\documentclass[10pt,twocolumn,letterpaper]{article}

\usepackage{cvpr}
\usepackage{times}
\usepackage{epsfig}
\usepackage{graphicx}
\usepackage{amsmath}
\usepackage{amssymb}
\usepackage{comment}

\usepackage[breaklinks=true,bookmarks=false]{hyperref}

\cvprfinalcopy 

\def\cvprPaperID{****} 

\begin{document}

\title{A Comprehensive Study of Deep Video Action Recognition}


\author{Yi Zhu, Xinyu Li, Chunhui Liu, Mohammadreza Zolfaghari, Yuanjun Xiong,\\ Chongruo Wu, Zhi Zhang, Joseph Tighe, R. Manmatha, Mu Li \\
Amazon Web Services \\
{\tt\small \{yzaws,xxnl,chunhliu,mozolf,yuanjx,chongrwu,zhiz,tighej,manmatha,mli\}@amazon.com}
}

\maketitle

\begin{abstract}
Video action recognition is one of the representative tasks for video understanding. Over the last decade, we have witnessed great advancements in video action recognition thanks to the emergence of deep learning. But we also encountered new challenges, including modeling long-range temporal information in videos, high computation costs, and incomparable results due to datasets and evaluation protocol variances. In this paper, we provide a comprehensive survey of over 200 existing papers on deep learning for video action recognition. We first introduce the 17 video action recognition datasets that influenced the design of models. Then we present video action recognition models in chronological order: starting with early attempts at adapting deep learning, then to the two-stream networks, followed by the adoption of 3D convolutional kernels, and finally to the recent compute-efficient models. In addition, we benchmark popular methods on several representative datasets and release code for reproducibility. In the end, we discuss open problems and shed light on opportunities for video action recognition to facilitate new research ideas. 
\end{abstract}

\section{Introduction}
\label{sec:introduction}

\begin{figure}[t]
\begin{center}
\includegraphics[width=1.0\columnwidth]{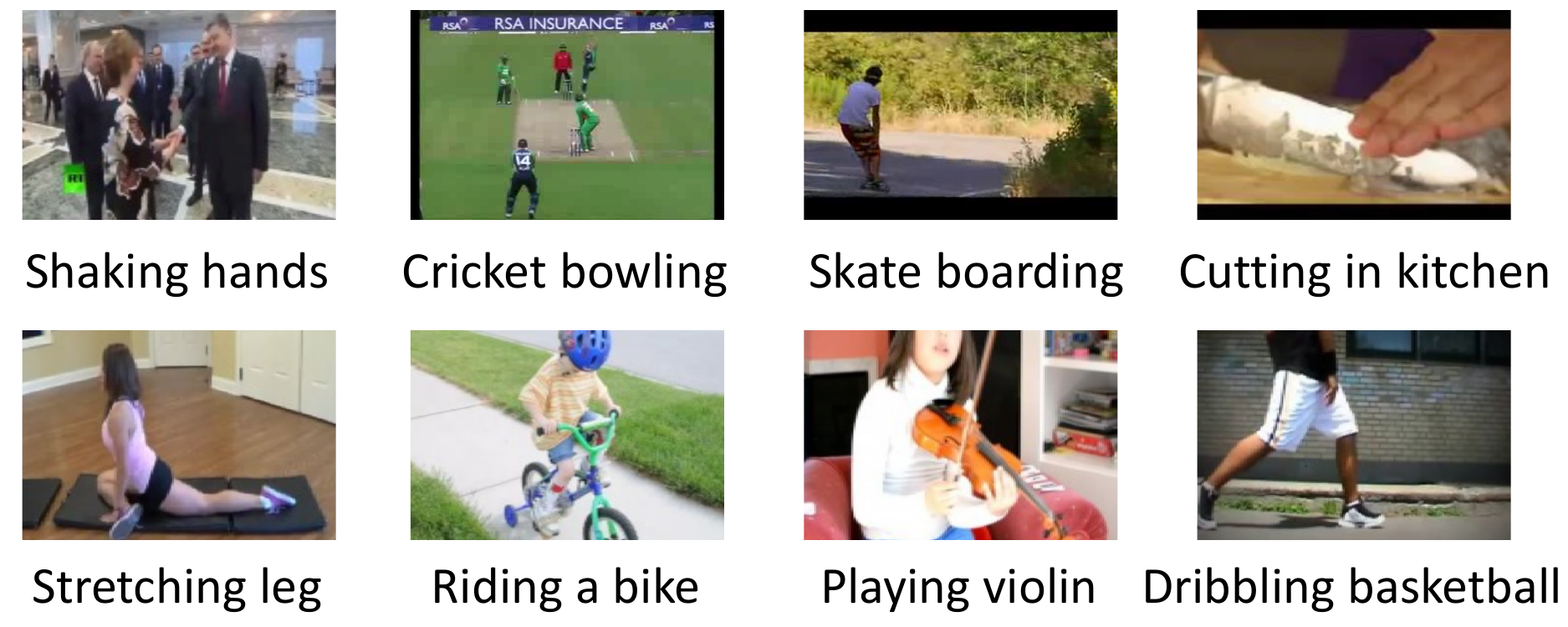}
\end{center}
\vspace{-2ex}
\caption{\textbf{Visual examples of categories in popular video action datasets}.}
\label{fig:dataset_teaser}
\end{figure}

One of the most important tasks in video understanding is to understand human actions. It has many real-world applications, including behavior analysis, video retrieval, human-robot interaction, gaming, and entertainment. Human action understanding involves recognizing, localizing, and predicting human behaviors. The task to recognize human actions in a video is called \textit{video action recognition}. In Figure \ref{fig:dataset_teaser}, we visualize several video frames with the associated action labels, which are typical human daily activities such as shaking hands and riding a bike. 

\begin{figure}[t]
\begin{center}
\includegraphics[width=1.0\columnwidth]{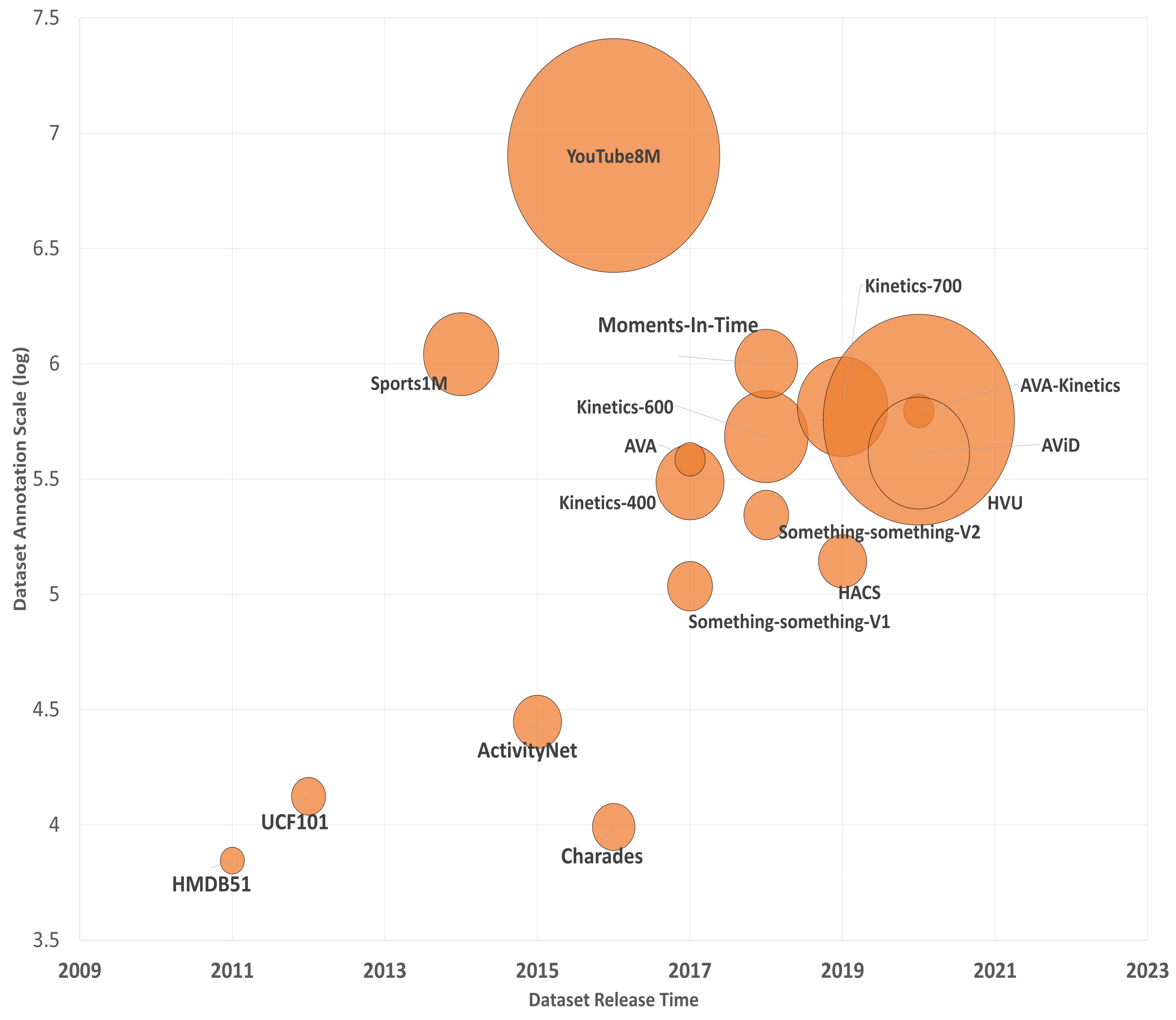}
\end{center}
\caption{\textbf{Statistics of most popular video action recognition datasets} from past 10 years. The area of an circle represents the scale of each dataset (i.e., number of videos).}
\label{fig:dataset_stat}
\end{figure}

\begin{figure*}[t]
\begin{center}
\includegraphics[width=1.0\linewidth]{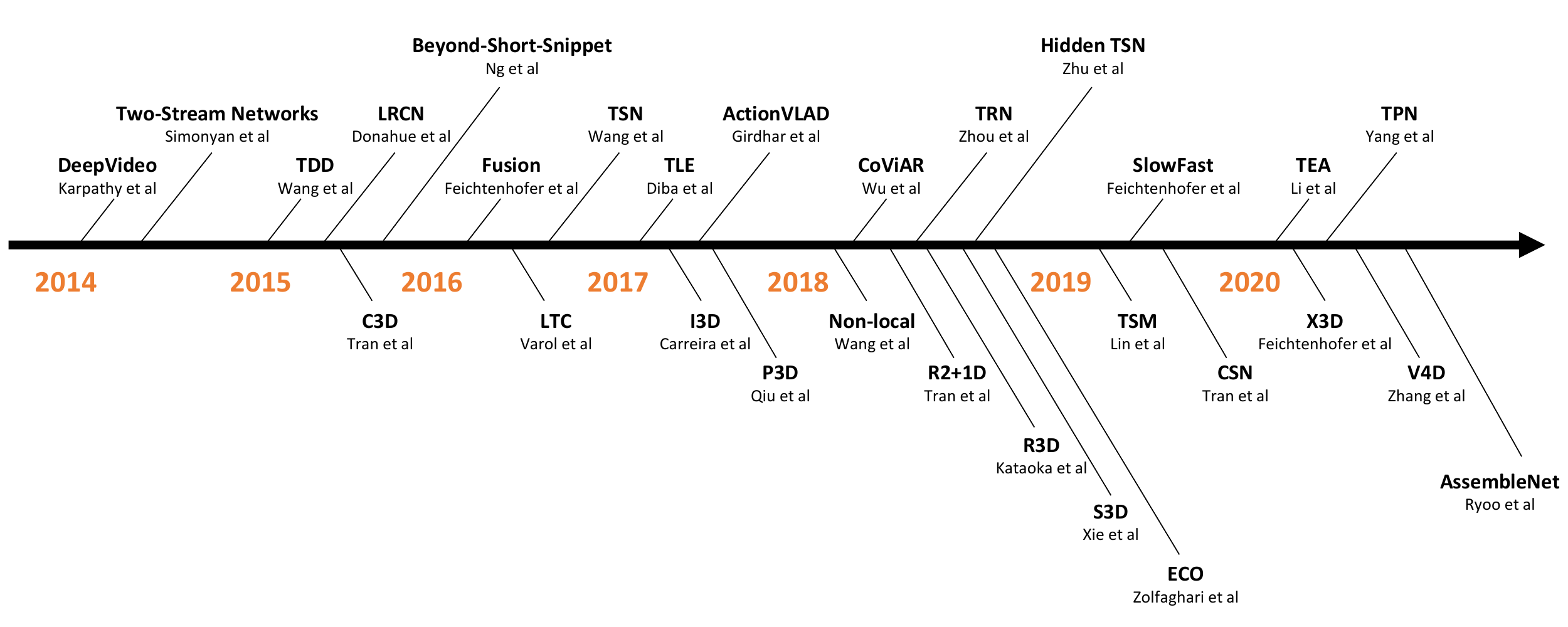}
\end{center}
\vspace{-2ex}
\caption{\textbf{A chronological overview of recent representative work in video action recognition}.} 
\label{fig:video_timeline}
\end{figure*}

Over the last decade, there has been growing research interest in video action recognition with the  emergence of high-quality large-scale action recognition datasets. We summarize the statistics of popular action recognition datasets in Figure~\ref{fig:dataset_stat}. We  see that both the number of videos and classes increase rapidly, e.g, from 7K videos over $51$ classes in HMDB51 \cite{hmdb51} to 8M videos over $3,862$ classes in YouTube8M \cite{youtube8m}. Also, the rate at which new datasets are released is  increasing: 3 datasets were released from 2011 to 2015 compared to 13 released from 2016 to 2020. 

Thanks to both the availability of large-scale datasets and the rapid progress in deep learning, there is also a rapid  growth in deep learning based models to recognize video actions. In Figure \ref{fig:video_timeline}, we present a chronological overview of recent representative work. DeepVideo \cite{karpathy_CVPR2014_videoCNN} is one of the earliest attempts to apply convolutional neural networks to videos. We observed three trends here. The first trend started by the seminal paper on Two-Stream Networks \cite{simonyan_NIPS2014_twoStream}, adds a second path to learn the temporal information in a video by training a convolutional neural network on the optical flow stream. Its great success inspired a large number of follow-up papers, such as TDD \cite{want_CVPR2015_trajectory}, LRCN \cite{donahue_CVPR2015_LRCN}, Fusion \cite{feichtenhofer_CVPR2016_fusion}, TSN \cite{wang_ECCV2016_TSN}, etc.  The second trend was the use of 3D convolutional kernels to model  video temporal information, such as I3D \cite{carreira_CVPR2017_I3D}, R3D \cite{hara_CVPR2018_3DResNet}, S3D \cite{xie_ECCV2018_S3D}, Non-local \cite{wang_CVPR2018_nonlocal}, SlowFast \cite{feichtenhofer_ICCV2019_slowfast}, etc. Finally, the third trend focused on computational efficiency to scale to even larger datasets so that they could be adopted in real applications. Examples include Hidden TSN \cite{zhu_ACCV2018_hidden}, TSM \cite{lin_ICCV2019_TSM}, X3D \cite{feichtenhofer_CVPR2020_X3D}, TVN \cite{piergiovanni_arxiv2019_TVN}, etc.

Despite the large number of deep learning based models for video action recognition, there is no comprehensive survey dedicated to these models. Previous survey papers either put more efforts into hand-crafted features \cite{herath_arxiv2016_survey,moreno_Sensors2019_survey} or focus on broader topics such as video captioning \cite{wu_arxiv2016_survey}, video prediction \cite{kong_arxiv2018_survey}, video action detection \cite{zhang_Sensors2018_survey} and zero-shot video action recognition \cite{estevam_arxiv2019_survey}. In this paper: 

\begin{itemize}
    \item We comprehensively review over 200 papers on deep learning for video action recognition. We walk the readers through the recent advancements chronologically and systematically, with popular papers explained in detail.
    \item We benchmark widely adopted methods on the same set of datasets in terms of both accuracy and efficiency. We also release our implementations for full reproducibility\footnote{Model zoo in both PyTorch and MXNet: \url{https://cv.gluon.ai/model_zoo/action_recognition.html}}.
    \item We elaborate on challenges, open problems, and opportunities in this field to facilitate future research.
\end{itemize}

The rest of the survey is organized as following. We first describe popular datasets used for benchmarking and existing challenges in section~\ref{sec:data_cha}. Then we present recent advancements using deep learning for video action recognition in section~\ref{sec:odyssey}, which is the major contribution of this survey. In section~\ref{sec:benchmark}, we evaluate widely adopted approaches on standard benchmark datasets, and provide discussions and future research opportunities in section~\ref{sec:future}.


\section{Datasets and Challenges}
\label{sec:data_cha}

\subsection{Datasets}
\label{subsec:datasets}
Deep learning methods usually improve in accuracy when the volume of the training data grows. In the case of video action recognition, this means we need large-scale annotated datasets to learn effective models. 

For the task of video action recognition, datasets are often built by the following process: (1) Define an action list, by combining labels from previous action recognition datasets and adding new categories depending on the use case. (2) Obtain videos from various sources, such as YouTube and movies, by matching the video title/subtitle to the action list. (3) Provide temporal annotations manually to indicate the start and end position of the action, and (4) finally clean up the dataset by de-duplication and filtering out noisy classes/samples. 
Below we review the most popular large-scale video action recognition datasets in Table~\ref{tab:dataset} and Figure~\ref{fig:dataset_stat}.

\begin{table}
\begin{center}
\scalebox{0.85}{
\begin{tabular}{l|c|c|c|c}
Dataset & Year & \# Samples& Ave. Len & \# Actions\\
\hline
HMDB51 \cite{hmdb51} & 2011 & 7K & $\sim$5s &51  \\
UCF101 \cite{ucf101} & 2012 & 13.3K & $\sim$6s &  101 \\
Sports1M \cite{karpathy_CVPR2014_videoCNN} & 2014 & 1.1M & $\sim$5.5m &487  \\
ActivityNet \cite{activityNet} & 2015 & 28K & $[5,10]m$&200  \\
YouTube8M \cite{youtube8m} & 2016 & 8M & $ 229.6s$ & 3862  \\
Charades \cite{charades} &2016 & 9.8K & $30.1s$ & 157  \\
Kinetics400 \cite{kinetics400} &2017 & 306K & $10s$ &400  \\
Kinetics600 \cite{carreira2018short} &2018 & 482K & $10s$ &600  \\
Kinetics700 \cite{carreira2019short} &2019 & 650K & $10s$ &700  \\
Sth-Sth V1 \cite{sthsth} &2017 & 108.5K & $[2,6]s$ &174  \\
Sth-Sth V2 \cite{sthsth} &2017 & 220.8K & $[2,6]s$ &174  \\
AVA \cite{ava} & 2017 & 385K & $15m$ & 80 \\
AVA-kinetics \cite{li2020ava} & 2020 & 624K & $15m,10s$ & 80 \\
MIT \cite{mit} & 2018 & 1M & $3s$  &339  \\
HACS Clips \cite{hacs} & 2019 & 1.55M & $2s$  &200  \\
HVU~\cite{hvu} & 2020 & 572K & $10s$ & 739\\
AViD~\cite{piergiovanni2020avid}& 2020 & 450K & $[3,15]s$ & 887 \\
\hline
\end{tabular}
}
\end{center}
\caption{\textbf{A list of popular datasets for video action recognition}}
\label{tab:dataset}
\vspace{-2ex}
\end{table}

\noindent \textbf{HMDB51} \cite{hmdb51} was introduced in 2011. It was collected mainly from movies, and a small proportion from public databases such as the Prelinger archive, YouTube and Google videos. The dataset contains $6,849$ clips divided into 51 action categories, each containing a minimum of 101 clips. The dataset has three official splits. Most previous papers either report the top-1 classification accuracy on split 1 or the average accuracy over three splits.\\
\textbf{UCF101} \cite{ucf101} was introduced in 2012 and is an extension of the previous UCF50 dataset. It contains $13,320$ videos from YouTube spreading over 101 categories of human actions. The dataset has three official splits similar to HMDB51, and is also evaluated in the same manner.\\
\textbf{Sports1M} \cite{karpathy_CVPR2014_videoCNN} was introduced in 2014 as the first large-scale video action dataset which consisted of more than 1 million YouTube videos annotated with 487 sports classes. The categories are fine-grained which leads to low inter-class variations. It has an official 10-fold cross-validation split for evaluation. \\ 
\noindent \textbf{ActivityNet} \cite{activityNet} was originally introduced in 2015 and the ActivityNet  family has several versions since its initial launch. The most recent ActivityNet 200 (V1.3) contains 200 human daily living actions. It has $10,024$ training, $4,926$ validation, and $5,044$ testing videos. On average there are 137 untrimmed videos per class and 1.41 activity instances per video.\\
\textbf{YouTube8M} \cite{youtube8m} was introduced in 2016 and is by far the largest-scale video dataset that contains 8 million YouTube videos (500K hours of video in total) and annotated with $3,862$ action classes. Each video is annotated with one or multiple labels by a YouTube video annotation system. 
This dataset is split into training, validation and test in the ratio 70:20:10. The validation set of this dataset is also extended with human-verified segment annotations to provide temporal localization information.\\
\textbf{Charades} \cite{charades} was introduced in 2016 as a dataset for real-life concurrent action understanding. It contains $9,848$ videos with an average length of 30 seconds. This dataset includes 157 multi-label daily indoor activities, performed by 267 different people. It has an official train-validation split that has $7,985$ videos for training and the remaining $1,863$ for validation.\\
\textbf{Kinetics Family} is now the most widely adopted benchmark. Kinetics400 \cite{kinetics400} was introduced in 2017 and it consists of approximately 240k training and 20k validation videos trimmed to 10 seconds from 400 human action categories. The Kinetics family continues to expand, with Kinetics-600 \cite{carreira2018short} released in 2018 with 480K videos and Kinetics700\cite{carreira2019short} in 2019 with 650K videos. \\
\textbf{20BN-Something-Something} \cite{sthsth} V1 was introduced in 2017 and V2 was introduced in 2018. This family is another popular benchmark that consists of 174 action classes that describe humans performing basic actions with everyday objects. There are $108,499$ videos in V1 and $220,847$ videos in V2. Note that the Something-Something dataset requires strong temporal modeling because most activities cannot be inferred based on spatial features alone (e.g. opening something, covering something with something). \\
\textbf{AVA} \cite{ava} was introduced in 2017 as the first large-scale spatio-temporal action detection dataset. It contains 430 15-minute video clips with 80 atomic actions labels (only 60 labels were used for evaluation). The annotations were provided at each key-frame which lead to $214,622$ training, $57,472$ validation and $120,322$ testing samples. The AVA dataset was recently expanded to AVA-Kinetics with $352,091$ training, $89,882$ validation and $182,457$ testing samples \cite{li2020ava}.\\
\textbf{Moments in Time} \cite{mit} was introduced in 2018 and it is a large-scale dataset designed for event understanding. It contains one million 3 second video clips, annotated with a dictionary of 339 classes. Different from other datasets designed for human action understanding, Moments in Time dataset involves people, animals, objects and natural phenomena. The dataset was extended to Multi-Moments in Time (M-MiT) \cite{monfort2019multi} in 2019 by increasing the number of videos to 1.02 million, pruning vague classes, and increasing the number of labels per video. \\
\textbf{HACS} \cite{hacs} was introduced in 2019 as a new large-scale dataset for recognition and localization of human actions collected from Web videos. 
It consists of two kinds of manual annotations. HACS Clips contains 1.55M 2-second clip annotations on 504K videos, and HACS Segments has 140K complete action segments (from action start to end) on 50K videos. The videos are annotated with the same 200 human action classes used in ActivityNet (V1.3) \cite{activityNet}.\\
\textbf{HVU}~\cite{hvu} dataset was released in 2020 for multi-label multi-task video understanding. This dataset has 572K videos and $3,142$ labels. The official split has 481K, 31K and 65K videos for train, validation, and test respectively. This dataset has six task categories: scene, object, action, event, attribute, and concept. On average, there are about $2,112$ samples for each label. The duration of the videos varies with a maximum
length of $10$ seconds.\\
\textbf{AViD}~\cite{piergiovanni2020avid} was introduced in 2020 as a dataset for anonymized action recognition. It contains 410K videos for training and 40K videos for testing. Each video clip duration is between 3-15 seconds and in total it has 887 action classes. During data collection, the authors tried to collect data from various countries to deal with data bias. They also remove face identities to protect privacy of video makers. Therefore, AViD dataset might not be a proper choice for recognizing face-related actions.

\begin{figure}[t]
\begin{center}
\includegraphics[width=1.0\columnwidth]{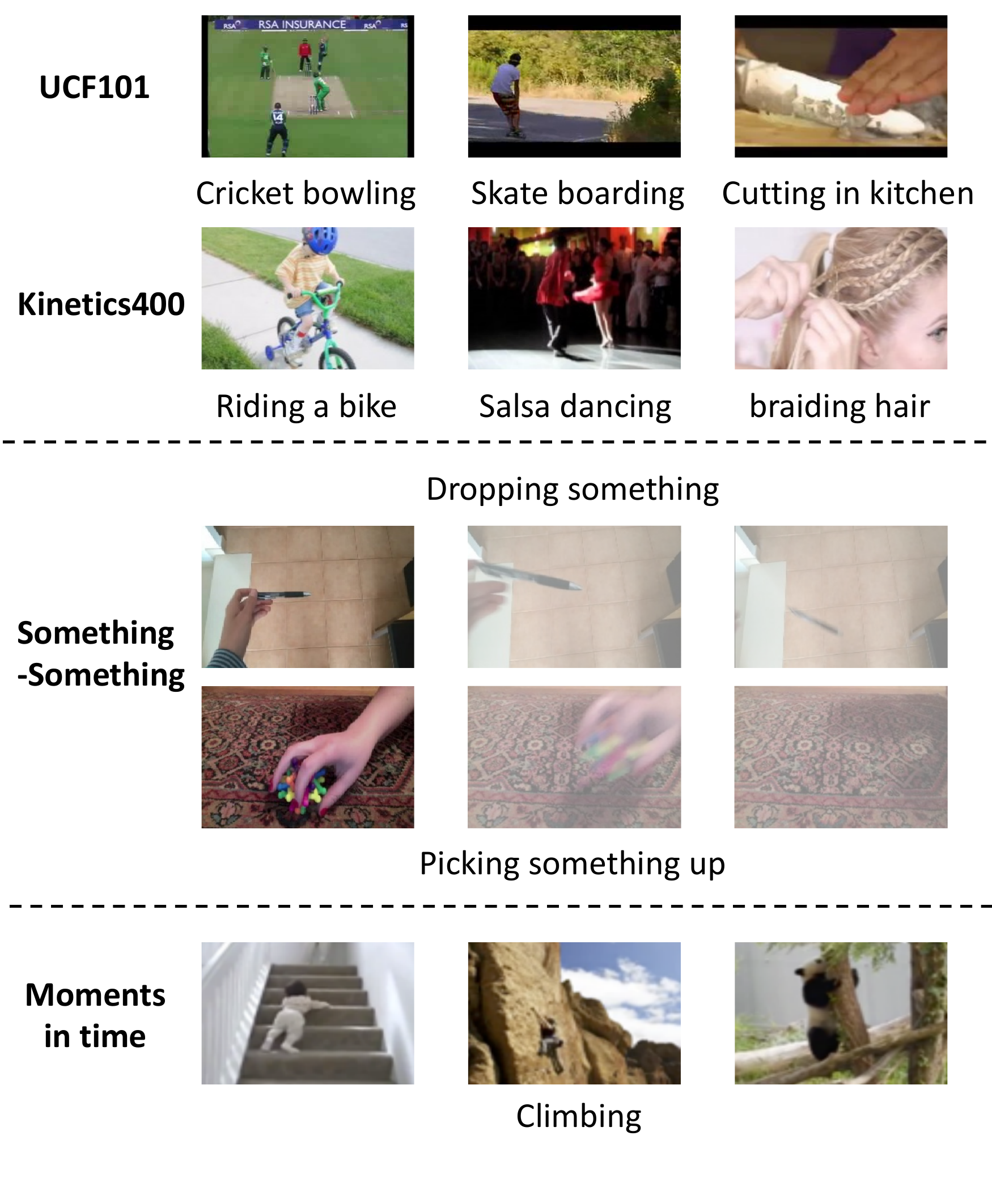}
\end{center}
\vspace{-6ex}
\caption{\textbf{Visual examples from popular video action datasets}. Top: individual video frames from action classes in UCF101 and Kinetics400. A single frame from these scene-focused datasets often contains enough information to correctly guess the category. Middle: consecutive video frames from classes in Something-Something. The 2nd and 3rd frames are made transparent to indicate the importance of temporal reasoning that we cannot tell these two actions apart by looking at the 1st frame alone. Bottom: individual video frames from classes in Moment in Time. Same action could have different actors in different environments.}
\vspace{-3ex}
\label{fig:dataset_comparison}
\end{figure}

Before we dive into the chronological review of methods, we present several visual examples from the above datasets in Figure \ref{fig:dataset_comparison} to show their different characteristics.
In the top two rows, we pick action classes from UCF101 \cite{ucf101} and Kinetics400 \cite{kinetics400} datasets. Interestingly, we find that these actions can sometimes be determined by the context or scene alone. For example, the model can predict the action riding a bike as long as it recognizes a bike in the video frame. The model may also predict the action cricket bowling if it recognizes the cricket pitch. Hence for these classes, video action recognition may become an object/scene classification problem without the need of reasoning motion/temporal information. 
In the middle two rows, we pick action classes from Something-Something dataset \cite{sthsth}. This dataset focuses on human-object interaction, thus it is more fine-grained and requires strong temporal modeling. For example, if we only look at the first frame of dropping something and picking something up without looking at other video frames, it is impossible to tell these two actions apart.
In the bottom row, we pick action classes from Moments in Time dataset \cite{mit}. This dataset is different from most video action recognition datasets, and is designed to have large inter-class and intra-class variation that represent dynamical events at different levels of abstraction. For example, the action climbing can have different actors (person or animal) in different environments (stairs or tree).

\subsection{Challenges}
\label{subsec:challenges}
There are several major challenges in developing effective video action recognition algorithms. 

In terms of dataset, first, defining the label space for training action recognition models is non-trivial. It's because human actions are usually composite concepts and the hierarchy of these concepts are not well-defined. 
Second, annotating videos for action recognition are laborious (e.g., need to watch all the video frames) and ambiguous (e.g, hard to determine the exact start and end of an action). 
Third, some popular benchmark datasets (e.g., Kinetics family) only release the video links for users to download and not the actual video, which leads to a situation that methods are evaluated on different data. It is impossible to do fair comparisons between methods and gain insights. 

In terms of modeling, first, videos capturing human actions have both strong intra- and inter-class variations. People can perform the same action in different speeds under various viewpoints. Besides, some actions share similar movement patterns that are hard to distinguish. 
Second, recognizing human actions requires simultaneous understanding of both short-term action-specific motion information and long-range temporal information. We might need a sophisticated model to handle different perspectives rather than using a single convolutional neural network.
Third, the computational cost is high for both training and inference, hindering both the development and deployment of action recognition models. In the next section, we will demonstrate how video action recognition methods developed over the last decade to address the aforementioned challenges.

\section{An Odyssey of Using Deep Learning for Video Action Recognition}
\label{sec:odyssey}
In this section, we review deep learning based methods for video action recognition from 2014 to present and introduce the related earlier work in context. 

\subsection{From hand-crafted features to CNNs}
\label{subsec:debut}

Despite there being some papers using Convolutional Neural Networks (CNNs) for video action recognition,  \cite{taylor_ECCV2010_CLST,baccouche_HBU2011_sequential,ji_PAMI2012_3DCNN}, hand-crafted features \cite{wang_CVPR2011_DT,wang_ICCV2013_IDT,peng_ECCV2014_stackedFV,lan_CVPR2015_MIFS}, particularly Improved Dense Trajectories (IDT) \cite{wang_ICCV2013_IDT}, dominated the video understanding literature before 2015, due to their high accuracy and good robustness. However, hand-crafted features have heavy computational cost \cite{xu_CVPR2015_LCD}, and are hard to scale and deploy.

With the rise of deep learning \cite{krizhevsky_NIPS2012_imagenetCNN}, researchers started to adapt CNNs for video problems. The seminal work DeepVideo \cite{karpathy_CVPR2014_videoCNN} proposed to use a single 2D CNN model on each video frame independently and investigated several temporal connectivity patterns to learn spatio-temporal features for video action recognition, such as late fusion, early fusion and slow fusion. Though this model made early progress with ideas that would prove to be useful later such as a multi-resolution network, its transfer learning performance on UCF101 \cite{ucf101} was $20\%$ less than hand-crafted IDT features ($65.4\%$ vs $87.9\%$). Furthermore, DeepVideo \cite{karpathy_CVPR2014_videoCNN} found that a network fed by individual video frames, performs equally well when the input is changed to a stack of frames. This observation might indicate that the learnt spatio-temporal features did not capture the motion well. 
It also encouraged people to think about why CNN models did not outperform traditional hand-crafted features in the video domain unlike in other computer vision tasks \cite{krizhevsky_NIPS2012_imagenetCNN,ren_NIPS2015_fasterRCNN}.

\subsection{Two-stream networks}
\label{subsec:two_stream}
Since video understanding intuitively needs motion information, finding an appropriate way to describe the temporal relationship between frames is essential to improving the performance of CNN-based video action recognition.

\begin{figure}[t]
\begin{center}
\includegraphics[width=1.0\linewidth]{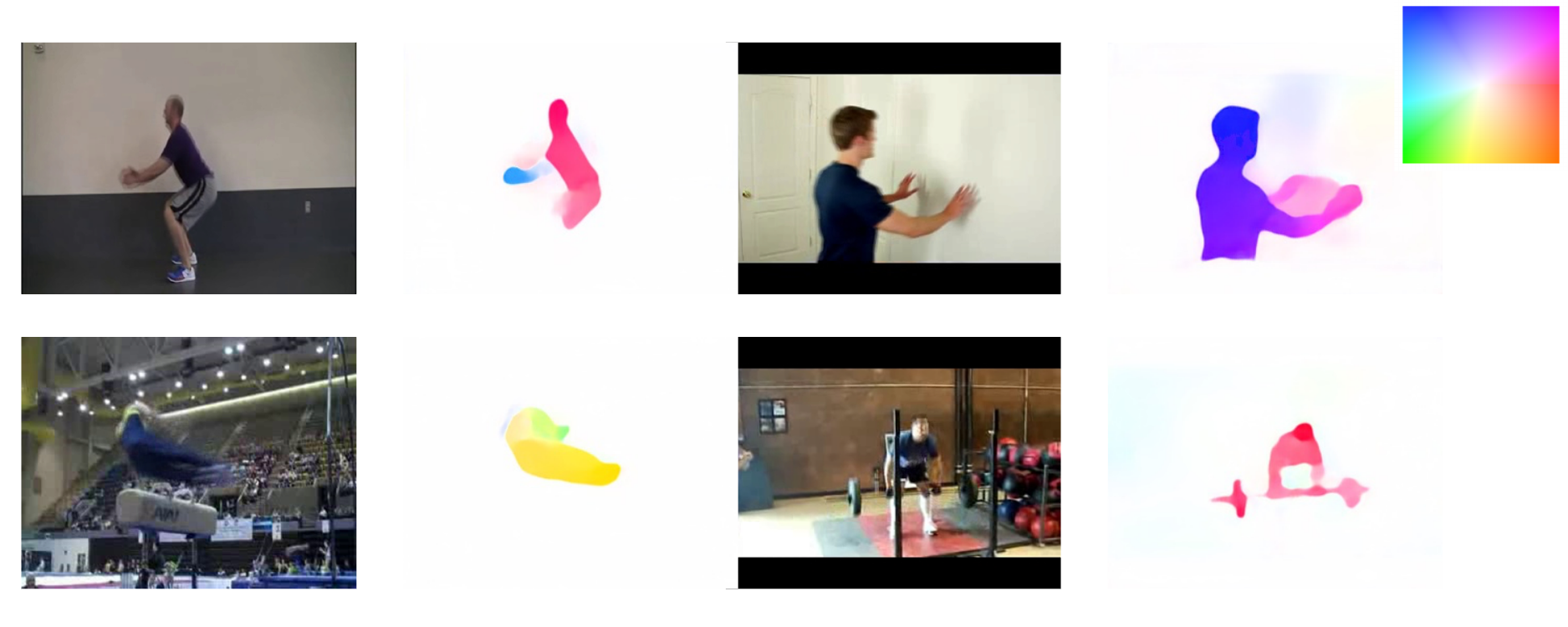}
\end{center}
\caption{Visualizations of optical flow. We show four image-flow pairs, left is original RGB image and right is the estimated optical flow by FlowNet2 \cite{flownet2}. Color of optical flow indicates the directions of motion, and we follow the color coding scheme of FlowNet2 \cite{flownet2} as shown in top right.}
\label{fig:optical_flow}
\end{figure}

Optical flow \cite{horn_AI1981_opticalFlow} is an effective motion representation to describe object/scene movement. To be precise, it is the pattern of apparent motion of objects, surfaces, and edges in a visual scene caused by the relative motion between an observer and a scene. We show several visualizations of optical flow in Figure \ref{fig:optical_flow}. As we can see, optical flow is able to describe the motion pattern of each action accurately. The advantage of using optical flow is it provides orthogonal information compared to the the RGB image. For example, the two images on the bottom of Figure \ref{fig:optical_flow} have cluttered backgrounds. Optical flow can effectively remove the non-moving background and result in a simpler learning problem compared to using the original RGB images as input. 
In addition, optical flow has been shown to work well on video problems. Traditional hand-crafted features such as IDT \cite{wang_ICCV2013_IDT} also contain optical-flow-like features, such as Histogram of Optical Flow (HOF) and Motion Boundary Histogram (MBH).

\begin{figure}[t]
\begin{center}
\includegraphics[width=1.0\linewidth]{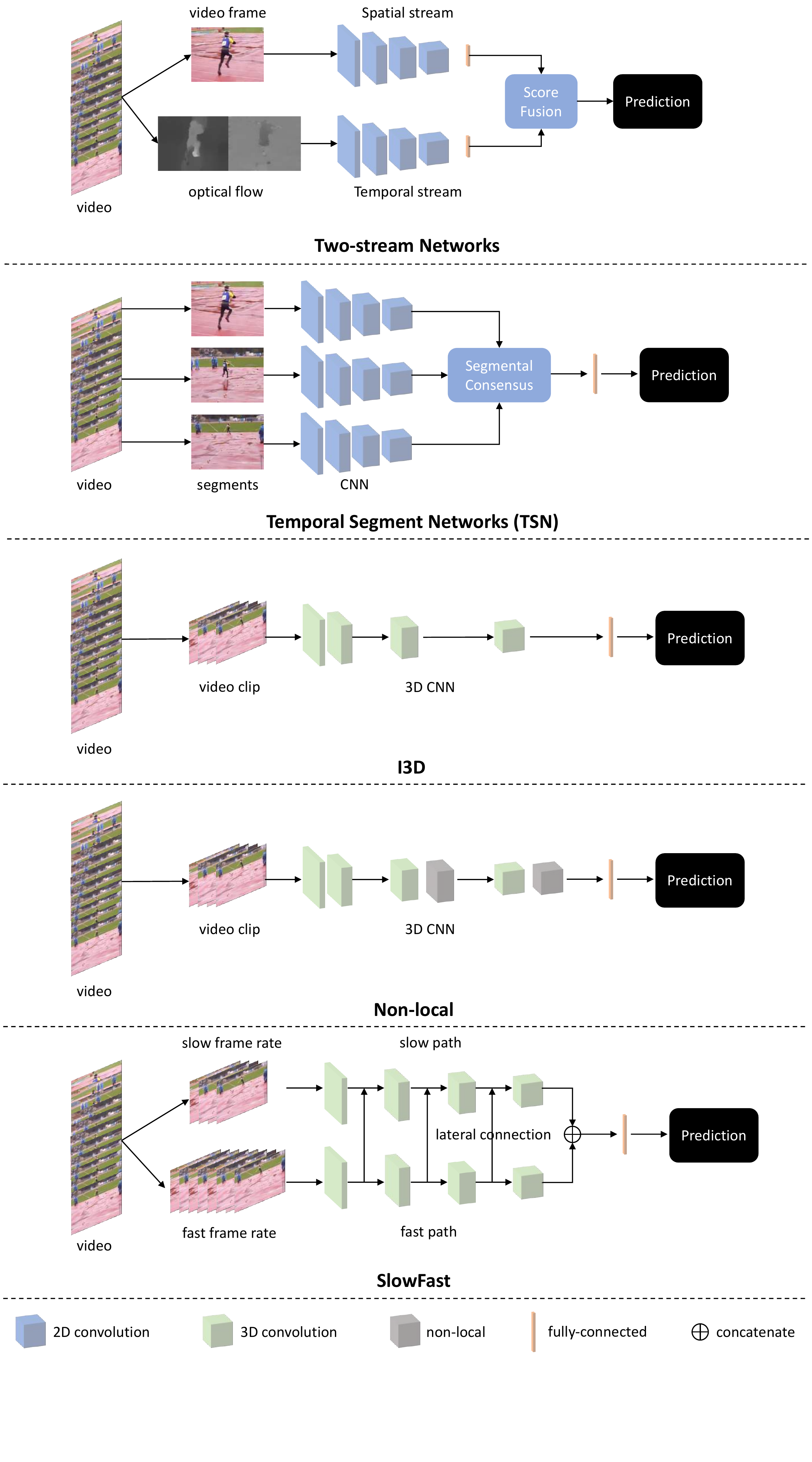}
\end{center}
\vspace{-6ex}
\caption{\textbf{Workflow of five important papers}: two-stream networks \cite{simonyan_NIPS2014_twoStream}, temporal segment networks \cite{wang_ECCV2016_TSN}, I3D \cite{carreira_CVPR2017_I3D}, Non-local \cite{wang_CVPR2018_nonlocal} and SlowFast \cite{feichtenhofer_ICCV2019_slowfast}. Best viewed in color.} 
\label{fig:milestone}
\end{figure}

Hence, Simonyan \etal \cite{simonyan_NIPS2014_twoStream} proposed two-stream networks, which included a spatial stream and a temporal stream as shown in Figure \ref{fig:milestone}. This method is related to the two-streams hypothesis \cite{Goodale_Neuro1992_separate}, according to which the human visual cortex contains two pathways: the ventral stream (which performs object recognition) and the dorsal stream (which recognizes motion). 
The spatial stream takes raw video frame(s) as input to capture visual appearance information. The temporal stream takes a stack of optical flow images as input to capture motion information between video frames.
To be specific, \cite{simonyan_NIPS2014_twoStream} linearly rescaled the horizontal and vertical components of the estimated flow (i.e., motion in the x-direction and y-direction) to a [0, 255] range and compressed using JPEG. The output corresponds to the two optical flow images shown in Figure \ref{fig:milestone}. 
The compressed optical flow images will then be concatenated as the input to the temporal stream with a dimension of $H \times W \times 2L$, where H, W and L indicates the height, width and the length of the video frames. In the end, the final prediction is obtained by averaging the prediction scores from both streams.

By adding the extra temporal stream, for the first time, a CNN-based approach achieved performance similar to the previous best hand-crafted feature IDT on UCF101 ($88.0\%$ vs $87.9\%$) and on HMDB51 \cite{hmdb51} ($59.4\%$ vs $61.1\%$).
\cite{simonyan_NIPS2014_twoStream} makes two important observations. First, motion information is important for video action recognition. Second, it is still challenging for CNNs to learn temporal information directly from raw video frames. Pre-computing optical flow as the motion representation is an effective way for deep learning to reveal its power.
Since \cite{simonyan_NIPS2014_twoStream} managed to close the gap between deep learning approaches and traditional hand-crafted features, many follow-up papers on two-stream networks emerged and greatly advanced the development of video action recognition. Here, we divide them into several categories and review them individually.

\subsubsection{Using deeper network architectures}
\label{subsubsec:deeper}
Two-stream networks \cite{simonyan_NIPS2014_twoStream} used a relatively shallow network architecture \cite{krizhevsky_NIPS2012_imagenetCNN}. Thus a natural extension to the two-stream networks involves using deeper networks. However, Wang \etal \cite{wang_THUMOS2015_challenge} finds that simply using deeper networks does not yield better results, possibly due to overfitting on the small-sized video datasets \cite{ucf101,hmdb51}. Recall from section \ref{subsec:datasets}, UCF101 and HMDB51 datasets only have thousands of training videos. Hence, Wang \etal \cite{wang_arxiv2015_good} introduce a series of good practices, including cross-modality initialization, synchronized batch normalization, corner cropping and multi-scale cropping data augmentation, large dropout ratio, etc. to prevent deeper networks from overfitting. With these good practices, \cite{wang_arxiv2015_good} was able to train a two-stream network with the VGG16 model \cite{simonyan_ICLR2015_VGG} that outperforms \cite{simonyan_NIPS2014_twoStream} by a large margin  on UCF101. These good practices have been widely adopted and are still being used. Later, Temporal Segment Networks (TSN) \cite{wang_ECCV2016_TSN}  performed a thorough investigation of network architectures, such as VGG16, ResNet \cite{he_CVPR2016_resnet}, Inception \cite{szegedy_CVPR2015_inception}, and demonstrated that deeper networks usually achieve higher recognition accuracy for video action recognition. We will describe more details about TSN in section \ref{subsubsec:tsn}.

\subsubsection{Two-stream fusion}
\label{subsubsec:ts_fusion}
Since there are two streams in a two-stream network, there will be a stage that  needs to merge the results from both networks to obtain the final prediction. This stage is usually referred to as the spatial-temporal fusion step. 

The easiest and most straightforward way is late fusion, which performs a weighted average of predictions from both streams. Despite late fusion being widely adopted \cite{simonyan_NIPS2014_twoStream,wang_arxiv2015_good}, many researchers claim that this may not be the optimal way to fuse the information between the spatial appearance stream and temporal motion stream. They believe that earlier interactions between the two networks could benefit both streams during model learning and this is termed as early fusion. 

Fusion \cite{feichtenhofer_CVPR2016_fusion} is one of the first of several papers investigating the early fusion paradigm, including how to perform spatial fusion (e.g., using operators such as sum, max, bilinear, convolution and concatenation), where to fuse the network (e.g., the network layer where early interactions happen), and how to perform temporal fusion (e.g., using 2D or 3D convolutional fusion in later stages of the network). \cite{feichtenhofer_CVPR2016_fusion} shows that early fusion is beneficial for both streams to learn richer features and leads to improved performance over late fusion. Following this line of research, Feichtenhofer \etal \cite{feichtenhofer_NIPS2016_stResidual} generalizes ResNet \cite{he_CVPR2016_resnet} to the spatio-temporal domain by introducing residual connections between the two streams. Based on \cite{feichtenhofer_NIPS2016_stResidual}, Feichtenhofer \etal \cite{feichtenhofer_CVPR2017_stMultiplier} further propose a multiplicative gating function for residual networks to learn better spatio-temporal features. Concurrently, \cite{wang_CVPR2017_pyramid} adopts a spatio-temporal pyramid to perform hierarchical early fusion between the two streams. 

\subsubsection{Recurrent neural networks}
\label{subsubsec:rnn}
Since a video is essentially a temporal sequence, researchers have explored Recurrent Neural Networks (RNNs) for temporal modeling inside a video, particularly the usage of Long Short-Term Memory (LSTM) \cite{hochreiter_NC1997_LSTM}.

LRCN \cite{donahue_CVPR2015_LRCN} and Beyond-Short-Snippets \cite{ng_CVPR2015_beyondLSTM} are the first of several papers that use LSTM for video action recognition under the two-stream networks setting. They take the feature maps from CNNs as an input to a deep LSTM network, and aggregate frame-level CNN features into video-level predictions. Note that they use LSTM on two streams separately, and the final results are still obtained by late fusion. However, there is no clear empirical improvement from LSTM models \cite{ng_CVPR2015_beyondLSTM} over the two-stream baseline \cite{simonyan_NIPS2014_twoStream}. Following the CNN-LSTM framework, several variants are proposed, such as bi-directional LSTM \cite{ullah_access2017_biLSTM}, CNN-LSTM fusion \cite{gammulle_WACV2017_TS_LSTM_fusion} and hierarchical multi-granularity LSTM network \cite{li_ICMR2016_granular}.
\cite{li_CVIU2018_videoLSTM} described VideoLSTM which includes a correlation-based spatial attention mechanism and a lightweight motion-based attention mechanism. VideoLSTM not only show improved results on action recognition, but also demonstrate how the learned attention can be used for action localization by relying on just the action class label. Lattice-LSTM \cite{sun_ICCV2017_L2STM} extends LSTM by learning independent hidden state transitions of memory cells for individual spatial locations, so that it can accurately model long-term and complex motions. ShuttleNet \cite{Shi_ICCV2017_bioinspire} is a concurrent work that considers both feedforward and feedback connections in a RNN to learn long-term dependencies. FASTER \cite{zhu2020faster} designed a FAST-GRU to aggregate clip-level features from an expensive backbone and a cheap backbone. This strategy reduces the processing cost of redundant clips and hence accelerates the inference speed.

However, the work mentioned above \cite{donahue_CVPR2015_LRCN,ng_CVPR2015_beyondLSTM,li_CVIU2018_videoLSTM,sun_ICCV2017_L2STM,Shi_ICCV2017_bioinspire} use different two-stream networks/backbones. The differences between various methods using RNNs are thus unclear. Ma \etal \cite{ma_SP2019_TS_LSTM} build a strong baseline for fair comparison and thoroughly study the effect of learning spatio-temporal features by using RNNs. They find that it requires proper care to achieve improved performance, e.g., LSTMs require pre-segmented data to fully exploit the temporal information. RNNs are also intensively studied in video action localization \cite{singh_CVPR2016_BiRNN} and video question answering \cite{zhu_IJCV2017_vqa}, but these are beyond the scope of this survey.

\subsubsection{Segment-based methods}
\label{subsubsec:tsn}
Thanks to optical flow, two-stream networks are able to reason about short-term motion information between frames. However, they still cannot capture long-range temporal information. 
Motivated by this weakness of two-stream networks , Wang \etal \cite{wang_ECCV2016_TSN} proposed a Temporal Segment Network (TSN) to perform video-level action recognition. Though initially proposed to be used with 2D CNNs, it is simple and generic. Thus recent work using either 2D or 3D CNNs, are still built upon this framework. 

To be specific, as shown in Figure \ref{fig:milestone}, TSN first divides a whole video into several segments, where the segments distribute uniformly along the temporal dimension. Then TSN randomly selects a single video frame within each segment and forwards them through the network. Here, the network shares weights for input frames from all the segments. In the end, a segmental consensus is performed to aggregate information from the sampled video frames. The segmental consensus could be operators like average pooling, max pooling, bilinear encoding, etc. In this sense, TSN is capable of modeling long-range temporal structure because the model sees the content from the entire video. In addition, this sparse sampling strategy lowers the training cost over long video sequences but preserves relevant information. 

Given TSN's good performance and simplicity, most two-stream methods afterwards become segment-based two-stream networks. 
Since the segmental consensus is simply doing a max or average pooling operation, a feature encoding step might generate a global video feature and lead to improved performance as suggested in traditional approaches \cite{sanchez_IJCV2013_fisherVector,jegou_CVPR2010_VLAD,peng_arxiv2014_bovwSurvey}. Deep Local Video Feature (DVOF) \cite{lan_CVPRW2017_DVOF} proposed to treat the deep networks that trained on local inputs as feature extractors and train another encoding function to map the
global features into global labels. Temporal Linear Encoding (TLE) network \cite{diba_CVPR2017_TLE} appeared concurrently with DVOF, but the encoding layer was embedded in the network so that the whole pipeline could be trained end-to-end. VLAD3 and ActionVLAD \cite{Li_CVPR2016_VLAD3,girdhar_CVPR2017_actionVLAD} also appeared concurrently. They extended the NetVLAD layer \cite{arandjelovic_CVPR2016_netVLAD} to the video domain to perform video-level encoding, instead of using compact bilinear encoding as in \cite{diba_CVPR2017_TLE}. To improve the temporal reasoning ability of TSN, Temporal Relation Network (TRN) \cite{zhou_ECCV2018_TRN} was proposed to learn and reason about temporal dependencies between video frames at multiple time scales. The recent state-of-the-art efficient model TSM \cite{lin_ICCV2019_TSM} is also segment-based. We will discuss it in more detail in section \ref{subsubsec:no3d}.

\subsubsection{Multi-stream networks}
\label{subsubsec:multi_stream}
Two-stream networks are successful because appearance and motion information are two of the most important properties of a video. However, there are other factors that can help video action recognition as well, such as pose, object, audio and depth, etc.

Pose information is closely related to human action. We can recognize most actions by just looking at a pose (skeleton) image without scene context. Although there is previous work on using pose for action recognition \cite{nie_CVPR2015_jointPose,stgcn2018aaai}, P-CNN \cite{cheron_ICCV2015_PCNN} was one of the first deep learning methods that successfully used pose to improve video action recognition. P-CNN proposed to aggregates motion and appearance information along tracks of human body parts, in a similar spirit to trajectory pooling \cite{want_CVPR2015_trajectory}. \cite{zolfaghari_ICCV2017_chained} extended this pipeline to a chained multi-stream framework, that computed and integrated appearance, motion and pose. They introduced a Markov chain model that added these cues successively and obtained promising results on both action recognition and action localization. PoTion \cite{vasileios_CVPR2018_potion} was a follow-up work to P-CNN, but introduced a more powerful feature representation that encoded the movement of human semantic keypoints. They first ran a decent human pose estimator and extracted heatmaps for the human joints in each frame. They then obtained the PoTion representation by temporally aggregating these probability maps. PoTion is lightweight and outperforms previous pose representations \cite{cheron_ICCV2015_PCNN,zolfaghari_ICCV2017_chained}. In addition, it was shown to be complementary to standard appearance and motion streams, e.g. combining PoTion with I3D \cite{carreira_CVPR2017_I3D} achieved state-of-the-art result on UCF101 ($98.2\%$). 

Object information is another important cue because most human actions involve human-object interaction. Wu \cite{wu_CVPR2016_objectScene} proposed to leverage both object features and scene features to help video action recognition. The object and scene features were extracted from state-of-the-art pre-trained object and scene detectors. Wang \etal \cite{wang_BMVC2016_SRCNN} took a step further to make the network end-to-end trainable. They introduced a two-stream semantic region based method, by replacing a standard spatial stream with a Faster RCNN network \cite{ren_NIPS2015_fasterRCNN}, to extract semantic information about the object, person and scene. 

Audio signals usually come with video, and are complementary to the visual information. Wu \etal \cite{wu_MM2016_multiStream} introduced a multi-stream framework that integrates spatial, short-term motion, long-term temporal and audio in videos to digest complementary clues. Recently, Xiao \etal \cite{xiao_arxiv2020_audioslowfast} introduced AudioSlowFast following \cite{feichtenhofer_ICCV2019_slowfast}, by adding another audio pathway to model vision and sound in an unified representation.

In RGB-D video action recognition field, using depth information is  standard practice \cite{garcia_ECCV2018_distill}. However, for vision-based video action recognition (e.g., only given monocular videos), we do not have access to ground truth depth information as in the RGB-D domain. An early attempt Depth2Action \cite{zhu_ECCVW2016_depth2action} uses off-the-shelf depth estimators to extract depth information from videos and use it for action recognition.

Essentially, multi-stream networks is a way of multi-modality learning, using different cues as input signals to help video action recognition. We will discuss more on multi-modality learning in section \ref{subsec:multi_modality}.

\subsection{The rise of 3D CNNs}
\label{subsec:3dcnn}
Pre-computing optical flow is computationally intensive and storage demanding, which is not friendly for large-scale training or real-time deployment. A conceptually easy way to understand a video is as a 3D tensor with two spatial and one time dimension. Hence, this leads to the usage of 3D CNNs as a processing unit to model the temporal information in a video.


The seminal work for using 3D CNNs for action recognition is \cite{ji_PAMI2012_3DCNN}. While inspiring, the network was not deep enough to show its potential. Tran \etal \cite{tran_ICCV2015_C3D} extended \cite{ji_PAMI2012_3DCNN} to a deeper 3D network, termed C3D. C3D follows the modular design of \cite{simonyan_ICLR2015_VGG}, which could be thought of as a 3D version of VGG16 network. Its performance on standard benchmarks is not satisfactory, but shows strong generalization capability and can be used as a generic feature extractor for various video tasks \cite{li_ICCV2015_tempStructure}. 

However, 3D networks are hard to optimize. In order to train a 3D convolutional filter well, people need a large-scale dataset with diverse video content and action categories. Fortunately, there exists a dataset, Sports1M \cite{karpathy_CVPR2014_videoCNN}  which is large enough to support the training of a deep 3D network. However, the training of C3D takes weeks to converge. Despite the popularity of C3D, most users just adopt it as a feature extractor for different use cases instead of modifying/fine-tuning the network. This is partially the reason why two-stream networks based on 2D CNNs dominated the video action recognition domain from year 2014 to 2017.

The situation changed when Carreira \etal \cite{carreira_CVPR2017_I3D} proposed I3D in year 2017. As shown in Figure \ref{fig:milestone}, I3D takes a video clip as input, and forwards it through stacked 3D convolutional layers. A video clip is a sequence of video frames, usually 16 or 32 frames are used. The major contributions of I3D are: 1) it adapts mature image classification architectures to use for 3D CNN; 2) For model weights, it adopts a method developed for initializing optical flow networks in~\cite{wang_arxiv2015_good} to inflate the ImageNet pre-trained 2D model weights to their counterparts in the 3D model. Hence, I3D bypasses the dilemma that 3D CNNs have to be trained from scratch.  With pre-training on a new large-scale dataset Kinetics400 \cite{kinetics400}, I3D achieved a  $95.6\%$ on UCF101 and $74.8\%$ on HMDB51. I3D ended the era where different methods reported numbers on small-sized datasets such as UCF101 and HMDB51\footnote{As we can see in Table \ref{tab:scene_results}}. Publications following I3D needed to report their performance on Kinetics400, or other large-scale benchmark datasets, which pushed video action recognition to the next level. In the next few years, 3D CNNs advanced quickly and became top performers on almost every benchmark dataset. We will review the 3D CNNs based literature in several categories below.

We want to point out that 3D CNNs are not replacing two-stream networks, and they are not mutually exclusive. They just use different ways to model the temporal relationship in a video.
Furthermore, the two-stream approach is a generic framework for video understanding, instead of a specific method. As long as there are two networks, one  for spatial appearance modeling using RGB frames, the other  for temporal motion modeling using optical flow, the method may be categorized into the family of two-stream networks. 
In \cite{carreira_CVPR2017_I3D}, they also build a temporal stream with I3D architecture and achieved even higher performance, $98.0\%$ on UCF101 and $80.9\%$ on HMDB51. Hence, the final I3D model is a combination of 3D CNNs and two-stream networks. However, the contribution of I3D does not lie in the usage of optical flow.

\subsubsection{Mapping from 2D to 3D CNNs}
\label{subsubsec:2dto3d}
2D CNNs enjoy the benefit of pre-training brought by the large-scale of image datasets such as ImageNet~\cite{imagenet_cvpr09} and Places205~\cite{zhou2017places}, which cannot be matched even with the largest video datasets available today. On these datasets numerous efforts have been devoted to the search for 2D CNN architectures that are more accurate and generalize better. Below we describe the efforts to capitalize on these advances for 3D CNNs.

ResNet3D \cite{hara_CVPR2018_3DResNet}  directly took 2D ResNet \cite{he_CVPR2016_resnet} and replaced all the 2D convolutional filters with 3D kernels. They believed that by using deep 3D CNNs together with large-scale datasets one can exploit the success  of 2D CNNs on ImageNet. Motivated by ResNeXt \cite{xie_CVPR2017_resnext}, Chen \etal \cite{chen_ECCV2018_multiFiber} presented a multi-fiber architecture that slices a complex neural network into an ensemble of lightweight networks (fibers) that facilitate information flow between fibers, reduces the computational cost at the same time. Inspired by SENet \cite{hu_CVPR2018_senet}, STCNet \cite{diba_ECCV2018_stcnet} propose to integrate channel-wise information inside a 3D block to capture both spatial-channels and temporal-channels correlation information throughout the network. 

\subsubsection{Unifying 2D and 3D CNNs}
\label{subsubsec:unify}
To reduce the complexity of 3D network training, P3D \cite{qiu_ICCV2017_P3D} and R2+1D \cite{tran_CVPR2018_R2plus1D} explore the idea of 3D factorization. To be specific, a 3D kernel (e.g., $3 \times 3 \times 3$) can be factorized to two separate operations, a 2D spatial convolution (e.g., $1 \times 3 \times 3$) and a 1D temporal convolution (e.g., $3 \times 1 \times 1$). The differences between P3D and R2+1D are how they arrange the two factorized operations and how they formulate each residual block. Trajectory convolution~\cite{zhao_NIPS2018_trajconv} follows this idea but uses deformable convolution for the temporal component to better cope with motion.

Another way of simplifying 3D CNNs is to mix 2D and 3D convolutions in a single network. MiCTNet \cite{zhou_CVPR2018_MiCT} integrates 2D and 3D CNNs to generate deeper and more informative feature maps, while reducing training complexity in each round of spatio-temporal fusion. ARTNet \cite{wang_CVPR2018_ARTNet} introduces an appearance-and-relation network by using a new building block. The building block consists of a spatial branch using 2D CNNs and a relation branch using 3D CNNs. S3D \cite{xie_ECCV2018_S3D} combines the merits from approaches mentioned above. It first replaces the 3D convolutions at the bottom of the network with 2D kernels, and find that this kind of top-heavy network has higher recognition accuracy. Then S3D factorizes the remaining 3D kernels as P3D and R2+1D do, to further reduce the model size and training complexity. A concurrent work named ECO \cite{zolfaghari_ECCV2018_ECO} also adopts such a top-heavy network to achieve online video understanding.


\subsubsection{Long-range temporal modeling}
\label{subsubsec:long_range_3d}
In 3D CNNs, long-range temporal connection may be achieved by stacking multiple short temporal convolutions, e.g., $3 \times 3 \times 3$ filters. However, useful temporal information may be lost in the later stages of a deep network, especially for frames far apart. 

In order to perform long-range temporal modeling, LTC \cite{varol_PAMI18_ltc} introduces and evaluates long-term temporal
convolutions over a large number of video frames. However, limited by GPU memory, they have to sacrifice input resolution to use more frames. After that, T3D \cite{diba_arxiv2017_T3D} adopted a densely connected structure \cite{huang_CVPR2017_densenet} to keep the original temporal information as complete as possible to make the final prediction. Later, Wang \etal \cite{wang_CVPR2018_nonlocal} introduced a new building block, termed non-local. Non-local is a generic operation similar to self-attention \cite{vaswani_NIPS2017_attention}, which can be used for many computer vision tasks in a plug-and-play manner. As shown in Figure \ref{fig:milestone}, they used a spacetime non-local module after later residual blocks to capture the long-range dependence in both space and temporal domain, and achieved improved performance over baselines without bells and whistles.
Wu \etal \cite{wu_CVPR2019_featureBank} proposed a feature bank representation, which embeds information of the entire video into a memory cell, to make context-aware prediction. Recently, V4D \cite{zhang_ICLR2020_V4D} proposed video-level 4D CNNs, to model the evolution of long-range spatio-temporal representation with 4D convolutions.

\subsubsection{Enhancing 3D efficiency}
\label{subsubsec:3d_efficiency}
In order to further improve the efficiency of 3D CNNs (i.e., in terms of GFLOPs, model parameters and latency), many variants of 3D CNNs begin to emerge.

Motivated by the development in efficient 2D networks, researchers started to adopt  channel-wise separable convolution and extend it for video classification \cite{kopuklu_ICCVW2019_resource,tran_ICCV2019_CSN}. CSN \cite{tran_ICCV2019_CSN} reveals that it is a good practice to factorize 3D convolutions by separating channel interactions and spatio-temporal interactions, and is able to obtain state-of-the-art performance while being 2 to 3 times faster than the previous best approaches. These methods are also related to multi-fiber networks \cite{chen_ECCV2018_multiFiber} as they are all inspired by group convolution. 

Recently, Feichtenhofer \etal \cite{feichtenhofer_ICCV2019_slowfast} proposed SlowFast, an efficient network with a slow pathway and a fast pathway. The network design is partially inspired by the biological Parvo- and Magnocellular cells in the primate visual systems.
As shown in Figure \ref{fig:milestone}, the slow pathway operates at low frame rates to capture detailed semantic information, while the fast pathway operates at high temporal resolution to capture rapidly changing motion. In order to incorporate motion information such as in two-stream networks, SlowFast adopts a lateral connection to fuse the representation learned by each pathway.
Since the fast pathway can be made very lightweight by reducing its channel capacity, the overall efficiency of SlowFast is largely improved. 
Although SlowFast has two pathways, it is different from the two-stream networks \cite{simonyan_NIPS2014_twoStream}, because the two pathways are designed to model different temporal speeds, not spatial and temporal modeling. There are several concurrent papers using multiple pathways to balance the accuracy and efficiency \cite{fan_NIPS2019_bLVNet}.

Following this line, Feichtenhofer \cite{feichtenhofer_CVPR2020_X3D} introduced X3D that progressively expand a 2D image classification architecture along multiple network axes, such as temporal duration, frame rate, spatial resolution, width, bottleneck width, and depth. X3D pushes the 3D model modification/factorization to an extreme, and is a family of efficient video networks to meet different requirements of target complexity. 
With similar spirit, A3D~\cite{zhu2020a3d} also leverages multiple network configurations. However, A3D trains these configurations jointly and during inference deploys only one model. This makes the model at the end more efficient. In the next section, we will continue to talk about efficient video modeling, but not based on 3D convolutions.

\subsection{Efficient Video Modeling}
\label{subsec:2dreturn}
With the increase of dataset size and the need for deployment, efficiency becomes an important concern.

If we use methods based on two-stream networks, we need to pre-compute optical flow and store them on local disk. Taking Kinetics400 dataset as an illustrative example, storing all the optical flow images requires 4.5TB disk space. Such a huge amount of data would make I/O become the tightest bottleneck during training, leading to a waste of GPU resources and longer experiment cycle. In addition, pre-computing optical flow is not cheap, which means all the two-stream networks methods are not real-time.

If we use methods based on 3D CNNs, people still find that 3D CNNs are hard to train and challenging to deploy. In terms of training, a standard SlowFast network trained on Kinetics400 dataset using a high-end 8-GPU machine takes 10 days to complete. Such a long experimental cycle and huge computing cost makes video understanding research only accessible to big companies/labs with abundant computing resources. There are several recent attempts to speed up the training of deep video models \cite{wu_CVPR2020_multigrid}, but these are still expensive compared to most image-based computer vision tasks.
In terms of deployment, 3D convolution is not as well supported as 2D convolution for different platforms. Furthermore, 3D CNNs require more video frames as input which adds additional IO cost. 

Hence, starting from year 2018, researchers started to investigate other alternatives to see how they could improve accuracy and efficiency at the same time for video action recognition. We will review recent efficient video modeling methods in several categories below.

\subsubsection{Flow-mimic approaches}
\label{subsubsec:flow_mimic}
One of the major drawback of two-stream networks is its need for optical flow. Pre-computing optical flow is computationally expensive, storage demanding, and not end-to-end trainable for video action recognition. It is appealing if we can find a way to encode motion information without using optical flow, at least during inference time.

\cite{ng_WACV2018_actionflownet} and \cite{diba_arxiv2016_efficient3DTS} are early attempts for learning to estimate optical flow inside a network for video action recognition. Although these two approaches do not need optical flow during inference, they require optical flow during training in order to train the flow estimation network. Hidden two-stream networks \cite{zhu_ACCV2018_hidden} proposed MotionNet to replace the traditional optical flow computation. MotionNet is a lightweight network to learn motion information in an unsupervised manner, and when concatenated with the temporal stream, is end-to-end trainable. Thus, hidden two-stream CNNs \cite{zhu_ACCV2018_hidden} only take raw video frames as input and directly predict action classes without explicitly computing optical flow, regardless of whether its the training or inference stage.
PAN~\cite{zhang2020pan} mimics the optical flow features by computing the difference between consecutive feature maps. Following this direction, \cite{sun_CVPR2018_OFF,fan_CVPR2018_end2end,lee_ECCV2018_motionFilter,piergiovanni_CVPR2019_repFlow} continue to investigate end-to-end trainable CNNs to learn optical-flow-like features from data. They derive such features directly from the definition of optical flow \cite{zach_PRS2007_TVL1}. 
MARS \cite{crasto_CVPR2019_MARS} and D3D \cite{stroud_WACV2020_D3D} used knowledge distillation to combine two-stream networks into a single stream, e.g., by tuning the spatial stream to predict the outputs of the temporal stream. 
Recently, Kwon \etal ~\cite{kwon2020motionsqueeze} introduced MotionSqueeze module to estimate the motion features. The proposed module is end-to-end trainable and can be plugged into any network, similar to \cite{zhu_ACCV2018_hidden}.



\subsubsection{Temporal modeling without 3D convolution}
\label{subsubsec:no3d}
A simple and natural choice to model temporal relationship between frames is using 3D convolution. However, there are many alternatives to achieve this goal. Here, we will review some recent work that performs temporal modeling without 3D convolution. 

Lin \etal \cite{lin_ICCV2019_TSM} introduce a new method, termed temporal shift module (TSM). TSM extends the shift operation \cite{wu_CVPR2018_shift} to video understanding. It shifts part of the channels along the temporal dimension, thus facilitating information exchanged among neighboring frames. In order to keep spatial feature learning capacity, they put temporal shift module inside the residual branch in a residual block. Thus all the information in the original activation is still accessible after temporal shift through identity mapping. The biggest advantage of TSM is that it can be inserted into a 2D CNN to achieve temporal modeling at zero computation and zero parameters.
Similar to TSM, TIN~\cite{shao2020temporal} introduces a temporal interlacing module to model the temporal convolution.

There are several recent 2D CNNs approaches using attention to perform long-term temporal modeling \cite{jiang_ICCV2019_STM,li_CVPR2020_TEA,liu_AAAI2020_TEINet,liu_arxiv2020_TAM}. STM \cite{jiang_ICCV2019_STM} proposes a channel-wise spatio-temporal module to present the spatio-temporal features and a channel-wise motion
module to efficiently encode motion features. TEA \cite{li_CVPR2020_TEA} is similar to STM, but inspired by SENet \cite{hu_CVPR2018_senet}, TEA uses motion features to recalibrate the spatio-temporal features to enhance the motion pattern. Specifically, TEA has two components: motion excitation and multiple temporal aggregation, while the first one handles short-range motion modeling and the second one efficiently enlarge the temporal receptive field for long-range temporal modeling. They are complementary and both light-weight, thus TEA is able to achieve competitive results with previous best approaches while keeping FLOPs as low as many 2D CNNs. Recently, TEINet \cite{liu_AAAI2020_TEINet} also adopts attention to enhance temporal modeling. Note that, the above attention-based methods are different from non-local \cite{wang_CVPR2018_nonlocal}, because they use channel attention while non-local uses spatial attention.

\subsection{Miscellaneous}
\label{subsec:misc}
In this section, we are going to show several other directions that are popular for video action recognition in the last decade. 


\begin{table*}[t]
\begin{center}
\begin{tabular}{|l|c|c|c|c|c|c|c|}
\hline
Method & Pre-train & Flow & Backbone & Venue & UCF101 & HMDB51 & Kinetics400 \\
\hline\hline
DeepVideo \cite{karpathy_CVPR2014_videoCNN} & I & - &  AlexNet & CVPR 2014 & 65.4 & - & -  \\
Two-stream \cite{simonyan_NIPS2014_twoStream} & I & \checkmark &  CNN-M & NeurIPS 2014 & 88.0 & 59.4 & -  \\
LRCN \cite{donahue_CVPR2015_LRCN} & I & \checkmark &  CNN-M & CVPR 2015 & 82.3 & - & -  \\
TDD \cite{want_CVPR2015_trajectory} & I & \checkmark &  CNN-M & CVPR 2015 & 90.3 & 63.2 & - \\
Fusion \cite{feichtenhofer_CVPR2016_fusion} & I & \checkmark &  VGG16 & CVPR 2016 & 92.5 & 65.4 & - \\
TSN \cite{wang_ECCV2016_TSN} & I & \checkmark &  BN-Inception & ECCV 2016 & 94.0 & 68.5 & \textbf{73.9} \\
TLE \cite{diba_CVPR2017_TLE} & I & \checkmark &  BN-Inception & CVPR 2017 & \textbf{95.6} & \textbf{71.1} & - \\
\hline\hline
C3D \cite{tran_ICCV2015_C3D} & S & - & VGG16-like & ICCV 2015 & 82.3 & 56.8 & 59.5 \\
I3D \cite{carreira_CVPR2017_I3D} & I,K & - & BN-Inception-like & CVPR 2017 & 95.6 & 74.8 & 71.1 \\
P3D \cite{qiu_ICCV2017_P3D} & S & - & ResNet50-like & ICCV 2017 & 88.6 & - & 71.6 \\
ResNet3D \cite{hara_CVPR2018_3DResNet} & K & - &  ResNeXt101-like & CVPR 2018 & 94.5 & 70.2 & 65.1 \\
R2+1D \cite{tran_CVPR2018_R2plus1D} & K & - & ResNet34-like & CVPR 2018 & 96.8 & 74.5 & 72.0 \\
NL I3D \cite{wang_CVPR2018_nonlocal} & I & - & ResNet101-like & CVPR 2018 & - & - & 77.7 \\
S3D \cite{xie_ECCV2018_S3D} & I,K & - & BN-Inception-like &  ECCV 2018 & 96.8 & \textbf{75.9} & 74.7 \\
SlowFast \cite{feichtenhofer_ICCV2019_slowfast} & - & - & ResNet101-NL-like & ICCV 2019 & - & - & 79.8 \\
X3D-XXL \cite{feichtenhofer_CVPR2020_X3D} & - & - & ResNet-like & CVPR 2020 & - & - & \textbf{80.4} \\
TPN \cite{yang_CVPR2020_TPN} & - & - & ResNet101-like & CVPR 2020 & - & - & 78.9 \\
CIDC \cite{li2020directional} & - & - & ResNet50-like & ECCV 2020 & \textbf{97.9} & 75.2 & 75.5 \\
\hline\hline
Hidden TSN \cite{zhu_ACCV2018_hidden} & I &  -& BN-Inception & ACCV 2018 & 93.2 & 66.8 & 72.8 \\
OFF \cite{sun_CVPR2018_OFF} & I & - & BN-Inception & CVPR 2018 & 96.0 & 74.2 & - \\
TSM \cite{lin_ICCV2019_TSM} & I & - & ResNet50 & ICCV 2019 & 95.9 & 73.5 & 74.1 \\
STM \cite{jiang_ICCV2019_STM} & I,K & - & ResNet50-like & ICCV 2019 & 96.2 & 72.2 & 73.7 \\
TEINet \cite{liu_AAAI2020_TEINet} & I,K & - & ResNet50-like & AAAI 2020 & 96.7 & 72.1 & 76.2 \\
TEA \cite{li_CVPR2020_TEA} & I,K & - & ResNet50-like & CVPR 2020 & \textbf{96.9} & 73.3 & 76.1 \\
MSNet \cite{kwon2020motionsqueeze} & I,K & - & ResNet50-like & ECCV 2020 & - & \textbf{77.4} & \textbf{76.4} \\
\hline
\end{tabular}
\end{center}
\caption{\textbf{Results of widely adopted methods on three scene-focused datasets}. Pre-train indicates which dataset the model is pre-trained on. I: ImageNet, S: Sports1M and K: Kinetics400. NL represents non local.}
\label{tab:scene_results}
\end{table*}

\subsubsection{Trajectory-based methods}
\label{subsubsec:both_worlds}

While CNN-based approaches have demonstrated their superiority and gradually replaced the traditional hand-crafted methods, the traditional local feature pipeline still has its merits which should not be ignored, such as the usage of trajectory. 

Inspired by the good performance of trajectory-based methods \cite{wang_ICCV2013_IDT}, Wang \etal ~\cite{want_CVPR2015_trajectory} proposed to conduct trajectory-constrained pooling to aggregate deep convolutional features into effective descriptors, which they term as TDD. Here, a trajectory is defined as a path tracking down pixels in the temporal dimension. This new video representation shares the merits of both hand-crafted features and deep-learned features, and became one of the top performers on both UCF101 and HMDB51 datasets in the year 2015. Concurrently, Lan \etal \cite{lan_arxiv2015_bothWorlds} incorporated both Independent Subspace Analysis (ISA) and dense trajectories into the standard two-stream networks, and show the complementarity between data-independent and data-driven approaches. Instead of treating CNNs as a fixed feature extractor, Zhao \etal \cite{zhao_NIPS2018_trajconv} proposed trajectory convolution to learn features along the temporal dimension with the help of trajectories.

\subsubsection{Rank pooling}
\label{subsubsec:rank_pooling}
There is another way to model temporal information inside a video, termed rank pooling (a.k.a learning-to-rank). The seminal work in this line starts from VideoDarwin \cite{fernando_CVPR2015_videoDarwin}, that uses a ranking machine to learn the evolution of the appearance over time and returns a ranking function. The ranking function should be able to order the frames of a video temporally, thus they use the parameters of this ranking function as a new video representation. VideoDarwin \cite{fernando_CVPR2015_videoDarwin} is not a deep learning based method, but achieves comparable performance and efficiency.

To adapt rank pooling to deep learning, Fernando \cite{fernando_ICML2016_e2eRankPool} introduces a differentiable rank pooling layer to achieve end-to-end feature learning. Following this direction, Bilen \etal \cite{bilen_CVPR2016_dynamicImage} apply rank pooling on the raw image pixels of a video producing a single RGB image per video, termed dynamic images. Another concurrent work by Fernando \cite{fernando_CVPR2016_HrankPool} extends rank pooling to hierarchical rank pooling by stacking multiple levels of temporal encoding. Finally, \cite{cherian_CVPR2017_gRankPool} propose a generalization of the original ranking formulation \cite{fernando_CVPR2015_videoDarwin} using subspace representations and show that it leads to significantly better representation of the dynamic evolution of actions, while being computationally cheap.

\subsubsection{Compressed video action recognition}
\label{subsubsec:compressed}
Most video action recognition approaches use raw videos (or decoded video frames) as input. However, there are several drawbacks of using raw videos, such as the huge amount of data and high temporal redundancy. Video compression methods usually store one frame by reusing contents from another frame (i.e., I-frame) and only store the difference (i.e., P-frames and B-frames) due to the fact that adjacent frames are similar. Here, the I-frame is the original RGB video frame, and P-frames and B-frames include the motion vector and residual, which are used to store the difference. Motivated by the developments in the video compression domain, researchers started to adopt compressed video representations as input to train effective video models. 

Since the motion vector has coarse structure and may contain inaccurate movements, Zhang \etal \cite{zhang_CVPR2016_motionVec} adopted knowledge distillation to help the motion-vector-based temporal stream mimic the optical-flow-based temporal stream. However, their approach required extracting and processing each frame. They obtained comparable recognition accuracy with standard two-stream networks, but were 27 times faster. Wu \etal \cite{wu_CVPR2018_compressed} used a heavyweight CNN for the I frame and lightweight CNN's for the P frames. This required that the motion vectors and residuals for each P frame be referred back to the I frame by accumulation. DMC-Net \cite{shou_CVPR2019_DMCNet} is a follow-up work to  \cite{wu_CVPR2018_compressed} using adversarial loss. It adopts a lightweight generator network to help the motion vector capturing fine motion details, instead of knowledge distillation as in \cite{zhang_CVPR2016_motionVec}. A recent paper SCSampler \cite{korbar_ICCV2019_scsampler}, also adopts compressed video representation for sampling salient clips and we will discuss it in the next section~\ref{subsubsec:sampling}. As yet none of the compressed approaches can deal with B-frames due to the added complexity.

\subsubsection{Frame/Clip sampling}
\label{subsubsec:sampling}

Most of the aforementioned deep learning methods  treat every video frame/clip equally for the final prediction. However, discriminative actions only happen in a few moments, and most of the other video content is irrelevant or weakly related to the labeled action category. There are several drawbacks of this paradigm. First, training with a large proportion of irrelevant video frames may hurt performance. Second, such uniform sampling is not efficient during inference. 

Partially inspired by how human  understand a video using just a few glimpses over the entire video~\cite{yeung_CVPR2016_glimpses}, many methods were proposed to sample the most informative video frames/clips for both improving the performance and making the model more efficient during inference. 

KVM~\cite{zhu_CVPR2016_keyVolume} is one of the first attempts to propose an end-to-end framework to simultaneously identify key volumes and do action classification. 
Later, \cite{kar_CVPR2017_adascan} introduce AdaScan that predicts the importance score of each video frame in an online fashion, which they term as adaptive temporal pooling.
Both of these methods achieve improved performance, but they still adopt the standard evaluation scheme which does not show efficiency during inference.
Recent approaches focus more on the efficiency~\cite{fan2018watching,wu_CVPR2019_adaframe,bhardwaj_CVPR2019_fewer,korbar_ICCV2019_scsampler}.
AdaFrame~\cite{wu_CVPR2019_adaframe} follows \cite{yeung_CVPR2016_glimpses,kar_CVPR2017_adascan} but uses a reinforcement learning based approach to search more informative video clips. 
Concurrently, \cite{bhardwaj_CVPR2019_fewer} uses a teacher-student framework, i.e., a see-it-all teacher can be used to train a compute efficient see-very-little student. They demonstrate that the efficient student network can reduce the inference time by $30\%$ and the number of FLOPs by approximately $90\%$ with negligible performance drop.
Recently, SCSampler~\cite{korbar_ICCV2019_scsampler} trains a lightweight network to sample the most salient video clips based on compressed video representations, and achieve state-of-the-art performance on both Kinetics400 and Sports1M dataset. They also empirically show that such saliency-based sampling is not only efficient, but also enjoys higher accuracy than using all the video frames.





\begin{table*}[t]
\label{tab:sthsth}
\begin{center}
\begin{tabular}{|l|c|c|c|c|c|c|c|c|}
\hline
Method & Pre-train & Backbone  & Frames$\times$Views & Venue &V1 Top1  &V2 Top1 \\
\hline\hline
TSN \cite{wang_ECCV2016_TSN} & I & BN-Inception & 8$\times$1 & ECCV 2016 & 19.7 & -  \\
I3D \cite{carreira_CVPR2017_I3D} & I,K  & ResNet50-like & 32$\times$6 &CVPR 2017& 41.6  & -  \\
NL I3D \cite{wang_CVPR2018_nonlocal} & I,K & ResNet50-like & 32$\times$6 &CVPR 2018& 44.4 & -  \\
NL I3D + GCN \cite{wang_ECCV2018_videoGraph} & I,K & ResNet50-like & 32$\times$6 & ECCV 2018 & 46.1  & -  \\ 
ECO \cite{zolfaghari_ECCV2018_ECO} & K & BNIncep+ResNet18 & 16$\times$1 & ECCV 2018 & 41.4  & -  \\
TRN \cite{zhou_ECCV2018_TRN} & I & BN-Inception & 8$\times$1 &ECCV 2018 &  42.0 &  48.8  \\ 
STM \cite{jiang_ICCV2019_STM} & I & ResNet50-like & 8$\times$30 & ICCV 2019& 49.2 & -  \\ 
STM \cite{jiang_ICCV2019_STM} & I & ResNet50-like & 16$\times$30 & ICCV 2019 & 50.7 & -  \\
TSM \cite{lin_ICCV2019_TSM} & K & ResNet50 & 8$\times$1 &ICCV 2019 & 45.6 &  59.1 \\
TSM \cite{lin_ICCV2019_TSM} & K & ResNet50 & 16$\times$1 &ICCV 2019 & 47.2 &  63.4 \\
bLVNet-TAM \cite{fan_NIPS2019_bLVNet} & I & BLNet-like & 8$\times$2 &NeurIPS  2019& 46.4 & 59.1  \\
bLVNet-TAM \cite{fan_NIPS2019_bLVNet} & I & BLNet-like & 16$\times$2 &NeurIPS 2019 &48.4  & 61.7 \\
TEA \cite{li_CVPR2020_TEA} & I & ResNet50-like & 8$\times$1 &  CVPR 2020 &48.9 &  - \\
TEA \cite{li_CVPR2020_TEA} & I & ResNet50-like  & 16$\times$1 & CVPR 2020& 51.9 &  -  \\
TSM + TPN \cite{yang_CVPR2020_TPN} & K & ResNet50-like & 8$\times$1 &CVPR 2020 & 49.0  & 62.0  \\
MSNet \cite{kwon2020motionsqueeze} & I & ResNet50-like & 8$\times$1 &ECCV 2020 & 50.9  & 63.0  \\
MSNet \cite{kwon2020motionsqueeze} & I & ResNet50-like &  16$\times$1 &ECCV 2020 & \textbf{52.1}  & \textbf{64.7} \\
TIN \cite{shao2020temporal} & K & ResNet50-like & 16$\times$1 & AAAI 2020 &47.0 &  60.1 \\
TEINet \cite{liu_AAAI2020_TEINet} & I & ResNet50-like &  8$\times$1 & AAAI 2020 & 47.4  & 61.3  \\
TEINet \cite{liu_AAAI2020_TEINet} & I & ResNet50-like &  16$\times$1 & AAAI 2020 & 49.9  & 62.1  \\
\hline
\end{tabular}
\end{center}
\caption{\textbf{Results of widely adopted methods on Something-Something V1 and V2 datasets}. We only report numbers without using optical flow. Pre-train indicates which dataset the model is pre-trained on. I: ImageNet and K: Kinetics400. View means number of temporal clip multiples spatial crop, e.g., 30 means 10 temporal clips with 3 spatial crops each clip.}
\end{table*}

\subsubsection{Visual tempo}
\label{subsubsec:tempo}

Visual tempo is a concept to describe how fast an action goes. Many action classes have different visual tempos. In most cases, the key to distinguish them is their visual tempos, as they might share high similarities in visual appearance, such as walking, jogging and running \cite{yang_CVPR2020_TPN}. 
There are several papers exploring different temporal rates (tempos) for improved temporal modeling \cite{zhu_CVPR2017_multirate,ng_WACV2018_TDN,huang_CVPR2018_makeVideo,zhu_ACCV2018_RTS,feichtenhofer_ICCV2019_slowfast,yang_CVPR2020_TPN}.
Initial attempts usually capture the video tempo through sampling raw videos at multiple rates and constructing an input-level frame pyramid \cite{zhu_CVPR2017_multirate,ng_WACV2018_TDN,zhu_ACCV2018_RTS}.
Recently, SlowFast \cite{feichtenhofer_ICCV2019_slowfast}, as we discussed in section~\ref{subsubsec:3d_efficiency}, utilizes the characteristics of visual tempo to design a two-pathway network for better accuracy and efficiency trade-off.
CIDC \cite{li2020directional} proposed  directional temporal modeling along with a local backbone for video temporal modeling.
TPN \cite{yang_CVPR2020_TPN} extends the tempo modeling to the feature-level and shows consistent improvement over previous approaches.

We would like to point out that visual tempo is also widely used in self-supervised video representation learning~\cite{benaim_CVPR2020_speednet,yang_arxiv2020_vthcl, chen2020rspnet} since it can naturally provide supervision signals to train a deep network. We will discuss more details on self-supervised video representation learning in section~\ref{subsec:selfsup}.






\section{Evaluation and Benchmarking}
\label{sec:benchmark}

In this section, we compare popular approaches on benchmark datasets. To be specific, we first introduce standard evaluation schemes in section \ref{subsec:evaluation}. Then we divide common benchmarks into three categories, scene-focused (UCF101, HMDB51 and Kinetics400 in section \ref{subsec:scene_datasets}), motion-focused (Sth-Sth V1 and V2 in section \ref{subsec:temporal_datasets}) and multi-label (Charades in section \ref{subsec:multilabel_datasets}). In the end, we present a fair comparison among popular methods in terms of both recognition accuracy and efficiency in section \ref{subsec:speed}.

\subsection{Evaluation scheme}
\label{subsec:evaluation}
During model training, we usually randomly pick a video frame/clip to form mini-batch samples. However, for evaluation, we need a standardized pipeline in order to perform fair comparisons.

For 2D CNNs, a widely adopted evaluation scheme is to evenly sample $25$ frames from each video following \cite{simonyan_NIPS2014_twoStream,wang_arxiv2015_good}. For each frame, we perform ten-crop data augmentation by cropping the 4 corners and 1 center, flipping them horizontally and averaging the prediction scores (before softmax operation) over all crops of the samples, i.e., this means we use 250 frames per video for inference.

For 3D CNNs, a widely adopted evaluation scheme termed 30-view strategy is to evenly sample $10$ clips from each video following \cite{wang_CVPR2018_nonlocal}. For each video clip, we perform three-crop data augmentation. To be specific, we scale the shorter spatial side to $256$ pixels and take three crops of $256 \times 256$ to cover the spatial dimensions and average the prediction scores. 

However, the evaluation schemes are not fixed. They are evolving and adapting to new network architectures and different datasets. For example, TSM \cite{lin_ICCV2019_TSM} only uses two clips per video for small-sized datasets \cite{ucf101,hmdb51}, and perform three-crop data augmentation for each clip despite its being a 2D CNN. We will mention any deviations from the standard evaluation pipeline.

In terms of evaluation metric, we report \textit{accuracy} for single-label action recognition, and \textit{mAP (mean average precision)} for multi-label action recognition.

\subsection{Scene-focused datasets}
\label{subsec:scene_datasets}
Here, we compare recent state-of-the-art approaches on scene-focused datasets: UCF101, HMDB51 and Kinetics400. The reason we call them scene-focused is because most action videos in these datasets are short, and can be recognized by static scene appearance alone as shown in Figure \ref{fig:dataset_comparison}.

Following the chronology, we first present results for early attempts of using deep learning and the two-stream networks at the top of Table \ref{tab:scene_results}. We make several observations. First, without motion/temporal modeling, the performance of DeepVideo \cite{karpathy_CVPR2014_videoCNN} is inferior to all other approaches. Second, it is helpful to transfer knowledge from traditional methods (non-CNN-based) to deep learning. For example, TDD \cite{want_CVPR2015_trajectory} uses trajectory pooling to extract motion-aware CNN features. TLE \cite{diba_CVPR2017_TLE} embeds global feature encoding, which is an important step in traditional video action recognition pipeline, into a deep network. 

We then compare 3D CNNs based approaches in the middle of Table \ref{tab:scene_results}. Despite training on a large corpus of videos, C3D \cite{tran_ICCV2015_C3D} performs inferior to concurrent work \cite{simonyan_NIPS2014_twoStream,want_CVPR2015_trajectory,wang_arxiv2015_good}, possibly due to difficulties in optimization of 3D kernels. Motivated by this, several papers - I3D \cite{carreira_CVPR2017_I3D}, P3D \cite{qiu_ICCV2017_P3D}, R2+1D \cite{tran_CVPR2018_R2plus1D} and S3D \cite{xie_ECCV2018_S3D} factorize 3D convolution filters to 2D spatial kernels and 1D temporal kernels to ease the training. In addition, I3D introduces an inflation strategy to avoid training from scratch by bootstrapping the 3D model weights from well-trained 2D networks. By using these techniques, they achieve comparable performance to the best two-stream network methods \cite{diba_CVPR2017_TLE} without the need for optical flow.
Furthermore, recent 3D models obtain even higher accuracy, by using more training samples \cite{tran_ICCV2019_CSN}, additional pathways \cite{feichtenhofer_ICCV2019_slowfast}, or architecture search \cite{feichtenhofer_CVPR2020_X3D}.

Finally, we show recent efficient models in the bottom of Table \ref{tab:scene_results}. We can see that these methods are able to achieve higher recognition accuracy than two-stream networks (top), and comparable performance to 3D CNNs (middle). Since they are 2D CNNs and do not use optical flow, these methods are efficient in terms of both training and inference. Most of them are real-time approaches, and some can do online video action recognition \cite{lin_ICCV2019_TSM}. We believe 2D CNN plus temporal modeling is a promising direction due to the need of efficiency. Here, temporal modeling could be attention based, flow based or 3D kernel based.

\begin{table*}[t]
\begin{center}
\begin{tabular}{|l|c|c|c|c|c|}
\hline
Method & Extra-information & Backbone & Pre-train & Venue & mAP\\
\hline\hline
2D CNN \cite{charades} &-& AlexNet & I & ECCV 2016& 11.2\\
Two-stream \cite{charades} &flow& VGG16 & I & ECCV 2016& 22.4\\
ActionVLAD \cite{girdhar_CVPR2017_actionVLAD} &-& VGG16 & I & CVPR 2017& 21.0\\
CoViAR \cite{wu_CVPR2018_compressed} &-& ResNet50-like & - &CVPR 2018& 21.9 \\
MultiScale TRN \cite{zhou_ECCV2018_TRN} &-& BN-Inception-like & I  &ECCV 2018& \textbf{25.2}\\
\hline\hline
I3D \cite{carreira_CVPR2017_I3D} &-& BN-Inception-like &K400&CVPR 2017& 32.9\\
STRG \cite{wang_ECCV2018_videoGraph} &-& ResNet101-NL-like&K400 &ECCV 2018& 39.7\\
LFB \cite{wu_CVPR2019_featureBank} &-& ResNet101-NL-like&K400 &CVPR 2019& 42.5\\
TC \cite{hussein_ICCV2019_timeception} && ResNet101-NL-like&K400 &ICCV 2019& 41.1\\
HAF \cite{wang_ICCV2019_hallucinating} &IDT + flow& BN-Inception-like &K400 &ICCV 2019& 43.1\\
SlowFast \cite{feichtenhofer_ICCV2019_slowfast} &-& ResNet-like & K400 &ICCV 2019& 42.5\\
SlowFast \cite{feichtenhofer_ICCV2019_slowfast} &-& ResNet-like & K600 &ICCV 2019& \textbf{45.2}\\
\hline\hline
Action-Genome \cite{ji2020action} &person + object& ResNet-like &  - &CVPR 2020& \textbf{60.1}\\
AssembleNet++ \cite{ryoo2020assemblenet++} &flow + object& ResNet-like & - &ECCV 2020& 59.9\\
\hline
\end{tabular}
\end{center}
\caption{\textbf{Charades evaluation using mAP}, calculated using the officially provided script. NL: non-local network. Pre-train indicates which dataset the model is pre-trained on. I: ImageNet, K400: Kinetics400 and K600: Kinetics600.}
\label{tab:charades}
\end{table*}

\subsection{Motion-focused datasets}
\label{subsec:temporal_datasets}
In this section, we compare the recent state-of-the-art approaches on the 20BN-Something-Something (Sth-Sth) dataset. We report top1 accuracy on both V1 and V2. Sth-Sth datasets focus on humans performing basic actions with daily objects. Different from scene-focused datasets, background scene in Sth-Sth datasets contributes little to the final action class prediction.
In addition, there are classes such as ``Pushing something from left to right'' and ``Pushing something from right to left'', and which  require strong motion reasoning.

By comparing the previous work in Table 3, we observe that using longer input (e.g., 16 frames) is generally better. 
Moreover, methods that focus on temporal modeling \cite{lin_ICCV2019_TSM, li_CVPR2020_TEA, jiang_ICCV2019_STM} work  better than stacked 3D kernels \cite{carreira_CVPR2017_I3D}. 
For example, TSM \cite{lin_ICCV2019_TSM}, TEA \cite{li_CVPR2020_TEA} and MSNet \cite{kwon2020motionsqueeze} insert an explicit temporal reasoning module into 2D ResNet backbones and achieves state-of-the-art results. This implies that the Sth-Sth dataset needs strong temporal motion reasoning as well as spatial semantics information.

\subsection{Multi-label datasets}
\label{subsec:multilabel_datasets}
In this section, we first compare the recent state-of-the-art approaches on Charades dataset \cite{charades} and then we list some recent work that use assemble model or additional object information on Charades.

Comparing the previous work in Table~\ref{tab:charades}, we make the following observations. 
First, 3D models \cite{wu_CVPR2019_featureBank,feichtenhofer_ICCV2019_slowfast} generally perform better than 2D models \cite{charades,wu_CVPR2018_compressed} and 2D models with optical flow input. This indicates that the spatio-temporal reasoning is critical for long-term complex concurrent action understanding. 
Second, longer input helps with the recognition \cite{wu_CVPR2019_featureBank} probably because some actions require long-term feature to recognize. 
Third, models with strong backbones that are pre-trained on larger datasets generally have better performance \cite{feichtenhofer_ICCV2019_slowfast}. This is because Charades is a medium-scaled dataset which doesn't contain enough diversity to train a deep model. 

Recently, researchers explored the alternative direction for complex concurrent action recognition by assembling models \cite{ryoo2020assemblenet++} or providing additional human-object interaction information \cite{ji2020action}. These papers significantly outperformed previous literature that only finetune a single model on Charades. It demonstrates that exploring spatio-temporal human-object interactions and finding a way to avoid overfitting are the keys for concurrent action understanding. 

\begin{table*}
\begin{center}
\scalebox{0.95}{
\begin{tabular}{|l|c|c|c|c|c|c|}
\hline
Model & Input  & FLOPS (G) & $\#$ of params (M) & FPS  & Latency (s) & Acc ($\%$) \\
\hline\hline
TSN-ResNet18 \cite{wang_ECCV2016_TSN} & $3\times224\times224$ & 3.671          & 21.49      & 151.96              & 0.0066          & 69.85     \\
TSN-ResNet34 \cite{wang_ECCV2016_TSN} & $3\times224\times224$  & 1.819          & 11.382     & 264.01              & 0.0038          & 66.73  \\
TSN-ResNet50 \cite{wang_ECCV2016_TSN}  & $3\times224\times224$ & 4.110          & 24.328     & 114.05              & 0.0088          & 70.88    \\
TSN-ResNet101 \cite{wang_ECCV2016_TSN}  & $3\times224\times224$ & 7.833          & 43.320     & 59.56               & 0.0167          & 72.25     \\
TSN-ResNet152 \cite{wang_ECCV2016_TSN}  & $3\times224\times224$ & 11.558         & 58.963     & 36.93               & 0.0271          & 72.45      \\
\hline\hline
I3D-ResNet50 \cite{carreira_CVPR2017_I3D}  & $3\times32\times224\times224$ & 33.275         & 28.863     & 1719.50             & 0.0372          & 74.87        \\
I3D-ResNet101 \cite{carreira_CVPR2017_I3D} & $3\times32\times224\times224$ & 51.864         & 52.574     & 1137.74             & 0.0563          & 75.10       \\
I3D-ResNet50-NL \cite{wang_CVPR2018_nonlocal}  & $3\times32\times224\times224$ & 47.737         & 38.069     & 1403.16             & 0.0456          & 75.17          \\
I3D-ResNet101-NL \cite{wang_CVPR2018_nonlocal} & $3\times32\times224\times224$
       & 66.326         & 61.780     & 999.94              & 0.0640          & 75.81     \\
R2+1D-ResNet18 \cite{tran_CVPR2018_R2plus1D} & $3\times16\times112\times112$  & 40.645         & 31.505     & 804.31              & 0.0398          & 71.72       \\  
R2+1D-ResNet34 \cite{tran_CVPR2018_R2plus1D}  & $3\times16\times112\times112$  & 75.400         & 61.832     & 503.17              & 0.0636          & 72.63      \\    
R2+1D-ResNet50 \cite{tran_CVPR2018_R2plus1D}  & $3\times16\times112\times112$  & 65.543         & 53.950     & 667.06              & 0.0480          & 74.92          \\    
R2+1D-ResNet152$\ast$ \cite{tran_CVPR2018_R2plus1D} & $3\times32\times112\times112$  & 252.900        & 118.227    & 546.19              & 0.1172          & 81.34      \\    
CSN-ResNet152$\ast$ \cite{tran_ICCV2019_CSN}  & $3\times32\times224\times224$  & 74.758         & 29.704     & 435.77              & 0.1469          & 83.18      \\  
I3D-Slow-ResNet50 \cite{feichtenhofer_ICCV2019_slowfast} & $3\times8\times224\times224$  & 41.919         & 32.454     & 1702.60             & 0.0376          & 74.41       \\
I3D-Slow-ResNet50 \cite{feichtenhofer_ICCV2019_slowfast}  & $3\times16\times224\times224$  & 83.838         & 32.454     & 1406.00             & 0.0455          & 76.36     \\
I3D-Slow-ResNet50 \cite{feichtenhofer_ICCV2019_slowfast} & $3\times32\times224\times224$ & 167.675        & 32.454     & 860.74              & 0.0744          & 77.89      \\
I3D-Slow-ResNet101 \cite{feichtenhofer_ICCV2019_slowfast}  & $3\times8\times224\times224$  & 85.675         & 60.359     & 1114.22             & 0.0574          & 76.15    \\
I3D-Slow-ResNet101 \cite{feichtenhofer_ICCV2019_slowfast}  & $3\times16\times224\times224$ & 171.348        & 60.359     & 876.20              & 0.0730          & 77.11          \\
I3D-Slow-ResNet101 \cite{feichtenhofer_ICCV2019_slowfast}  & $3\times32\times224\times224$ & 342.696        & 60.359     & 541.16              & 0.1183          & 78.57        \\
SlowFast-ResNet50-4x16 \cite{feichtenhofer_ICCV2019_slowfast}  & $3\times32\times224\times224$ & 27.820         & 34.480     & 1396.45             & 0.0458          & 75.25      \\
SlowFast-ResNet50-8x8 \cite{feichtenhofer_ICCV2019_slowfast}   & $3\times32\times224\times224$ & 50.583         & 34.566     & 1297.24             & 0.0493          & 76.66       \\
SlowFast-ResNet101-8x8 \cite{feichtenhofer_ICCV2019_slowfast}  & $3\times32\times224\times224$ & 96.794         & 62.827     & 889.62              & 0.0719          & 76.95          \\
TPN-ResNet50 \cite{yang_CVPR2020_TPN}   & $3\times8\times224\times224$  & 50.457         & 71.800     & 1350.39             & 0.0474          & 77.04             \\
TPN-ResNet50 \cite{yang_CVPR2020_TPN}  & $3\times16\times224\times224$ & 99.929         & 71.800     & 1128.39             & 0.0567          & 77.33      \\
TPN-ResNet50 \cite{yang_CVPR2020_TPN}   & $3\times32\times224\times224$  & 198.874        & 71.800     & 716.89              & 0.0893          & 78.90         \\
TPN-ResNet101 \cite{yang_CVPR2020_TPN}  & $3\times8\times224\times224$  & 94.366         & 99.705     & 942.61              & 0.0679          & 78.10      \\
TPN-ResNet101\cite{yang_CVPR2020_TPN}  & $3\times16\times224\times224$ & 187.594        & 99.705     & 754.00              & 0.0849          & 79.39      \\
TPN-ResNet101\cite{yang_CVPR2020_TPN}   & $3\times32\times224\times224$ & 374.048        & 99.705     & 479.77              & 0.1334          & 79.70       \\
\hline
\end{tabular}
}
\end{center}
\caption{\textbf{Comparison on both efficiency and accuracy}. Top: 2D models and bottom: 3D models. FLOPS means floating point operations per second. FPS indicates how many video frames can the model process per second. Latency is the actual running time to complete one network forward given the input. Acc is the top-1 accuracy on Kinetics400 dataset. TSN, I3D, I3D-slow families are pretrained on ImageNet. R2+1D, SlowFast and TPN families are trained from scratch.}
\label{tab:speed}
\end{table*}

\subsection{Speed comparison}
\label{subsec:speed}
To deploy a model in real-life applications, we usually need to know whether it meets the speed requirement before we can proceed. In this section, we evaluate the approaches mentioned above to perform a thorough comparison in terms of (1) number of parameters, (2) FLOPS, (3) latency and (4) frame per second. 

We present the results in Table \ref{tab:speed}. Here, we use the models in the GluonCV video action recognition model zoo\footnote{To reproduce the numbers in Table \ref{tab:speed}, please visit \url{https://github.com/dmlc/gluon-cv/blob/master/scripts/action-recognition/README.md}} since all these models are trained using the same data, same data augmentation strategy and under same 30-view evaluation scheme, thus fair comparison. All the timings are done on a single Tesla V100 GPU with 105 repeated runs, while the first 5 runs are ignored for warming up. We always use a batch size of 1. In terms of model input, we use the suggested settings in the original paper. 

As we can see in Table \ref{tab:speed}, if we compare latency, 2D models are much faster than all other 3D variants. This is probably why most real-world video applications still adopt frame-wise methods.
Secondly, as mentioned in \cite{radosavovic_CVPR2020_RegNet,zhang2020resnest}, FLOPS is not strongly correlated with the actual inference time (i.e., latency). Third, if comparing performance, most 3D models give similar accuracy around $75\%$, but pre-training with a larger dataset can significantly boost the performance\footnote{Note that, R2+1D-ResNet152$\ast$ and CSN-ResNet152$\ast$ in Table \ref{tab:speed} are pretrained on a large-scale IG65M dataset~\cite{ghadiyaram_CVPR2019_IG65M}.}. This indicates the importance of training data and partially suggests that self-supervised pre-training might be a promising way to further improve existing methods.

\section{Discussion and Future Work}
\label{sec:future}
We have surveyed more than 200 deep learning based methods for video action recognition since year 2014. Despite the performance on benchmark datasets plateauing, there are many active and promising directions in this task worth exploring.

\subsection{Analysis and insights}
\label{subsec:insight}
More and more methods haven been developed to improve video action recognition, at the same time, there are some papers summarizing these methods and providing analysis and insights. Huang \etal \cite{huang_CVPR2018_makeVideo} perform an explicit analysis of the effect of temporal information for video understanding. They try to answer the question ``how important is the motion in the video for recognizing the action''. Feichtenhofer \etal \cite{feichtenhofer_CVPR2018_learned,feichtenhofer_IJCV2019_insights} provide an amazing visualization of what two-stream models have learned in order to understand how these deep representations work and what they are capturing. Li \etal \cite{li_ECCV2018_resound} introduce a concept, representation bias of a dataset, and find that current datasets are biased towards static representations. Experiments on such biased datasets may lead to erroneous conclusions, which is indeed a big problem that limits the development of video action recognition. Recently, Piergiovanni \etal introduced the AViD~\cite{piergiovanni2020avid} dataset to cope with data bias by collecting data from diverse groups of people. These papers provide great insights to help fellow researchers to understand the challenges, open problems and where the next breakthrough might reside. 

\subsection{Data augmentation}
\label{subsec:augmentation}
Numerous data augmentation methods have been proposed in image recognition domain, such as mixup \cite{zhang_ICLR2018_mixup}, cutout \cite{devries_arxiv2017_cutout}, CutMix \cite{yun_ICCV2019_cutmix}, AutoAugment \cite{cubuk_CVPR2019_autoaug}, FastAutoAug \cite{lim_NIPS2019_fastautoaug}, etc. However, video action recognition still adopts basic data augmentation techniques introduced before year 2015 \cite{wang_arxiv2015_good,simonyan_ICLR2015_VGG}, including random resizing, random cropping and random horizontal flipping. Recently, SimCLR \cite{chen_arxiv2020_simCLR} and other papers have shown that color jittering and random rotation greatly help representation learning. Hence, an investigation of using different data augmentation techniques for video action recognition is particularly useful. This may change the data pre-processing pipeline for all existing methods.

\subsection{Video domain adaptation}
\label{subsec:domain_adaptation}
Domain adaptation (DA) has been studied extensively in recent years to address the domain shift problem. Despite the accuracy on standard datasets getting higher and higher, the generalization capability of current video models across datasets or domains is less explored. 

There is early work on video domain adaptation \cite{sultani_CVPR2014_FWHD,xu_IVC2016_many2one,jamal_BMVC2018_DAspace,toby_CVPR2019_DDLSTM}. However, these literature focus on small-scale video DA with only a few overlapping categories, which may not reflect the actual domain discrepancy and may lead to biased conclusions. Chen \etal \cite{chen_ICCV2019_TAA} introduce two larger-scale datasets to investigate video DA and find that aligning temporal dynamics is particularly useful. Pan \etal \cite{pan_AAAI2020_coAttention} adopts co-attention to solve the temporal misalignment problem. Very recently, Munro \etal \cite{munro_CVPR2020_MMDA} explore a multi-modal self-supervision method for fine-grained video action recognition and show the effectiveness of multi-modality learning in video DA. Shuffle and Attend \cite{Jinwoo_shuffle_ECCV2020} argues that aligning features of all sampled clips results in a sub-optimal solution due to the fact that all clips do not include relevant semantics. Therefore, they propose to use an attention mechanism to focus more on informative clips and discard the non-informative ones. In conclusion, video DA is a promising direction, especially for researchers with less computing resources.

\subsection{Neural architecture search}
\label{subsec:nas}
Neural architecture search (NAS) has attracted great interest in recent years and is a promising research direction. However, given its greedy need for computing resources, only a few papers have been published in this area \cite{peng_ICIP2019_nas,piergiovanni_ICCV2019_evolve,piergiovanni_arxiv2019_TVN,ryoo_ICLR2020_assembleNet}. The TVN family \cite{piergiovanni_arxiv2019_TVN}, which jointly optimize parameters and runtime, can achieve competitive accuracy with human-designed contemporary models, and run much faster (within 37 to 100 ms on a CPU and 10 ms on a GPU per 1 second video clip).
AssembleNet \cite{ryoo_ICLR2020_assembleNet} and AssembleNet++ \cite{ryoo2020assemblenet++} provide a generic approach to learn the connectivity among feature representations across input modalities, and show surprisingly good performance on Charades and other benchmarks.
AttentionNAS~\cite{wang2020attentionnas} proposed a solution for spatio-temporal attention cell search. The found cell can be plugged into any network to improve the spatio-temporal features.
All previous papers do show their potential for video understanding.

Recently, some efficient ways of searching architectures have been proposed in the image recognition domain, such as DARTS \cite{liu_ICLR2019_darts}, Proxyless NAS \cite{cai_ICLR2019_proxylessnas}, ENAS \cite{pham_ICML2018_enas}, one-shot NAS \cite{gabriel_ICML2018_oneShot}, etc. It would be interesting to combine efficient 2D CNNs and efficient searching algorithms to perform video NAS for a reasonable cost.

\subsection{Efficient model development}
\label{subsec:realtime}
Despite their accuracy, it is difficult to deploy deep learning based methods for video understanding problems in terms of real-world applications. There are several major challenges: (1) most methods are developed in offline settings, which means the input is a short video clip, not a video stream in an online setting; (2) most methods do not meet the real-time requirement; (3) incompatibility of 3D convolutions or other non-standard operators on non-GPU devices (e.g., edge devices). 

Hence, the development of efficient network architecture based on 2D convolutions is a promising direction. The approaches proposed in the image classification domain can be easily adapted to video action recognition, e.g. model compression, model quantization, model pruning, distributed training \cite{goyal_arxiv2017_oneHour,lin_NIPSW2019_kinetics15min}, mobile networks \cite{howard_ICCV2019_mobilenetv3,zhang_CVPR2018_shufflenet}, mixed precision training, etc. However, more effort is needed for the online setting since the input to most action recognition applications is a video stream, such as surveillance monitoring. We may need a new and more comprehensive dataset for benchmarking online video action recognition methods. Lastly, using compressed videos might be desirable because most videos are already compressed, and we have free access to motion information.

\subsection{New datasets}
\label{subsec:new_datasets}
Data is more or at least as important as model development for machine learning. For video action recognition, most datasets are biased towards spatial representations ~\cite{li_ECCV2018_resound}, i.e., most actions can be recognized by a single frame inside the video without considering the temporal movement. Hence, a new dataset in terms of long-term temporal modeling is required to advance video understanding. Furthermore, most current datasets are collected from YouTube. Due to copyright/privacy issues, the dataset organizer often only releases the YouTube id or video link for users to download and not the actual video. The first problem is that downloading the large-scale datasets might be slow for some regions. In particular, YouTube recently started to block massive downloading from a single IP. Thus, many researchers may not even get the dataset to start working in this field. The second problem is, due to region limitation and privacy issues, some videos are not accessible anymore. For example, the original Kinetcis400 dataset has over 300K videos, but at this moment, we can only crawl about 280K videos. On average, we lose 5$\%$ of the videos every year. It is impossible to do fair comparisons between methods when they are trained and evaluated on different data.

\subsection{Video adversarial attack}
\label{subsec:adversarial}
Adversarial examples have been well studied on image models. \cite{szegedy_2013_intriguing} first shows that an adversarial sample, computed by inserting a small amount of noise on the original image, may lead to a wrong prediction. However, limited work has been done on attacking video models.

This task usually considers two settings, a white-box attack \cite{inkawhich2018adversarial,li2018adversarial,goodfellow2014explaining,chen2019appending} where the adversary can always get the full access to the model including exact gradients of a given input, or a black-box one \cite{jiang2019black,yan2020sparse,wei2019heuristic}, in which the structure and parameters of the model are blocked so that the attacker can only access the (input, output) pair through queries.
Recent work ME-Sampler \cite{zhang_ECCV2020_mesampler} leverages the motion information directly in generating adversarial videos, and is shown to successfully attack a number of video classification models using many fewer queries. 
In summary, this direction is useful since many companies provide APIs for services such as video classification, anomaly detection, shot detection, face detection, etc. In addition, this topic is also related to detecting DeepFake videos. Hence, investigating both attacking and defending methods is crucial to keeping these video services safe.

\subsection{Zero-shot action recognition}
\label{subsec:few_zero}
Zero-shot learning (ZSL) has been trending in the image understanding domain, and has recently been adapted to video action recognition. Its goal is to transfer the learned knowledge to classify previously unseen categories.
Due to (1) the expensive data sourcing and annotation and (2) the set of possible human actions is huge, zero-shot action recognition is a very useful task for real-world applications. 

There are many early attempts \cite{mtl_zsar_Xu_ECCV16,2015_ICCV_ZSL_detection,tran_zsar_Xu_IJCV17,spatial_aware_Mettes_ICCV17,ecoc_zsar_Qin_CVPR17,SIR_Gan_AAAI15} in this direction. Most of them follow a standard framework, which is to first extract visual features from videos using a pretrained network, and then train a joint model that maps the visual embedding to a semantic embedding space. In this manner, the model can be applied to new classes by finding the test class whose embedding is the nearest-neighbor of the model’s output. 
A recent work URL \cite{Zhu_CVPR2018_URL} proposes to learn a universal representation that generalizes across datasets.
Following URL \cite{Zhu_CVPR2018_URL}, \cite{Brattoli_CVPR2020_end2endZero} present the first end-to-end ZSL action recognition model. They also establish a new ZSL training and evaluation protocol, and provide an in-depth analysis to further advance this field. Inspired by the success of pre-training and then zero-shot in NLP domain, we believe ZSL action recognition is a promising research topic.

\subsection{Weakly-supervised video action recognition}
\label{subsec:weak_supervised}
Building a high-quality video action recognition dataset~\cite{ucf101,kinetics400} usually requires multiple laborious steps: 1) first sourcing a large amount of raw videos, typically from the internet; 2) removing videos irrelevant to the categories in the dataset; 3) manually trimming the video segments that have actions of interest; 4) refining the categorical labels. 
Weakly-supervised action recognition explores how to reduce the cost for curating training data.

The first direction of research~\cite{chen2015webly,ghadiyaram_CVPR2019_IG65M,gan2016webly} aims to reduce the cost of sourcing videos and accurate categorical labeling. They design training methods that use training data such as action-related images or partially annotated videos, gathered from publicly available sources such as Internet. Thus this paradigm is also referred to as webly-supervised learning~\cite{chen2015webly,gan2016webly}. Recent work on omni-supervised learning~\cite{ghadiyaram_CVPR2019_IG65M,girdhar2019distinit,duan2016omnisource} also follows this paradigm but features bootstrapping on unlabelled videos by distilling the models' own inference results.  

The second direction aims at removing trimming, the most time consuming part in annotation. UntrimmedNet~\cite{wang2017untrimmednets} proposed a method to learn action recognition model on untrimmed videos with only categorical labels~\cite{nguyen2018weakly,richard2017weakly}. This task is also related to weakly-supervised temporal action localization which aims to automatically generate the temporal span of the actions. Several papers propose to simultaneously~\cite{paul2018w} or iteratively~\cite{shou2018autoloc} learn models for these two tasks.

\subsection{Fine-grained video action recognition}
\label{subsec:fine_grained}
Popular action recognition datasets, such as UCF101~\cite{ucf101} or Kinetics400~\cite{kinetics400}, mostly comprise actions happening in various scenes. However, models learned on these datasets could overfit to contextual information irrelevant to the actions~\cite{wang2018pulling,weinzaepfel2019mimetics,choi2019can}. Several datasets have been proposed to study the problem of fine-grained action recognition, which could examine the models' capacities in modeling action specific information. These datasets comprise fine-grained actions in human activities such as cooking~\cite{Damen2020Collection,kuehne2014language,rohrbach15ijcv}, working~\cite{kobayashi2019fine} and sports~\cite{shao_CVPR2020_finegym,li_ECCV2018_resound}. 
For example, FineGym~\cite{shao_CVPR2020_finegym} is a recent large dataset annotated with different moves and sub-actions in gymnastic videos.

\subsection{Egocentric action recognition}
\label{subsec:ego-centric}
Recently, large-scale egocentric action recognition \cite{damen2020rescaling,Damen2020Collection} has attracted increasing interest with the emerging of wearable cameras devices.
Egocentric action recognition requires a fine understanding of hand motion and the interacting objects in the complex environment. 
A few papers leverage object detection features to offer fine object context to improve egocentric video recognition \cite{ma2016going,wang2020symbiotic,wu_CVPR2019_featureBank,sener2020temporal}.
Others  incorporate spatio-temporal attention \cite{Sudhakaran_2019_CVPR} or gaze annotations \cite{liu2019forecasting} to localize the interacting object to facilitate action recognition.
Similar to third-person action recognition, multi-modal inputs (e.g., optical flow and audio) have been demonstrated to be effective in egocentric action recognition \cite{Kazakos_2019_ICCV}.


\begin{table*}[t]
\small
	\begin{center}
		\begin{tabular}{l|l|ccc|cc|c|cc|cc} \hline
		 &  &  &  &  &  &  &   & \multicolumn{2}{c}{UCF101} & \multicolumn{2}{c}{HMDB51}  \\ 
		Method & Dataset & \rotatebox[origin=c]{45}{ Video} & \rotatebox[origin=c]{45}{ Audio} & \rotatebox[origin=c]{45}{ Text} & Size & Backbone & Venue  & Linear & FT & Linear & FT  \\\hline \hline
		AVTS~\cite{korbar2018cooperative}    & K400   & \checkmark & \checkmark &$-$   & $224$  & R(2+1)D-18   & NeurIPS 2018     &$-$& 86.2 &$-$& 52.3  \\
		AVTS~\cite{korbar2018cooperative}    & AS   & \checkmark & \checkmark &$-$   & $224$  & R(2+1)D-18   & NeurIPS 2018     &$-$& 89.1 &$-$& 58.1  \\
		CBT~\cite{sun2019learning}       & K600+	& \checkmark  & $-$  & \checkmark & $112$ & S3D & arXiv 2019 & 54.0 & 79.5 & 29.5 & 44.6\\
		MIL-NCE~\cite{miech20endtoend} & HTM   &   \checkmark   & $-$ & \checkmark    & $224$ & S3D   & CVPR 2020 & 82.7 & 91.3 & 53.1 & 61.0 \\
		ELO~\cite{pier20evolving}  & YT8M & \checkmark & \checkmark &$-$ & $224$   & R(2+1)D-50 &   CVPR 2020 & -- & 93.8 & 64.5 & 67.4\\ %
		XDC~\cite{alwassel2020selfsupervised}  & K400 	& \checkmark    & \checkmark  & $-$ & $224 $  & R(2+1)D-18 & NeurIPS 2020    &$-$& 86.8 &$-$& 52.6\\ 
		XDC~\cite{alwassel2020selfsupervised}  & AS 	& \checkmark    & \checkmark  & $-$ & $224 $  & R(2+1)D-18 & NeurIPS 2020    &$-$& 93.0 &$-$& 63.7\\
		XDC~\cite{alwassel2020selfsupervised}  & IG65M 	& \checkmark    & \checkmark  & $-$ & $224 $  & R(2+1)D-18 & NeurIPS 2020    &$-$& 94.6 &$-$& 66.5\\
		XDC~\cite{alwassel2020selfsupervised}  & IG-K 	& \checkmark    & \checkmark  & $-$ & $224 $  & R(2+1)D-18 & NeurIPS 2020    &$-$& \textbf{95.5} &$-$& 68.9\\
		AVID~\cite{morgado_avid_cma}    & AS   & \checkmark & \checkmark &$-$   & $224$  & R(2+1)D-50   & arXiv 2020     &$-$& 91.5 &$-$& 64.7  \\
		
		GDT~\cite{Patrick20}    & K400 & \checkmark & \checkmark &$-$   & $112$     & R(2+1)D-18 & arXiv 2020 &$-$& 89.3 &$-$& 60.0  \\ 
		GDT~\cite{Patrick20}    & AS & \checkmark & \checkmark &$-$   & $112$     & R(2+1)D-18 & arXiv 2020 &$-$& 92.5 &$-$& 66.1  \\ 
		GDT~\cite{Patrick20}    & IG65M & \checkmark & \checkmark &$-$   & $112$     & R(2+1)D-18 & arXiv 2020 &$-$& 95.2 &$-$& 72.8  \\ 
		MMV~\cite{alayrac2020selfsupervised}    & AS+ HTM & \checkmark & \checkmark & \checkmark    & $200$     & S3D & NeurIPS 2020 & 89.6 & 92.5 & 62.6 & 69.6  \\ 
		MMV~\cite{alayrac2020selfsupervised}    & AS+ HTM & \checkmark & \checkmark & \checkmark    & $200$     & TSM-50x2 & NeurIPS 2020 & \textbf{91.8} & 95.2 & \textbf{67.1} & \textbf{75.0}  \\ 
		\hline
		\hline
		OPN~\cite{OPN}        & UCF101       &  \checkmark &$-$&$-$  & $227$  & VGG & ICCV 2017    &$-$& 59.6 &$-$& 23.8  \\ 
		
		3D-RotNet~\cite{jing2018self} & K400       & \checkmark &$-$&$-$   & $112$      & R3D & arXiv 2018 &$-$& 62.9 &$-$& 33.7  \\ 
		
		ST-Puzzle~\cite{Kim19}  & K400       & \checkmark &$-$&$-$   & $224$      & R3D & AAAI 2019 &$-$& 63.9 &$-$& 33.7 \\ 
		VCOP~\cite{Xu19vcop}    & UCF101  &  \checkmark &$-$&$-$  & $112$     & R(2+1)D & CVPR 2019 &$-$& 72.4 &$-$& 30.9  \\ 
		DPC~\cite{Han19}        & K400      & \checkmark &$-$&$-$   & $128$     & R-2D3D & ICCVW 2019 &$-$& 75.7 &$-$& 35.7  \\ 
		SpeedNet~\cite{benaim_CVPR2020_speednet} &  K400  & \checkmark &$-$&$-$& $224$  & S3D-G & CVPR 2020 &$-$& 81.1 &$-$& 48.8  \\
		MemDPC~\cite{Han20_dpc}& K400   & \checkmark  &   $-$  & $-$ & 224 & R-2D3D  & ECCV 2020   & 54.1 & 86.1 & 30.5 & 54.5 \\ 
		CoCLR~\cite{Han20}        & K400   &  \checkmark &$-$&$-$   & $128$     & S3D &  NeurIPS 2020   & \textbf{74.5} & 87.9 & \textbf{46.1} &  54.6\\ 
		CVRL~\cite{Qian20}   & K400     & \checkmark &$-$&$-$   & $224$     & R3D-50 & arXiv 2020  &$-$& 92.2 &$-$& 66.7  \\ 
		CVRL~\cite{Qian20}   & K600     & \checkmark &$-$&$-$   & $224$     & R3D-50 & arXiv 2020  &$-$& \textbf{93.4} &$-$& \textbf{68.0}  \\ 

		\hline
		\end{tabular}
	\end{center}
	\vspace{5pt}
	\caption{\textbf{Comparison of self-supervised video representation learning methods.} Top section shows the multi-modal video representation learning approaches and bottom section shows the video-only representation learning methods. From left to right, we show the self-supervised training setting, e.g. dataset, modalities, resolution, and architecture. Two last right columns show the action recognition results on two datasets UCF101 and HMDB51 to measure the quality of self-supervised pre-trained model. HTM: HowTo100M, YT8M: YouTube8M, AS: AudioSet, IG-K: IG-Kinetics, K400: Kinetics400 and K600: Kinetics600.}
	\label{table:sota-self-supervised}
\end{table*}

\subsection{Multi-modality}
\label{subsec:multi_modality}

Multi-modal video understanding has attracted increasing attention in recent years \cite{gabeur2020multimodal,alwassel2020selfsupervised,Liu2019a,Qian20,Patrick20,alayrac2020selfsupervised,korbar2018cooperative}. There are two main categories for multi-modal video understanding. The first group of approaches use multi-modalities such as scene, object, motion, and audio to enrich the video representations. In the second group, the goal is to design a model which utilizes modality information as a supervision signal for pre-training models~\cite{sun_ICCV2019_videoBERT,miech20endtoend,yang2020hierarchical,ging2020coot,alayrac2020selfsupervised}.

\paragraph{Multi-modality for comprehensive video understanding} Learning a robust and comprehensive representation of video is extremely challenging due to the complexity of semantics in videos. Video data often includes variations in different forms including appearance, motion, audio, text or scene~\cite{gabeur2020multimodal,Liu2019a, piergiovanni2020learning}. Therefore, utilizing these multi-modal representations is a critical step in understanding video content more efficiently. The multi-modal representations of video can be approximated by gathering representations of various modalities such as scene, object, audio, motion, appearance and text. Ngiam \etal \cite{ngiam2011multimodal} was an early attempt to suggest using multiple modalities to obtain better features. They utilized videos of lips and their corresponding speech for multi-modal representation learning. Miech \etal \cite{miech2018learning} proposed a mixture-of embedding-experts model to combine multiple modalities including motion, appearance, audio, and face features and learn the shared embedding space between these modalities and text.
Roig \etal ~\cite{sarmiento2019mmp} combines multiple modalities such as action, scene, object and acoustic event features in a pyramidal structure for action recognition. They show that adding each modality improves the final action recognition accuracy. Both CE~\cite{Liu2019a} and MMT~\cite{gabeur2020multimodal}, follow a similar research line to \cite{miech2018learning} where the goal is to combine multiple-modalities to obtain a comprehensive representation of video for joint video-text representation learning.
Piergiovanni \etal \cite{piergiovanni2020learning} utilized textual data together with video data to learn a joint embedding space. Using this learned joint embedding space, the method is capable of doing zero-shot action recognition. This line of research is promising due to the availability of strong semantic extraction models and also success of transformers on both vision and language tasks.

\paragraph{Multi-modality for self-supervised video representation learning}
Most videos contain multiple modalities such as audio or text/caption. These modalities are great source of supervision for learning video representations \cite{alwassel2020selfsupervised,morgado_avid_cma,Patrick20,alayrac2020selfsupervised,pier20evolving}. 
Korbar \etal \cite{korbar2018cooperative} incorporated the natural synchronization between audio and video as a supervision signal in their contrastive learning objective for self-supervised representation learning. In multi-modal self-supervised representation learning, the dataset plays an important role. VideoBERT~\cite{sun_ICCV2019_videoBERT} collected 310K cooking videos from YouTube. However, this dataset is not publicly available. Similar to BERT, VideoBERT used a ``masked language model'' training objective and also quantized the visual representations into ``visual words''.  Miech \etal \cite{miech19howto100m} introduced HowTo100M dataset in 2019. This dataset includes 136M clips from 1.22M videos with their corresponding text. They collected the dataset from YouTube with the aim of obtaining instructional videos (how to perform an activity). In total, it covers 23.6K instructional tasks. MIL-NCE~\cite{miech20endtoend} used this dataset for self-supervised cross-modal representation learning. They tackled the problem of visually misaligned narrations, by considering multiple positive pairs in the contrastive learning objective. ActBERT~\cite{zhu_CVPR2020_ActBERT}, utilized HowTo100M dataset for pre-training of the model in a self-supervised way. They incorporated visual, action, text and object features for cross modal representation learning. Recently AVLnet~\cite{rouditchenko2020avlnet} and MMV~\cite{alayrac2020selfsupervised} considered three modalities visual, audio and language for self-supervised representation learning. This research direction is also increasingly getting more attention due to the success of contrastive learning on many vision and language tasks and the access to the abundance of unlabeled multi-modal video data on platforms such as  YouTube, Instagram or Flickr. The top section of Table~\ref{table:sota-self-supervised} compares multi-modal self-supervised representation learning methods. We will discuss more work on video-only representation learning in the next section.

\subsection{Self-supervised video representation learning}
\label{subsec:selfsup}

Self-supervised learning has attracted more attention recently as it is able to leverage a large amount of unlabeled data by designing a pretext task to obtain free supervisory signals from data itself. It first emerged in image representation learning. On images, the first stream of papers aimed at designing pretext tasks for completing missing information, such as image coloring~\cite{zhang2016colorful} and image reordering~\cite{pathakCVPR16context, gidaris2018unsupervised, Zhang_2017_CVPR}. The second stream of papers uses instance discrimination~\cite{wu2018unsupervised} as the pretext task and  contrastive losses~\cite{wu2018unsupervised,oord2018representation} for supervision. They learn visual representation by modeling visual similarity of object instances without class labels~\cite{wu2018unsupervised, he2019momentum, tian2019contrastive, chen2020improved, chen_arxiv2020_simCLR}.

Self-supervised learning is also viable for videos. Compared with images, videos has another axis, temporal dimension, which we can use to craft pretext tasks. Information completion tasks for this purpose include predicting the correct order of shuffled frames~\cite{misra2016shuffle, fernando2017self} and video clips~\cite{Xu19vcop}. Jing \etal~\cite{jing2018self} focus on the spatial dimension only by predicting the rotation angles of rotated video clips. Combining temporal and spatial information, several tasks have been introduced to solve a space-time cubic puzzle, anticipate future frames~\cite{vondrick2016anticipating}, forecast long-term motions~\cite{luo2017unsupervised} and predict motion and appearance statistics~\cite{wang2019self}. RSPNet~\cite{chen2020rspnet} and visual tempo~\cite{yang_arxiv2020_vthcl} exploit the relative speed between video clips as a supervision signal.

The added temporal axis can also provide flexibility in designing instance discrimination pretexts \cite{gordon2020watching,Qian20}. Inspired by the decoupling of 3D convolution to spatial and temporal separable convolutions~\cite{xie_ECCV2018_S3D}, Zhang \etal ~\cite{zhang2020hierarchically} proposed to decouple the video representation learning into two sub-tasks: spatial contrast and temporal contrast. Recently, Han \etal~\cite{Han20_dpc} proposed memory augmented dense predictive coding for self-supervised video representation learning. They split each video into several blocks and the embedding of future block is predicted by the combination of condensed representations in memory. 


The temporal continuity in videos inspires researchers to design other pretext tasks around correspondence. Wang \etal~\cite{CVPR2019_CycleTime} proposed to learn representation by performing cycle-consistency tracking.
Specifically, they track the same object backward and then forward in the consecutive video frames, and use the inconsistency between the start and end points as the loss function. TCC \cite{dwibedi2019temporal} is a concurrent paper. Instead of tracking local objects, \cite{dwibedi2019temporal} used cycle-consistency to perform frame-wise temporal alignment as a supervision signal.
\cite{li2019joint} was a follow-up work of \cite{CVPR2019_CycleTime}, and utilized both object-level and pixel-level correspondence across video frames.
Recently,  long-range temporal correspondence is modelled as a random walk graph to help learning video representation in~\cite{jabri2020walk}.

We compare video self-supervised representation learning methods at the bottom section of Table~\ref{table:sota-self-supervised}.
A clear trend can be observed that recent papers have achieved much better linear evaluation accuracy and fine-tuning accuracy comparable to supervised pre-training.
This shows that self-supervised learning could be a promising direction towards learning better video representations.

\section{Conclusion}
\label{sec:conclusion}
In this survey, we present a comprehensive review of 200+ deep learning based recent approaches to video action recognition. Although this is not an exhaustive list, we hope the survey serves as an easy-to-follow tutorial for those seeking to enter the field, and an inspiring discussion for those seeking to find new research directions.

\section*{Acknowledgement}
We would like to thank Peter Gehler, Linchao Zhu and Thomas Brady for constructive feedback and fruitful discussions.

{\small
\bibliographystyle{ieee_fullname}
\bibliography{video_cls_survey}

\begin{thebibliography}{100}\itemsep=-1pt

\bibitem{youtube8m}
Sami Abu-El-Haija, Nisarg Kothari, Joonseok Lee, Paul Natsev, George Toderici,
  Balakrishnan Varadarajan, and Sudheendra Vijayanarasimhan.
\newblock {YouTube-8M: A Large-Scale Video Classification Benchmark}.
\newblock {\em arXiv preprint arXiv:1609.08675}, 2016.

\bibitem{alayrac2020selfsupervised}
Jean-Baptiste Alayrac, Adrià Recasens, Rosalia Schneider, Relja Arandjelović,
  Jason Ramapuram, Jeffrey~De Fauw, Lucas Smaira, Sander Dieleman, and Andrew
  Zisserman.
\newblock {Self-Supervised MultiModal Versatile Networks}, 2020.

\bibitem{alwassel2020selfsupervised}
Humam Alwassel, Dhruv Mahajan, Bruno Korbar, Lorenzo Torresani, Bernard Ghanem,
  and Du Tran.
\newblock {Self-Supervised Learning by Cross-Modal Audio-Video Clustering}.
\newblock In {\em Advances in Neural Information Processing Systems (NeurIPS)},
  2020.

\bibitem{arandjelovic_CVPR2016_netVLAD}
Relja Arandjelović, Petr Gronat, Akihiko Torii, Tomas Pajdla, and Josef Sivic.
\newblock {NetVLAD: CNN Architecture for Weakly Supervised Place Recognition}.
\newblock In {\em The IEEE Conference on Computer Vision and Pattern
  Recognition (CVPR)}, 2016.

\bibitem{baccouche_HBU2011_sequential}
Moez Baccouche, Franck Mamalet, Christian Wolf, Christophe Garcia, and Atilla
  Baskurt.
\newblock {Sequential Deep Learning for Human Action Recognition}.
\newblock In {\em the Second International Conference on Human Behavior
  Understanding}, 2011.

\bibitem{benaim_CVPR2020_speednet}
Sagie Benaim, Ariel Ephrat, Oran Lang, Inbar Mosseri, William~T. Freeman,
  Michael Rubinstein, Michal Irani, and Tali Dekel.
\newblock {SpeedNet: Learning the Speediness in Videos}.
\newblock In {\em The IEEE Conference on Computer Vision and Pattern
  Recognition (CVPR)}, 2020.

\bibitem{gabriel_ICML2018_oneShot}
Gabriel Bender, Pieter-Jan Kindermans, Barret Zoph, Vijay Vasudevan, and Quoc
  Le.
\newblock {Understanding and Simplifying One-Shot Architecture Search}.
\newblock In {\em The International Conference on Machine Learning (ICML)},
  2018.

\bibitem{bhardwaj_CVPR2019_fewer}
Shweta Bhardwaj, Mukundhan Srinivasan, and Mitesh~M. Khapra.
\newblock {Efficient Video Classification Using Fewer Frames}.
\newblock In {\em The IEEE Conference on Computer Vision and Pattern
  Recognition (CVPR)}, 2019.

\bibitem{bilen_CVPR2016_dynamicImage}
Hakan Bilen, Basura Fernando, Efstratios Gavves, Andrea Vedaldi, and Stephen
  Gould.
\newblock {Dynamic Image Networks for Action Recognition}.
\newblock In {\em The IEEE Conference on Computer Vision and Pattern
  Recognition (CVPR)}, 2016.

\bibitem{Brattoli_CVPR2020_end2endZero}
Biagio Brattoli, Joseph Tighe, Fedor Zhdanov, Pietro Perona, and Krzysztof
  Chalupka.
\newblock {Rethinking Zero-Shot Video Classification: End-to-End Training for
  Realistic Applications}.
\newblock In {\em The IEEE Conference on Computer Vision and Pattern
  Recognition (CVPR)}, 2020.

\bibitem{cai_ICLR2019_proxylessnas}
Han Cai, Ligeng Zhu, and Song Han.
\newblock {Proxyless{NAS}: Direct Neural Architecture Search on Target Task and
  Hardware}.
\newblock In {\em The International Conference on Learning Representations
  (ICLR)}, 2019.

\bibitem{carreira2018short}
Joao Carreira, Eric Noland, Andras Banki-Horvath, Chloe Hillier, and Andrew
  Zisserman.
\newblock A short note about kinetics-600.
\newblock {\em arXiv preprint arXiv:1808.01340}, 2018.

\bibitem{carreira2019short}
Joao Carreira, Eric Noland, Chloe Hillier, and Andrew Zisserman.
\newblock A short note on the kinetics-700 human action dataset.
\newblock {\em arXiv preprint arXiv:1907.06987}, 2019.

\bibitem{carreira_CVPR2017_I3D}
Joao Carreira and Andrew Zisserman.
\newblock {Quo Vadis, Action Recognition? A New Model and the Kinetics
  Dataset}.
\newblock In {\em The IEEE Conference on Computer Vision and Pattern
  Recognition (CVPR)}, 2017.

\bibitem{chen_ICCV2019_TAA}
Min-Hung Chen, Zsolt Kira, Ghassan AlRegib, Jaekwon Yoo, Ruxin Chen, and Jian
  Zheng.
\newblock {Temporal Attentive Alignment for Large-Scale Video Domain
  Adaptation}.
\newblock In {\em The IEEE International Conference on Computer Vision (ICCV)},
  2019.

\bibitem{chen2020rspnet}
Peihao Chen, Deng Huang, Dongliang He, Xiang Long, Runhao Zeng, Shilei Wen,
  Mingkui Tan, and Chuang Gan.
\newblock {RSPNet: Relative Speed Perception for Unsupervised Video
  Representation Learning}, 2020.

\bibitem{chen_arxiv2020_simCLR}
Ting Chen, Simon Kornblith, Mohammad Norouzi, and Geoffrey Hinton.
\newblock {A Simple Framework for Contrastive Learning of Visual
  Representations}.
\newblock {\em arXiv preprint arXiv:2002.05709}, 2020.

\bibitem{chen2020improved}
Xinlei Chen, Haoqi Fan, Ross Girshick, and Kaiming He.
\newblock {Improved Baselines with Momentum Contrastive Learning}.
\newblock {\em arXiv preprint arXiv:2003.04297}, 2020.

\bibitem{chen2015webly}
Xinlei Chen and Abhinav Gupta.
\newblock {Webly Supervised Learning of Convolutional Networks}.
\newblock In {\em Proceedings of the IEEE International Conference on Computer
  Vision (ICCV)}, pages 1431--1439, 2015.

\bibitem{chen_ECCV2018_multiFiber}
Yunpeng Chen, Yannis Kalantidis, Jianshu Li, Shuicheng Yan, and Jiashi Feng.
\newblock {Multi-Fiber Networks for Video Recognition}.
\newblock In {\em The European Conference on Computer Vision (ECCV)}, 2018.

\bibitem{chen2019appending}
Zhikai Chen, Lingxi Xie, Shanmin Pang, Yong He, and Qi Tian.
\newblock {Appending Adversarial Frames for Universal Video Attack}.
\newblock {\em arXiv preprint arXiv:1912.04538}, 2019.

\bibitem{cherian_CVPR2017_gRankPool}
Anoop Cherian, Basura Fernando, Mehrtash Harandi, and Stephen Gould.
\newblock {Generalized Rank Pooling for Activity Recognition}.
\newblock In {\em The IEEE Conference on Computer Vision and Pattern
  Recognition (CVPR)}, 2017.

\bibitem{cheron_ICCV2015_PCNN}
Guilhem Cheron, Ivan Laptev, and Cordelia Schmid.
\newblock {P-CNN: Pose-based CNN Features for Action Recognition}.
\newblock In {\em The IEEE International Conference on Computer Vision (ICCV)},
  2015.

\bibitem{choi2019can}
Jinwoo Choi, Chen Gao, Joseph~CE Messou, and Jia-Bin Huang.
\newblock {Why Can't I Dance in the Mall? Learning to Mitigate Scene Bias in
  Action Recognition}.
\newblock In {\em Advances in Neural Information Processing Systems (NeurIPS)},
  pages 853--865, 2019.

\bibitem{vasileios_CVPR2018_potion}
Vasileios Choutas, Philippe Weinzaepfel, Jérôme Revaud, and Cordelia Schmid.
\newblock {PoTion: Pose MoTion Representation for Action Recognition}.
\newblock In {\em The IEEE Conference on Computer Vision and Pattern
  Recognition (CVPR)}, 2018.

\bibitem{crasto_CVPR2019_MARS}
Nieves Crasto, Philippe Weinzaepfel, Karteek Alahari, and Cordelia Schmid.
\newblock {MARS: Motion-Augmented RGB Stream for Action Recognition}.
\newblock In {\em The IEEE Conference on Computer Vision and Pattern
  Recognition (CVPR)}, 2019.

\bibitem{cubuk_CVPR2019_autoaug}
Ekin~D. Cubuk, Barret Zoph, Dandelion Mane, Vijay Vasudevan, and Quoc~V. Le.
\newblock {AutoAugment: Learning Augmentation Strategies From Data}.
\newblock In {\em The IEEE Conference on Computer Vision and Pattern
  Recognition (CVPR)}, 2019.

\bibitem{Damen2020Collection}
Dima Damen, Hazel Doughty, Giovanni~Maria Farinella, Sanja Fidler, Antonino
  Furnari, Evangelos Kazakos, Davide Moltisanti, Jonathan Munro, Toby Perrett,
  Will Price, and Michael Wray.
\newblock {The EPIC-KITCHENS Dataset: Collection, Challenges and Baselines}.
\newblock {\em IEEE Transactions on Pattern Analysis and Machine Intelligence
  (TPAMI)}, 2020.

\bibitem{damen2020rescaling}
Dima Damen, Hazel Doughty, Giovanni~Maria Farinella, Antonino Furnari,
  Evangelos Kazakos, Jian Ma, Davide Moltisanti, Jonathan Munro, Toby Perrett,
  Will Price, et~al.
\newblock {Rescaling Egocentric Vision}.
\newblock {\em arXiv preprint arXiv:2006.13256}, 2020.

\bibitem{imagenet_cvpr09}
J. Deng, W. Dong, R. Socher, L.-J. Li, K. Li, and L. Fei-Fei.
\newblock {ImageNet: A Large-Scale Hierarchical Image Database}.
\newblock In {\em CVPR}, 2009.

\bibitem{devries_arxiv2017_cutout}
Terrance DeVries and Graham~W Taylor.
\newblock {Improved Regularization of Convolutional Neural Networks with
  Cutout}.
\newblock {\em arXiv preprint arXiv:1708.04552}, 2017.

\bibitem{diba_arxiv2017_T3D}
Ali Diba, Mohsen Fayyaz, Vivek Sharma, Amir~Hossein Karami, Mohammad~Mahdi
  Arzani, Rahman Yousefzadeh, and Luc~Van Gool.
\newblock {Temporal 3D ConvNets: New Architecture and Transfer Learning for
  Video Classification}.
\newblock {\em arXiv preprint arXiv:1711.08200}, 2017.

\bibitem{diba_ECCV2018_stcnet}
Ali Diba, Mohsen Fayyaz, Vivek Sharma, M. Mahdi~Arzani, Rahman Yousefzadeh,
  Juergen Gall, and Luc Van~Gool.
\newblock {Spatio-Temporal Channel Correlation Networks for Action
  Classification}.
\newblock In {\em The European Conference on Computer Vision (ECCV)}, 2018.

\bibitem{hvu}
Ali Diba, Mohsen Fayyaz, Vivek Sharma, Manohar Paluri, J{\"u}rgen Gall, Rainer
  Stiefelhagen, and Luc Van~Gool.
\newblock {Large Scale Holistic Video Understanding}.
\newblock In {\em European Conference on Computer Vision}, pages 593--610.
  Springer, 2020.

\bibitem{diba_arxiv2016_efficient3DTS}
Ali Diba, Ali~Mohammad Pazandeh, and Luc~Van Gool.
\newblock {Efficient Two-Stream Motion and Appearance 3D CNNs for Video
  Classification}.
\newblock {\em arXiv preprint arXiv:1608.08851}, 2016.

\bibitem{diba_CVPR2017_TLE}
Ali Diba, Vivek Sharma, and Luc Van~Gool.
\newblock {Deep Temporal Linear Encoding Networks}.
\newblock In {\em The IEEE Conference on Computer Vision and Pattern
  Recognition (CVPR)}, 2017.

\bibitem{donahue_CVPR2015_LRCN}
Jeff Donahue, Lisa~Anne Hendricks, Marcus Rohrbach, Subhashini Venugopalan,
  Sergio Guadarrama, Kate Saenko, and Trevor Darrell.
\newblock {Long-term Recurrent Convolutional Networks for Visual Recognition
  and Description}.
\newblock In {\em The IEEE Conference on Computer Vision and Pattern
  Recognition (CVPR)}, 2015.

\bibitem{duan2016omnisource}
Hao~Dong Duan, Yue Zhao, Yuanjun Xiong, Wentao Liu, and Dahu Lin.
\newblock {Omni-sourced Webly-supervised Learning for Video Recognition}.
\newblock In {\em European Conference on Computer Vision}, 2020.

\bibitem{dwibedi2019temporal}
Debidatta Dwibedi, Yusuf Aytar, Jonathan Tompson, Pierre Sermanet, and Andrew
  Zisserman.
\newblock {Temporal Cycle-Consistency Learning}.
\newblock In {\em Proceedings of the IEEE Conference on Computer Vision and
  Pattern Recognition}, pages 1801--1810, 2019.

\bibitem{activityNet}
Bernard~Ghanem Fabian Caba~Heilbron, Victor~Escorcia and Juan~Carlos Niebles.
\newblock {ActivityNet: A Large-Scale Video Benchmark for Human Activity
  Understanding}.
\newblock In {\em The IEEE Conference on Computer Vision and Pattern
  Recognition (CVPR)}, 2015.

\bibitem{fan2018watching}
Hehe Fan, Zhongwen Xu, Linchao Zhu, Chenggang Yan, Jianjun Ge, and Yi Yang.
\newblock Watching a small portion could be as good as watching all: Towards
  efficient video classification.
\newblock In {\em IJCAI International Joint Conference on Artificial
  Intelligence}, 2018.

\bibitem{fan_CVPR2018_end2end}
Lijie Fan, Wenbing Huang, Chuang Gan, Stefano Ermon, Boqing Gong, and Junzhou
  Huang.
\newblock {End-to-End Learning of Motion Representation for Video
  Understanding}.
\newblock In {\em The IEEE Conference on Computer Vision and Pattern
  Recognition (CVPR)}, 2018.

\bibitem{fan_NIPS2019_bLVNet}
Quanfu Fan, Chun-Fu~(Richard) Chen, Hilde Kuehne, Marco Pistoia, and David Cox.
\newblock {More Is Less: Learning Efficient Video Representations by Big-Little
  Network and Depthwise Temporal Aggregation}.
\newblock In {\em Advances in Neural Information Processing Systems (NeurIPS)},
  2019.

\bibitem{feichtenhofer_CVPR2020_X3D}
Christoph Feichtenhofer.
\newblock {X3D: Expanding Architectures for Efficient Video Recognition}.
\newblock In {\em The IEEE Conference on Computer Vision and Pattern
  Recognition (CVPR)}, 2020.

\bibitem{feichtenhofer_ICCV2019_slowfast}
Christoph Feichtenhofer, Haoqi Fan, Jitendra Malik, and Kaiming He.
\newblock {SlowFast Networks for Video Recognition}.
\newblock In {\em The IEEE International Conference on Computer Vision (ICCV)},
  2019.

\bibitem{feichtenhofer_NIPS2016_stResidual}
Christoph Feichtenhofer, Axel Pinz, and Richard~P. Wildes.
\newblock {Spatiotemporal Residual Networks for Video Action Recognition}.
\newblock In {\em Advances in Neural Information Processing Systems (NeurIPS)},
  2016.

\bibitem{feichtenhofer_CVPR2017_stMultiplier}
Christoph Feichtenhofer, Axel Pinz, and Richard~P Wildes.
\newblock {Spatiotemporal Multiplier Networks for Video Action Recognition}.
\newblock In {\em The IEEE Conference on Computer Vision and Pattern
  Recognition (CVPR)}, 2017.

\bibitem{feichtenhofer_CVPR2018_learned}
Christoph Feichtenhofer, Axel Pinz, Richard~P. Wildes, and Andrew Zisserman.
\newblock {What Have We Learned From Deep Representations for Action
  Recognition?}
\newblock In {\em The IEEE Conference on Computer Vision and Pattern
  Recognition (CVPR)}, 2018.

\bibitem{feichtenhofer_IJCV2019_insights}
Christoph Feichtenhofer, Axel Pinz, Richard~P. Wildes, and Andrew Zisserman.
\newblock {Deep insights into convolutional networks for video recognition}.
\newblock {\em International Journal of Computer Vision (IJCV)}, 2019.

\bibitem{feichtenhofer_CVPR2016_fusion}
Christoph Feichtenhofer, Axel Pinz, and Andrew Zisserman.
\newblock {Convolutional Two-Stream Network Fusion for Video Action
  Recognition}.
\newblock In {\em The IEEE Conference on Computer Vision and Pattern
  Recognition (CVPR)}, 2016.

\bibitem{fernando_CVPR2016_HrankPool}
Basura Fernando, Peter Anderson, Marcus Hutter, and Stephen Gould.
\newblock {Discriminative Hierarchical Rank Pooling for Activity Recognition}.
\newblock In {\em The IEEE Conference on Computer Vision and Pattern
  Recognition (CVPR)}, 2016.

\bibitem{fernando2017self}
Basura Fernando, Hakan Bilen, Efstratios Gavves, and Stephen Gould.
\newblock Self-supervised video representation learning with odd-one-out
  networks.
\newblock In {\em Proceedings of the IEEE conference on computer vision and
  pattern recognition}, pages 3636--3645, 2017.

\bibitem{fernando_CVPR2015_videoDarwin}
Basura Fernando, Efstratios Gavves, Jose~Oramas M., Amir Ghodrati, and Tinne
  Tuytelaars.
\newblock {Modeling Video Evolution For Action Recognition}.
\newblock In {\em The IEEE Conference on Computer Vision and Pattern
  Recognition (CVPR)}, 2015.

\bibitem{fernando_ICML2016_e2eRankPool}
Basura Fernando and Stephen Gould.
\newblock {Learning End-to-end Video Classification with Rank-Pooling}.
\newblock In {\em The International Conference on Machine Learning (ICML)},
  2016.

\bibitem{gabeur2020multimodal}
Gabeur et~al.
\newblock {Multi-modal Transformer for Video Retrieval}.
\newblock {\em arxiv:2007.10639}, 2020.

\bibitem{gammulle_WACV2017_TS_LSTM_fusion}
Harshala Gammulle, Simon Denman, Sridha Sridharan, and Clinton Fookes.
\newblock {Two Stream LSTM: A Deep Fusion Framework for Human Action
  Recognition}.
\newblock In {\em The IEEE Winter Conference on Applications of Computer Vision
  (WACV)}, 2017.

\bibitem{SIR_Gan_AAAI15}
Chuang Gan, Ming Lin, Yi Yang, Yueting Zhuang, and Alexander G.Hauptmann.
\newblock {Exploring Semantic Inter-Class Relationships (SIR) for Zero-Shot
  Action Recognition}.
\newblock In {\em AAAI}, 2015.

\bibitem{gan2016webly}
Chuang Gan, Chen Sun, Lixin Duan, and Boqing Gong.
\newblock Webly-supervised video recognition by mutually voting for relevant
  web images and web video frames.
\newblock In {\em European Conference on Computer Vision}, pages 849--866.
  Springer, 2016.

\bibitem{garcia_ECCV2018_distill}
Nuno~C. Garcia, Pietro Morerio, and Vittorio Murino.
\newblock {Modality Distillation with Multiple Stream Networks for Action
  Recognition}.
\newblock In {\em The European Conference on Computer Vision (ECCV)}, 2018.

\bibitem{ghadiyaram_CVPR2019_IG65M}
Deepti Ghadiyaram, Matt Feiszli, Du Tran, Xueting Yan, Heng Wang, and D.
  Mahajan.
\newblock {Large-Scale Weakly-Supervised Pre-Training for Video Action
  Recognition}.
\newblock {\em The IEEE Conference on Computer Vision and Pattern Recognition
  (CVPR)}, 2019.

\bibitem{gidaris2018unsupervised}
Spyros Gidaris, Praveer Singh, and Nikos Komodakis.
\newblock Unsupervised representation learning by predicting image rotations.
\newblock {\em arXiv preprint arXiv:1803.07728}, 2018.

\bibitem{ging2020coot}
Simon Ging, Mohammadreza Zolfaghari, Hamed Pirsiavash, and Thomas Brox.
\newblock Coot: Cooperative hierarchical transformer for video-text
  representation learning.
\newblock In {\em Advances in Neural Information Processing Systems}, 2020.

\bibitem{girdhar_CVPR2017_actionVLAD}
Rohit Girdhar, Deva Ramanan, Abhinav Gupta, Josef Sivic, and Bryan Russell.
\newblock {ActionVLAD: Learning Spatio-Temporal Aggregation for Action
  Classification}.
\newblock In {\em The IEEE Conference on Computer Vision and Pattern
  Recognition (CVPR)}, 2017.

\bibitem{girdhar2019distinit}
Rohit Girdhar, Du Tran, Lorenzo Torresani, and Deva Ramanan.
\newblock Distinit: Learning video representations without a single labeled
  video.
\newblock In {\em Proceedings of the IEEE International Conference on Computer
  Vision}, pages 852--861, 2019.

\bibitem{Goodale_Neuro1992_separate}
M.~A. Goodale and A.~D. Milner.
\newblock {Separate Visual Pathways for Perception and Action}.
\newblock {\em Trends in Neurosciences}, 1992.

\bibitem{goodfellow2014explaining}
Ian Goodfellow, Jonathon Shlens, and Christian Szegedy.
\newblock Explaining and harnessing adversarial examples.
\newblock In {\em International Conference on Learning Representations}, 2015.

\bibitem{gordon2020watching}
Daniel Gordon, Kiana Ehsani, Dieter Fox, and Ali Farhadi.
\newblock {Watching the World Go By: Representation Learning from Unlabeled
  Videos}.
\newblock {\em arXiv preprint arXiv:2003.07990}, 2020.

\bibitem{goyal_arxiv2017_oneHour}
Priya Goyal, Piotr Dollár, Ross Girshick, Pieter Noordhuis, Lukasz Wesolowski,
  Aapo Kyrola, Andrew Tulloch, Yangqing Jia, and Kaiming He.
\newblock {Accurate, Large Minibatch SGD: Training ImageNet in 1 Hour}.
\newblock {\em arXiv preprint arXiv:1706.02677}, 2017.

\bibitem{sthsth}
Raghav Goyal, Samira~Ebrahimi Kahou, Vincent Michalski, Joanna Materzynska,
  Susanne Westphal, Heuna Kim, Valentin Haenel, Ingo Fruend, Peter Yianilos,
  Moritz Mueller-Freitag, et~al.
\newblock {The" Something Something" Video Database for Learning and Evaluating
  Visual Common Sense}.
\newblock In {\em The IEEE International Conference on Computer Vision (ICCV)},
  2017.

\bibitem{ava}
Chunhui Gu, Chen Sun, David~A. Ross, Carl Vondrick, Caroline Pantofaru, Yeqing
  Li, Sudheendra Vijayanarasimhan, George Toderici, Susanna Ricco, Rahul
  Sukthankar, Cordelia Schmid, and Jitendra Malik.
\newblock {AVA: A Video Dataset of Spatio-Temporally Localized Atomic Visual
  Actions}.
\newblock In {\em The IEEE Conference on Computer Vision and Pattern
  Recognition (CVPR)}, 2018.

\bibitem{Han19}
Tengda Han, Weidi Xie, and Andrew Zisserman.
\newblock Video representation learning by dense predictive coding.
\newblock In {\em Workshop on Large Scale Holistic Video Understanding, ICCV},
  2019.

\bibitem{Han20_dpc}
Tengda Han, Weidi Xie, and Andrew Zisserman.
\newblock Memory-augmented dense predictive coding for video representation
  learning.
\newblock In {\em European Conference on Computer Vision}, 2020.

\bibitem{Han20}
Tengda Han, Weidi Xie, and Andrew Zisserman.
\newblock Self-supervised co-training for video representation learning.
\newblock In {\em Neurips}, 2020.

\bibitem{hara_CVPR2018_3DResNet}
Kensho Hara, Hirokatsu Kataoka, and Yutaka Satoh.
\newblock {Can Spatiotemporal 3D CNNs Retrace the History of 2D CNNs and
  ImageNet?}
\newblock In {\em The IEEE Conference on Computer Vision and Pattern
  Recognition (CVPR)}, 2018.

\bibitem{he2019momentum}
Kaiming He, Haoqi Fan, Yuxin Wu, Saining Xie, and Ross Girshick.
\newblock Momentum contrast for unsupervised visual representation learning.
\newblock {\em arXiv preprint arXiv:1911.05722}, 2019.

\bibitem{he_CVPR2016_resnet}
Kaiming He, Xiangyu Zhang, Shaoqing Ren, and Jian Sun.
\newblock {Deep Residual Learning for Image Recognition}.
\newblock In {\em The IEEE Conference on Computer Vision and Pattern
  Recognition (CVPR)}, 2016.

\bibitem{herath_arxiv2016_survey}
Samitha Herath, Mehrtash Harandi, and Fatih Porikli.
\newblock {Going Deeper into Action Recognition: A Survey}.
\newblock {\em arXiv preprint arXiv:1605.04988}, 2016.

\bibitem{hochreiter_NC1997_LSTM}
Sepp Hochreiter and J{\"u}rgen Schmidhuber.
\newblock {Long Short-Term Memory}.
\newblock {\em Neural Computation}, 1997.

\bibitem{horn_AI1981_opticalFlow}
Berthold~K.P. Horn and Brian~G. Rhunck.
\newblock {Determining Optical Flow}.
\newblock {\em Artificial Intelligence}, 1981.

\bibitem{howard_ICCV2019_mobilenetv3}
Andrew Howard, Mark Sandler, Grace Chu, Liang-Chieh Chen, Bo Chen, Mingxing
  Tan, Weijun Wang, Yukun Zhu, Ruoming Pang, Vijay Vasudevan, Quoc~V. Le, and
  Hartwig Adam.
\newblock {Searching for MobileNetV3}.
\newblock In {\em The IEEE International Conference on Computer Vision (ICCV)},
  2019.

\bibitem{hu_CVPR2018_senet}
Jie Hu, Li Shen, and Gang Sun.
\newblock {Squeeze-and-Excitation Networks}.
\newblock In {\em The IEEE Conference on Computer Vision and Pattern
  Recognition (CVPR)}, 2018.

\bibitem{huang_CVPR2018_makeVideo}
De-An Huang, Vignesh Ramanathan, Dhruv Mahajan, Lorenzo Torresani, Manohar
  Paluri, Li Fei-Fei, and Juan Carlos~Niebles.
\newblock {What Makes a Video a Video: Analyzing Temporal Information in Video
  Understanding Models and Datasets}.
\newblock In {\em The IEEE Conference on Computer Vision and Pattern
  Recognition (CVPR)}, 2018.

\bibitem{huang_CVPR2017_densenet}
Gao Huang, Zhuang Liu, Laurens van~der Maaten, and Kilian~Q. Weinberger.
\newblock {Densely Connected Convolutional Networks}.
\newblock In {\em The IEEE Conference on Computer Vision and Pattern
  Recognition (CVPR)}, 2017.

\bibitem{hussein_ICCV2019_timeception}
Noureldien Hussein, Efstratios Gavves, and Arnold~WM Smeulders.
\newblock Timeception for complex action recognition.
\newblock In {\em Proceedings of the IEEE Conference on Computer Vision and
  Pattern Recognition}, pages 254--263, 2019.

\bibitem{flownet2}
E. Ilg, N. Mayer, T. Saikia, M. Keuper, A. Dosovitskiy, and T. Brox.
\newblock {FlowNet 2.0: Evolution of Optical Flow Estimation with Deep
  Networks}.
\newblock In {\em The IEEE Conference on Computer Vision and Pattern
  Recognition (CVPR)}, 2017.

\bibitem{inkawhich2018adversarial}
Nathan Inkawhich, Matthew Inkawhich, Yiran Chen, and Hai Li.
\newblock Adversarial attacks for optical flow-based action recognition
  classifiers.
\newblock {\em arXiv preprint arXiv:1811.11875}, 2018.

\bibitem{jabri2020walk}
Allan Jabri, Andrew Owens, and Alexei~A Efros.
\newblock Space-time correspondence as a contrastive random walk.
\newblock {\em Advances in Neural Information Processing Systems}, 2020.

\bibitem{2015_ICCV_ZSL_detection}
Mihir Jain, Jan~C van Gemert, Thomas Mensink, and Cees~GM Snoek.
\newblock {Objects2action: Classifying and Localizing Actions without Any Video
  Example}.
\newblock In {\em ICCV}, 2015.

\bibitem{jamal_BMVC2018_DAspace}
Arshad Jamal, Vinay~P Namboodiri, Dipti Deodhare, and KS Venkatesh.
\newblock {Deep Domain Adaptation in Action Space}.
\newblock In {\em The British Machine Vision Conference (BMVC)}, 2018.

\bibitem{ji2020action}
Jingwei Ji, Ranjay Krishna, Li Fei-Fei, and Juan~Carlos Niebles.
\newblock Action genome: Actions as compositions of spatio-temporal scene
  graphs.
\newblock In {\em Proceedings of the IEEE/CVF Conference on Computer Vision and
  Pattern Recognition}, pages 10236--10247, 2020.

\bibitem{ji_PAMI2012_3DCNN}
Shuiwang Ji, Wei Xu, Ming Yang, and Kai Yu.
\newblock {3D Convolutional Neural Networks for Human Action Recognition}.
\newblock {\em IEEE Transactions on Pattern Analysis and Machine Intelligence
  (PAMI)}, 2012.

\bibitem{jiang_ICCV2019_STM}
Boyuan Jiang, MengMeng Wang, Weihao Gan, Wei Wu, and Junjie Yan.
\newblock {STM: SpatioTemporal and Motion Encoding for Action Recognition}.
\newblock In {\em The IEEE International Conference on Computer Vision (ICCV)},
  2019.

\bibitem{jiang2019black}
Linxi Jiang, Xingjun Ma, Shaoxiang Chen, James Bailey, and Yu-Gang Jiang.
\newblock Black-box adversarial attacks on video recognition models.
\newblock In {\em Proceedings of the 27th ACM International Conference on
  Multimedia}, pages 864--872, 2019.

\bibitem{jing2018self}
Longlong Jing, Xiaodong Yang, Jingen Liu, and Yingli Tian.
\newblock Self-supervised spatiotemporal feature learning via video rotation
  prediction.
\newblock {\em arXiv preprint arXiv:1811.11387}, 2018.

\bibitem{Jinwoo_shuffle_ECCV2020}
Samuel~Schulter Jinwoo~Choi, Gaurav~Sharma and Jia-Bin Huang.
\newblock {Shuffle and Attend: Video Domain Adaptation}.
\newblock In {\em The European Conference on Computer Vision (ECCV)}, 2020.

\bibitem{estevam_arxiv2019_survey}
Valter Luís~Estevam Junior, Helio Pedrini, and David Menotti.
\newblock {Zero-Shot Action Recognition in Videos: A Survey}.
\newblock {\em arXiv preprint arXiv:1909.06423}, 2019.

\bibitem{jegou_CVPR2010_VLAD}
Hervé Jégou, Matthijs Douze, Cordelia Schmid, and Patrick Pérez.
\newblock {Aggregating Local Descriptors into a Compact Image Representation}.
\newblock In {\em The IEEE Conference on Computer Vision and Pattern
  Recognition (CVPR)}, 2010.

\bibitem{kar_CVPR2017_adascan}
Amlan Kar, Nishant Rai, Karan Sikka, and Gaurav Sharma.
\newblock {AdaScan: Adaptive Scan Pooling in Deep Convolutional Neural Networks
  for Human Action Recognition in Videos}.
\newblock In {\em The IEEE Conference on Computer Vision and Pattern
  Recognition (CVPR)}, 2017.

\bibitem{karpathy_CVPR2014_videoCNN}
Andrej Karpathy, George Toderici, Sanketh Shetty, Thomas Leung, Rahul
  Sukthankar, and Li Fei-Fei.
\newblock {Large-Scale Video Classification with Convolutional Neural
  Networks}.
\newblock In {\em The IEEE Conference on Computer Vision and Pattern
  Recognition (CVPR)}, 2014.

\bibitem{kinetics400}
Will Kay, Joao Carreira, Karen Simonyan, Brian Zhang, Chloe Hillier, Sudheendra
  Vijayanarasimhan, Fabio Viola, Tim Green, Trevor Back, Paul Natsev, Mustafa
  Suleyman, and Andrew Zisserman.
\newblock {The Kinetics Human Action Video Dataset}.
\newblock {\em arXiv preprint arXiv:1705.06950}, 2017.

\bibitem{Kazakos_2019_ICCV}
Evangelos Kazakos, Arsha Nagrani, Andrew Zisserman, and Dima Damen.
\newblock {EPIC-Fusion: Audio-Visual Temporal Binding for Egocentric Action
  Recognition}.
\newblock In {\em ICCV}, 2019.

\bibitem{Kim19}
Dahun Kim, Donghyeon Cho, and In~So Kweon.
\newblock {Self-Supervised Video Representation Learning with Space-Time Cubic
  Puzzles}.
\newblock In {\em AAAI}, 2019.

\bibitem{kobayashi2019fine}
Takuya Kobayashi, Yoshimitsu Aoki, Shogo Shimizu, Katsuhiro Kusano, and Seiji
  Okumura.
\newblock Fine-grained action recognition in assembly work scenes by drawing
  attention to the hands.
\newblock In {\em 2019 15th International Conference on Signal-Image Technology
  \& Internet-Based Systems (SITIS)}, pages 440--446. IEEE, 2019.

\bibitem{kong_arxiv2018_survey}
Yu Kong and Yun Fu.
\newblock {Human Action Recognition and Prediction: A Survey}.
\newblock {\em arXiv preprint arXiv:1806.11230}, 2018.

\bibitem{korbar2018cooperative}
Bruno Korbar, Du Tran, and Lorenzo Torresani.
\newblock Cooperative learning of audio and video models from self-supervised
  synchronization, 2018.

\bibitem{korbar_ICCV2019_scsampler}
Bruno Korbar, Du Tran, and Lorenzo Torresani.
\newblock {SCSampler: Sampling Salient Clips From Video for Efficient Action
  Recognition}.
\newblock In {\em The IEEE International Conference on Computer Vision (ICCV)},
  2019.

\bibitem{krizhevsky_NIPS2012_imagenetCNN}
Alex Krizhevsky, Ilya Sutskever, and Geoffrey~E. Hinton.
\newblock {ImageNet Classification with Deep Convolutional Neural Networks}.
\newblock In {\em Advances in Neural Information Processing Systems (NeurIPS)},
  2012.

\bibitem{kuehne2014language}
Hilde Kuehne, Ali Arslan, and Thomas Serre.
\newblock The language of actions: Recovering the syntax and semantics of
  goal-directed human activities.
\newblock In {\em Proceedings of the IEEE conference on computer vision and
  pattern recognition}, pages 780--787, 2014.

\bibitem{hmdb51}
Hildegard Kuehne, Hueihan Jhuang, Est{\'\i}baliz Garrote, Tomaso Poggio, and
  Thomas Serre.
\newblock {HMDB: A Large Video Database for Human Motion Recognition}.
\newblock In {\em The IEEE International Conference on Computer Vision (ICCV)},
  2011.

\bibitem{kwon2020motionsqueeze}
Heeseung Kwon, Manjin Kim, Suha Kwak, and Minsu Cho.
\newblock Motionsqueeze: Neural motion feature learning for video
  understanding.
\newblock In {\em ECCV}, 2020.

\bibitem{kopuklu_ICCVW2019_resource}
Okan Köpüklü, Neslihan Kose, Ahmet Gunduz, and Gerhard Rigoll.
\newblock {Resource Efficient 3D Convolutional Neural Networks}.
\newblock In {\em The IEEE International Conference on Computer Vision (ICCV)
  Workshop}, 2019.

\bibitem{lan_CVPR2015_MIFS}
Zhenzhong Lan, Ming Lin, Xuanchong Li, Alexander~G. Hauptmann, and Bhiksha Raj.
\newblock {Beyond Gaussian Pyramid: Multi-skip Feature Stacking for Action
  Recognition}.
\newblock In {\em The IEEE Conference on Computer Vision and Pattern
  Recognition (CVPR)}, 2015.

\bibitem{lan_arxiv2015_bothWorlds}
Zhenzhong Lan, Dezhong Yao, Ming Lin, Shoou-I Yu, and Alexander Hauptmann.
\newblock {The Best of Both Worlds: Combining Data-independent and Data-driven
  Approaches for Action Recognition}.
\newblock {\em arXiv preprint arXiv:1505.04427}, 2015.

\bibitem{lan_CVPRW2017_DVOF}
Zhenzhong Lan, Yi Zhu, Alexander~G. Hauptmann, and Shawn Newsam.
\newblock {Deep Local Video Feature for Action Recognition}.
\newblock In {\em The IEEE Conference on Computer Vision and Pattern
  Recognition (CVPR) Workshops}, 2017.

\bibitem{OPN}
Hsin-Ying Lee, Jia-Bin Huang, Maneesh~Kumar Singh, and Ming-Hsuan Yang.
\newblock {Unsupervised Representation Learning by Sorting Sequence}, 2017.

\bibitem{lee_ECCV2018_motionFilter}
Myunggi Lee, Seungeui Lee, Sungjoon Son, Gyutae Park, and Nojun Kwak.
\newblock {Motion Feature Network: Fixed Motion Filter for Action Recognition}.
\newblock In {\em The European Conference on Computer Vision (ECCV)}, 2018.

\bibitem{li2020ava}
Ang Li, Meghana Thotakuri, David~A Ross, Jo{\~a}o Carreira, Alexander
  Vostrikov, and Andrew Zisserman.
\newblock The ava-kinetics localized human actions video dataset.
\newblock {\em arXiv preprint arXiv:2005.00214}, 2020.

\bibitem{li_ICMR2016_granular}
Qing Li, Zhaofan Qiu, Ting Yao, Tao Mei, Yong Rui, and Jiebo Luo.
\newblock {Action Recognition by Learning Deep Multi-Granular Spatio-Temporal
  Video Representation}.
\newblock In {\em The ACM International Conference on Multimedia Retrieval
  (ICMR)}, 2016.

\bibitem{li2018adversarial}
Shasha Li, Ajaya Neupane, Sujoy Paul, Chengyu Song, Srikanth~V Krishnamurthy,
  Amit K~Roy Chowdhury, and Ananthram Swami.
\newblock Adversarial perturbations against real-time video classification
  systems.
\newblock {\em arXiv preprint arXiv:1807.00458}, 2018.

\bibitem{li2019joint}
Xueting Li, Sifei Liu, Shalini De~Mello, Xiaolong Wang, Jan Kautz, and
  Ming-Hsuan Yang.
\newblock {Joint-task self-supervised learning for temporal correspondence}.
\newblock In {\em Advances in Neural Information Processing Systems}, pages
  317--327, 2019.

\bibitem{li2020directional}
Xinyu Li, Bing Shuai, and Joseph Tighe.
\newblock Directional temporal modeling for action recognition.
\newblock In {\em European Conference on Computer Vision}, pages 275--291.
  Springer, 2020.

\bibitem{li_CVPR2020_TEA}
Yan Li, Bin Ji, Xintian Shi, Jianguo Zhang, Bin Kang, and Limin Wang.
\newblock {TEA: Temporal Excitation and Aggregation for Action Recognition}.
\newblock In {\em The IEEE Conference on Computer Vision and Pattern
  Recognition (CVPR)}, 2020.

\bibitem{Li_CVPR2016_VLAD3}
Yingwei Li, Weixin Li, Vijay Mahadevan, and Nuno Vasconcelos.
\newblock {VLAD3: Encoding Dynamics of Deep Features for Action Recognition}.
\newblock In {\em The IEEE Conference on Computer Vision and Pattern
  Recognition (CVPR)}, 2016.

\bibitem{li_ECCV2018_resound}
Yingwei Li, Yi Li, and Nuno Vasconcelos.
\newblock {RESOUND: Towards Action Recognition without Representation Bias}.
\newblock In {\em The European Conference on Computer Vision (ECCV)}, 2018.

\bibitem{li_CVIU2018_videoLSTM}
Zhenyang Li, Kirill Gavrilyuk, Efstratios Gavves, Mihir Jain, and Cees~GM
  Snoek.
\newblock {VideoLSTM Convolves, Attends and Flows for Action Recognition}.
\newblock {\em Computer Vision and Image Understanding (CVIU)}, 2018.

\bibitem{lim_NIPS2019_fastautoaug}
Sungbin Lim, Ildoo Kim, Taesup Kim, Chiheon Kim, and Sungwoong Kim.
\newblock {Fast AutoAugment}.
\newblock In {\em Advances in Neural Information Processing Systems (NeurIPS)},
  2019.

\bibitem{lin_NIPSW2019_kinetics15min}
Ji Lin, Chuang Gan, and Song Han.
\newblock {Training Kinetics in 15 Minutes: Large-scale Distributed Training on
  Videos}.
\newblock In {\em Advances in Neural Information Processing Systems (NeurIPS)
  Workshop}, 2019.

\bibitem{lin_ICCV2019_TSM}
Ji Lin, Chuang Gan, and Song Han.
\newblock {TSM: Temporal Shift Module for Efficient Video Understanding}.
\newblock In {\em The IEEE International Conference on Computer Vision (ICCV)},
  2019.

\bibitem{Liu2019a}
Liu et~al.
\newblock {Use What You Have: Video Retrieval using Representations from
  Collaborative Experts}.
\newblock {\em arxiv:1907.13487}, 2019.

\bibitem{liu_ICLR2019_darts}
Hanxiao Liu, Karen Simonyan, and Yiming Yang.
\newblock {DARTS: Differentiable Architecture Search}.
\newblock In {\em The International Conference on Learning Representations
  (ICLR)}, 2019.

\bibitem{liu2019forecasting}
Miao Liu, Siyu Tang, Yin Li, and James Rehg.
\newblock Forecasting human object interaction: Joint prediction of motor
  attention and egocentric activity.
\newblock In {\em ECCV}, 2020.

\bibitem{liu_AAAI2020_TEINet}
Zhaoyang Liu, Donghao Luo, Yabiao Wang, Limin Wang, Ying Tai, Chengjie Wang,
  Jilin Li, Feiyue Huang, and Tong Lu.
\newblock {TEINet: Towards an Efficient Architecture for Video Recognition}.
\newblock In {\em The Conference on Artificial Intelligence (AAAI)}, 2020.

\bibitem{liu_arxiv2020_TAM}
Zhaoyang Liu, Limin Wang, Wayne Wu, Chen Qian, and Tong Lu.
\newblock {TAM: Temporal Adaptive Module for Video Recognition}.
\newblock {\em arXiv preprint arXiv:2005.06803}, 2020.

\bibitem{luo2017unsupervised}
Zelun Luo, Boya Peng, De-An Huang, Alexandre Alahi, and Li Fei-Fei.
\newblock Unsupervised learning of long-term motion dynamics for videos.
\newblock In {\em Proceedings of the IEEE Conference on Computer Vision and
  Pattern Recognition}, pages 2203--2212, 2017.

\bibitem{ma_SP2019_TS_LSTM}
Chih-Yao Ma, Min-Hung Chen, Zsolt Kira, and Ghassan AlRegib.
\newblock {TS-LSTM and Temporal-Inception: Exploiting Spatiotemporal Dynamics
  for Activity Recognition}.
\newblock {\em Signal Processing: Image Communication}, 2019.

\bibitem{ma2016going}
Minghuang Ma, Haoqi Fan, and Kris~M Kitani.
\newblock Going deeper into first-person activity recognition.
\newblock In {\em CVPR}, 2016.

\bibitem{spatial_aware_Mettes_ICCV17}
Pascal Mettes and Cees G.~M. Snoek.
\newblock {Spatial-Aware Object Embeddings for Zero-Shot Localization and
  Classification of Actions}.
\newblock In {\em ICCV}, 2017.

\bibitem{miech20endtoend}
Antoine Miech, Jean-Baptiste Alayrac, Lucas Smaira, Ivan Laptev, Josef Sivic,
  and Andrew Zisserman.
\newblock {E}nd-to-{E}nd {L}earning of {V}isual {R}epresentations from
  {U}ncurated {I}nstructional {V}ideos.
\newblock In {\em CVPR}, 2020.

\bibitem{miech2018learning}
Antoine Miech, Ivan Laptev, and Josef Sivic.
\newblock Learning a text-video embedding from incomplete and heterogeneous
  data.
\newblock {\em arXiv preprint arXiv:1804.02516}, 2018.

\bibitem{miech19howto100m}
Antoine Miech, Dimitri Zhukov, Jean-Baptiste Alayrac, Makarand Tapaswi, Ivan
  Laptev, and Josef Sivic.
\newblock How{T}o100{M}: {L}earning a {T}ext-{V}ideo {E}mbedding by {W}atching
  {H}undred {M}illion {N}arrated {V}ideo {C}lips.
\newblock In {\em ICCV}, 2019.

\bibitem{misra2016shuffle}
Ishan Misra, C~Lawrence Zitnick, and Martial Hebert.
\newblock Shuffle and learn: unsupervised learning using temporal order
  verification.
\newblock In {\em European Conference on Computer Vision}, pages 527--544.
  Springer, 2016.

\bibitem{mit}
Mathew Monfort, Alex Andonian, Bolei Zhou, Kandan Ramakrishnan, Sarah~Adel
  Bargal, Tom Yan, Lisa Brown, Quanfu Fan, Dan Gutfruend, Carl Vondrick, et~al.
\newblock {Moments in Time Dataset: One Million Videos for Event
  Understanding}.
\newblock {\em IEEE Transactions on Pattern Analysis and Machine Intelligence
  (PAMI)}, 2019.

\bibitem{monfort2019multi}
Mathew Monfort, Kandan Ramakrishnan, Alex Andonian, Barry~A McNamara, Alex
  Lascelles, Bowen Pan, Quanfu Fan, Dan Gutfreund, Rogerio Feris, and Aude
  Oliva.
\newblock Multi-moments in time: Learning and interpreting models for
  multi-action video understanding.
\newblock {\em arXiv preprint arXiv:1911.00232}, 2019.

\bibitem{morgado_avid_cma}
Pedro Morgado, Nuno Vasconcelos, and Ishan Misra.
\newblock Audio-visual instance discrimination with cross-modal agreement.
\newblock 2020.

\bibitem{munro_CVPR2020_MMDA}
Jonathan Munro and Dima Damen.
\newblock {Multi-Modal Domain Adaptation for Fine-Grained Action Recognition}.
\newblock In {\em The IEEE Conference on Computer Vision and Pattern
  Recognition (CVPR)}, 2020.

\bibitem{ng_WACV2018_actionflownet}
Joe Yue-Hei Ng, Jonghyun Choi, Jan Neumann, and Larry~S. Davis.
\newblock {ActionFlowNet: Learning Motion Representation for Action
  Recognition}.
\newblock In {\em The IEEE Winter Conference on Applications of Computer Vision
  (WACV)}, 2018.

\bibitem{ng_WACV2018_TDN}
Joe Yue-Hei Ng and Larry~S. Davis.
\newblock {Temporal Difference Networks for Video Action Recognition}.
\newblock In {\em The IEEE Winter Conference on Applications of Computer Vision
  (WACV)}, 2018.

\bibitem{ngiam2011multimodal}
Jiquan Ngiam, Aditya Khosla, Mingyu Kim, Juhan Nam, Honglak Lee, and Andrew~Y
  Ng.
\newblock Multimodal deep learning.
\newblock In {\em ICML}, 2011.

\bibitem{nguyen2018weakly}
Phuc Nguyen, Ting Liu, Gautam Prasad, and Bohyung Han.
\newblock Weakly supervised action localization by sparse temporal pooling
  network.
\newblock In {\em Proceedings of the IEEE Conference on Computer Vision and
  Pattern Recognition}, pages 6752--6761, 2018.

\bibitem{nie_CVPR2015_jointPose}
Bruce~Xiaohan Nie, Caiming Xiong, and Song-Chun Zhu.
\newblock {Joint Action Recognition and Pose Estimation From Video}.
\newblock In {\em The IEEE Conference on Computer Vision and Pattern
  Recognition (CVPR)}, 2015.

\bibitem{oord2018representation}
Aaron van~den Oord, Yazhe Li, and Oriol Vinyals.
\newblock Representation learning with contrastive predictive coding.
\newblock {\em arXiv preprint arXiv:1807.03748}, 2018.

\bibitem{pan_AAAI2020_coAttention}
Boxiao Pan, Zhangjie Cao, Ehsan Adeli, and Juan~Carlos Niebles.
\newblock {Adversarial Cross-Domain Action Recognition with Co-Attention}.
\newblock In {\em The Conference on Artificial Intelligence (AAAI)}, 2020.

\bibitem{pathakCVPR16context}
Deepak Pathak, Philipp Kr\"ahenb\"uhl, Jeff Donahue, Trevor Darrell, and
  Alexei~A. Efros.
\newblock Context encoders: Feature learning by inpainting.
\newblock In {\em CVPR}, 2016.

\bibitem{Patrick20}
Mandela Patrick, Yuki~M. Asano, Ruth Fong, Jo{\~{a}}o~F. Henriques, Geoffrey
  Zweig, and Andrea Vedaldi.
\newblock Multi-modal self-supervision from generalized data transformations.
\newblock {\em arXiv preprint arXiv:2003.04298}, 2020.

\bibitem{paul2018w}
Sujoy Paul, Sourya Roy, and Amit~K Roy-Chowdhury.
\newblock W-talc: Weakly-supervised temporal activity localization and
  classification.
\newblock In {\em Proceedings of the European Conference on Computer Vision
  (ECCV)}, pages 563--579, 2018.

\bibitem{peng_ICIP2019_nas}
Wei Peng, Xiaopeng Hong, and Guoying Zhao.
\newblock {Video Action Recognition Via Neural Architecture Searching}.
\newblock In {\em The IEEE International Conference on Image Processing
  (ICIP)}, 2019.

\bibitem{peng_arxiv2014_bovwSurvey}
Xiaojiang Peng, Limin Wang, Xingxing Wang, and Yu Qiao.
\newblock {Bag of Visual Words and Fusion Methods for Action Recognition:
  Comprehensive Study and Good Practice}.
\newblock {\em arXiv preprint arXiv:1405.4506}, 2014.

\bibitem{peng_ECCV2014_stackedFV}
Xiaojiang Peng, Changqing Zou, Yu Qiao, and Qiang Peng.
\newblock {Action Recognition with Stacked Fisher Vectors}.
\newblock In {\em The European Conference on Computer Vision (ECCV)}, 2014.

\bibitem{toby_CVPR2019_DDLSTM}
Toby Perrett and Dima Damen.
\newblock {DDLSTM: Dual-Domain LSTM for Cross-Dataset Action Recognition}.
\newblock In {\em The IEEE Conference on Computer Vision and Pattern
  Recognition (CVPR)}, 2019.

\bibitem{pham_ICML2018_enas}
Hieu Pham, Melody~Y. Guan, Barret Zoph, Quoc~V. Le, and Jeff Dean.
\newblock {Efficient Neural Architecture Search via Parameter Sharing}.
\newblock In {\em The International Conference on Machine Learning (ICML)},
  2018.

\bibitem{piergiovanni_arxiv2019_TVN}
AJ Piergiovanni, Anelia Angelova, and Michael~S. Ryoo.
\newblock {Tiny Video Networks}.
\newblock {\em arXiv preprint arXiv:1910.06961}, 2019.

\bibitem{pier20evolving}
AJ Piergiovanni, Anelia Angelova, and Michael~S Ryoo.
\newblock {Evolving Losses for Unsupervised Video Representation Learning}.
\newblock In {\em Proceedings of the IEEE/CVF Conference on Computer Vision and
  Pattern Recognition}, pages 133--142, 2020.

\bibitem{piergiovanni_ICCV2019_evolve}
AJ Piergiovanni, Anelia Angelova, Alexander Toshev, and Michael~S. Ryoo.
\newblock {Evolving Space-Time Neural Architectures for Videos}.
\newblock In {\em The IEEE International Conference on Computer Vision (ICCV)},
  2019.

\bibitem{piergiovanni_CVPR2019_repFlow}
AJ Piergiovanni and Michael~S. Ryoo.
\newblock {Representation Flow for Action Recognition}.
\newblock In {\em The IEEE Conference on Computer Vision and Pattern
  Recognition (CVPR)}, 2019.

\bibitem{piergiovanni2020avid}
AJ Piergiovanni and Michael~S. Ryoo.
\newblock Avid dataset: Anonymized videos from diverse countries, 2020.

\bibitem{piergiovanni2020learning}
AJ Piergiovanni and Michael~S. Ryoo.
\newblock Learning multimodal representations for unseen activities, 2020.

\bibitem{Qian20}
Rui Qian, Tianjian Meng, Boqing Gong, Ming-Hsuan Yang, Huisheng Wang, Serge
  Belongie, and Yin Cui.
\newblock Spatiotemporal contrastive video representation learning.
\newblock {\em arXiv preprint arXiv:2008.03800}, 2020.

\bibitem{ecoc_zsar_Qin_CVPR17}
Jie Qin, Li Liu, Ling Shao, Fumin Shen, Bingbing Ni, Jiaxin Chen, and Yunhong
  Wang.
\newblock {Zero-Shot Action Recognition with Error-Correcting Output Codes}.
\newblock In {\em CVPR}, 2017.

\bibitem{qiu_ICCV2017_P3D}
Zhaofan Qiu, Ting Yao, and Tao Mei.
\newblock {Learning Spatio-Temporal Representation with Pseudo-3D Residual
  Networks}.
\newblock In {\em The IEEE International Conference on Computer Vision (ICCV)},
  2017.

\bibitem{radosavovic_CVPR2020_RegNet}
Ilija Radosavovic, Raj~Prateek Kosaraju, Ross Girshick, Kaiming He, and Piotr
  Doll{\'a}r.
\newblock {Designing Network Design Spaces}.
\newblock In {\em The IEEE Conference on Computer Vision and Pattern
  Recognition (CVPR)}, 2020.

\bibitem{ren_NIPS2015_fasterRCNN}
Shaoqing Ren, Kaiming He, Ross Girshick, and Jian Sun.
\newblock {Faster {R-CNN}: Towards Real-Time Object Detection with Region
  Proposal Networks}.
\newblock In {\em Advances in Neural Information Processing Systems (NeurIPS)},
  2015.

\bibitem{richard2017weakly}
Alexander Richard, Hilde Kuehne, and Juergen Gall.
\newblock Weakly supervised action learning with rnn based fine-to-coarse
  modeling.
\newblock In {\em Proceedings of the IEEE Conference on Computer Vision and
  Pattern Recognition}, pages 754--763, 2017.

\bibitem{moreno_Sensors2019_survey}
Itsaso Rodríguez-Moreno, José~María Martínez-Otzeta, Basilio Sierra, Igor
  Rodriguez, and Ekaitz Jauregi.
\newblock {Video Activity Recognition: State-of-the-Art}.
\newblock {\em Sensors}, 2019.

\bibitem{rohrbach15ijcv}
Marcus Rohrbach, Anna Rohrbach, Michaela Regneri, Sikandar Amin, Mykhaylo
  Andriluka, Manfred Pinkal, and Bernt Schiele.
\newblock Recognizing fine-grained and composite activities using hand-centric
  features and script data.
\newblock {\em International Journal of Computer Vision}, pages 1--28, 2015.

\bibitem{sarmiento2019mmp}
C. {Roig}, M. {Sarmiento}, D. {Varas}, I. {Masuda}, J.~C. {Riveiro}, and E.
  {Bou-Balust}.
\newblock Multi-modal pyramid feature combination for human action recognition.
\newblock In {\em 2019 IEEE/CVF International Conference on Computer Vision
  Workshop (ICCVW)}, pages 3742--3746, 2019.

\bibitem{rouditchenko2020avlnet}
Andrew Rouditchenko, Angie Boggust, David Harwath, Dhiraj Joshi, Samuel Thomas,
  Kartik Audhkhasi, Rogerio Feris, Brian Kingsbury, Michael Picheny, Antonio
  Torralba, and James Glass.
\newblock Avlnet: Learning audio-visual language representations from
  instructional videos, 2020.

\bibitem{ryoo2020assemblenet++}
Michael~S Ryoo, AJ Piergiovanni, Juhana Kangaspunta, and Anelia Angelova.
\newblock Assemblenet++: Assembling modality representations via attention
  connections.
\newblock {\em arXiv preprint arXiv:2008.08072}, 2020.

\bibitem{ryoo_ICLR2020_assembleNet}
Michael~S. Ryoo, AJ Piergiovanni, Mingxing Tan, and Anelia Angelova.
\newblock {AssembleNet: Searching for Multi-Stream Neural Connectivity in Video
  Architectures}.
\newblock In {\em The International Conference on Learning Representations
  (ICLR)}, 2020.

\bibitem{sanchez_IJCV2013_fisherVector}
Jorge Sanchez, Florent Perronnin, Thomas Mensink, and Jakob Verbeek.
\newblock {Image Classification with the Fisher Vector: Theory and Practice}.
\newblock {\em International Journal of Computer Vision (IJCV)}, 2013.

\bibitem{sener2020temporal}
Fadime Sener, Dipika Singhania, and Angela Yao.
\newblock Temporal aggregate representations for long-range video
  understanding.
\newblock In {\em European Conference on Computer Vision}, pages 154--171.
  Springer, 2020.

\bibitem{shao_CVPR2020_finegym}
Dian Shao, Yue Zhao, Bo Dai, and Dahua Lin.
\newblock {FineGym: A Hierarchical Video Dataset for Fine-grained Action
  Understanding}.
\newblock In {\em The IEEE Conference on Computer Vision and Pattern
  Recognition (CVPR)}, 2020.

\bibitem{shao2020temporal}
Hao Shao, Shengju Qian, and Yu Liu.
\newblock Temporal interlacing network, 2020.

\bibitem{Shi_ICCV2017_bioinspire}
Yemin Shi, Yonghong Tian, Yaowei Wang, Wei Zeng, and Tiejun Huang.
\newblock {Learning Long-Term Dependencies for Action Recognition With a
  Biologically-Inspired Deep Network}.
\newblock In {\em The IEEE International Conference on Computer Vision (ICCV)},
  2017.

\bibitem{shou2018autoloc}
Zheng Shou, Hang Gao, Lei Zhang, Kazuyuki Miyazawa, and Shih-Fu Chang.
\newblock Autoloc: Weakly-supervised temporal action localization in untrimmed
  videos.
\newblock In {\em Proceedings of the European Conference on Computer Vision
  (ECCV)}, pages 154--171, 2018.

\bibitem{shou_CVPR2019_DMCNet}
Zheng Shou, Xudong Lin, Yannis Kalantidis, Laura Sevilla-Lara, Marcus Rohrbach,
  Shih-Fu Chang, and Zhicheng Yan.
\newblock {DMC-Net: Generating Discriminative Motion Cues for Fast Compressed
  Video Action Recognition}.
\newblock In {\em The IEEE Conference on Computer Vision and Pattern
  Recognition (CVPR)}, 2019.

\bibitem{charades}
Gunnar~A. Sigurdsson, Gül Varol, Xiaolong Wang, Ali Farhadi, Ivan Laptev, and
  Abhinav Gupta.
\newblock {Hollywood in Homes: Crowdsourcing Data Collection for Activity
  Understanding}.
\newblock In {\em The European Conference on Computer Vision (ECCV)}, 2016.

\bibitem{simonyan_NIPS2014_twoStream}
Karen Simonyan and Andrew Zisserman.
\newblock {Two-Stream Convolutional Networks for Action Recognition in Videos}.
\newblock In {\em Advances in Neural Information Processing Systems (NeurIPS)},
  2014.

\bibitem{simonyan_ICLR2015_VGG}
Karen Simonyan and Andrew Zisserman.
\newblock {Very Deep Convolutional Networks for Large-Scale Image Recognition}.
\newblock In {\em The International Conference on Learning Representations
  (ICLR)}, 2015.

\bibitem{singh_CVPR2016_BiRNN}
Bharat Singh, Tim~K. Marks, Michael Jones, Oncel Tuzel, and Ming Shao.
\newblock {A Multi-Stream Bi-Directional Recurrent Neural Network for
  Fine-Grained Action Detection}.
\newblock In {\em The IEEE Conference on Computer Vision and Pattern
  Recognition (CVPR)}, 2016.

\bibitem{ucf101}
Khurram Soomro, Amir~Roshan Zamir, and Mubarak Shah.
\newblock {UCF101: A Dataset of 101 Human Actions Classes From Videos in The
  Wild}.
\newblock {\em arXiv preprint arXiv:1212.0402}, 2012.

\bibitem{stroud_WACV2020_D3D}
Jonathan~C. Stroud, David~A. Ross, Chen Sun, Jia Deng, and Rahul Sukthankar.
\newblock {D3D: Distilled 3D Networks for Video Action Recognition}.
\newblock In {\em The IEEE Winter Conference on Applications of Computer Vision
  (WACV)}, 2020.

\bibitem{Sudhakaran_2019_CVPR}
Swathikiran Sudhakaran, Sergio Escalera, and Oswald Lanz.
\newblock Lsta: Long short-term attention for egocentric action recognition.
\newblock In {\em CVPR}, 2019.

\bibitem{sultani_CVPR2014_FWHD}
Waqas Sultani and Imran Saleemi.
\newblock {Human Action Recognition across Datasets by Foreground-Weighted
  Histogram Decomposition}.
\newblock In {\em The IEEE Conference on Computer Vision and Pattern
  Recognition (CVPR)}, 2014.

\bibitem{sun2019learning}
Chen Sun, Fabien Baradel, Kevin Murphy, and Cordelia Schmid.
\newblock Learning video representations using contrastive bidirectional
  transformer, 2019.

\bibitem{sun_ICCV2019_videoBERT}
Chen Sun, Austin Myers, Carl Vondrick, Kevin Murphy, and Cordelia Schmid.
\newblock {VideoBERT: A Joint Model for Video and Language Representation
  Learning}.
\newblock In {\em The IEEE International Conference on Computer Vision (ICCV)},
  2019.

\bibitem{sun_ICCV2017_L2STM}
Lin Sun, Kui Jia, Kevin Chen, Dit-Yan Yeung, Bertram~E. Shi, and Silvio
  Savarese.
\newblock {Lattice Long Short-Term Memory for Human Action Recognition}.
\newblock In {\em The IEEE International Conference on Computer Vision (ICCV)},
  2017.

\bibitem{sun_CVPR2018_OFF}
Shuyang Sun, Zhanghui Kuang, Lu Sheng, Wanli Ouyang, and Wei Zhang.
\newblock {Optical Flow Guided Feature: A Fast and Robust Motion Representation
  for Video Action Recognition}.
\newblock In {\em The IEEE Conference on Computer Vision and Pattern
  Recognition (CVPR)}, 2018.

\bibitem{szegedy_CVPR2015_inception}
Christian Szegedy, Wei Liu, Yangqing Jia, Pierre Sermanet, Scott Reed, Dragomir
  Anguelov, Dumitru Erhan, Vincent Vanhoucke, and Andrew Rabinovich.
\newblock {Going Deeper with Convolutions}.
\newblock In {\em The IEEE Conference on Computer Vision and Pattern
  Recognition (CVPR)}, 2015.

\bibitem{szegedy_2013_intriguing}
Christian Szegedy, Wojciech Zaremba, Ilya Sutskever, Joan Bruna, Dumitru Erhan,
  Ian Goodfellow, and Rob Fergus.
\newblock {Intriguing Properties of Neural Networks}.
\newblock {\em arXiv preprint arXiv:1312.6199}, 2013.

\bibitem{taylor_ECCV2010_CLST}
Graham~W. Taylor, Rob Fergus, Yann LeCun, and Christoph Bregler.
\newblock {Convolutional Learning of Spatio-temporal Features}.
\newblock In {\em The European Conference on Computer Vision (ECCV)}, 2010.

\bibitem{tian2019contrastive}
Yonglong Tian, Dilip Krishnan, and Phillip Isola.
\newblock Contrastive multiview coding.
\newblock {\em arXiv preprint arXiv:1906.05849}, 2019.

\bibitem{tran_ICCV2015_C3D}
Du Tran, Lubomir Bourdev, Rob Fergus, Lorenzo Torresani, and Manohar Paluri.
\newblock {Learning Spatiotemporal Features with 3D Convolutional Networks}.
\newblock In {\em The IEEE International Conference on Computer Vision (ICCV)},
  2015.

\bibitem{tran_ICCV2019_CSN}
Du Tran, Heng Wang, Lorenzo Torresani, and Matt Feiszli.
\newblock {Video Classification With Channel-Separated Convolutional Networks}.
\newblock In {\em The IEEE International Conference on Computer Vision (ICCV)},
  2019.

\bibitem{tran_CVPR2018_R2plus1D}
Du Tran, Heng Wang, Lorenzo Torresani, Jamie Ray, Yann LeCun, and Manohar
  Paluri.
\newblock {A Closer Look at Spatiotemporal Convolutions for Action
  Recognition}.
\newblock In {\em The IEEE Conference on Computer Vision and Pattern
  Recognition (CVPR)}, 2018.

\bibitem{ullah_access2017_biLSTM}
A. {Ullah}, J. {Ahmad}, K. {Muhammad}, M. {Sajjad}, and S.~W. {Baik}.
\newblock {Action Recognition in Video Sequences using Deep Bi-Directional LSTM
  With CNN Features}.
\newblock {\em IEEE Access}, 2017.

\bibitem{varol_PAMI18_ltc}
G{\"u}l Varol, Ivan Laptev, and Cordelia Schmid.
\newblock {Long-term Temporal Convolutions for Action Recognition}.
\newblock {\em IEEE Transactions on Pattern Analysis and Machine Intelligence
  (PAMI)}, 2018.

\bibitem{vaswani_NIPS2017_attention}
Ashish Vaswani, Noam Shazeer, Niki Parmar, Jakob Uszkoreit, Llion Jones,
  Aidan~N Gomez, {\L}ukasz Kaiser, and Illia Polosukhin.
\newblock {Attention is All You Need}.
\newblock In {\em Advances in Neural Information Processing Systems (NeurIPS)},
  2017.

\bibitem{vondrick2016anticipating}
Carl Vondrick, Hamed Pirsiavash, and Antonio Torralba.
\newblock Anticipating visual representations from unlabeled video.
\newblock In {\em Proceedings of the IEEE Conference on Computer Vision and
  Pattern Recognition}, pages 98--106, 2016.

\bibitem{wang_CVPR2011_DT}
Heng Wang, Alexander Kläser, Cordelia Schmid, and Liu Cheng-Lin.
\newblock {Action Recognition by Dense Trajectories}.
\newblock In {\em The IEEE Conference on Computer Vision and Pattern
  Recognition (CVPR)}, 2011.

\bibitem{wang_ICCV2013_IDT}
Heng Wang and Cordelia Schmid.
\newblock {Action Recognition with Improved Trajectories}.
\newblock In {\em The IEEE International Conference on Computer Vision (ICCV)},
  2013.

\bibitem{wang2019self}
Jiangliu Wang, Jianbo Jiao, Linchao Bao, Shengfeng He, Yunhui Liu, and Wei Liu.
\newblock Self-supervised spatio-temporal representation learning for videos by
  predicting motion and appearance statistics.
\newblock In {\em Proceedings of the IEEE Conference on Computer Vision and
  Pattern Recognition}, pages 4006--4015, 2019.

\bibitem{wang_ICCV2019_hallucinating}
Lei Wang, Piotr Koniusz, and Du Huynh.
\newblock Hallucinating idt descriptors and i3d optical flow features for
  action recognition with cnns.
\newblock In {\em Proceedings of the 2019 International Conference on Computer
  Vision}. IEEE, Institute of Electrical and Electronics Engineers, 2019.

\bibitem{wang_CVPR2018_ARTNet}
Limin Wang, Wei Li, Wen Li, and Luc Van~Gool.
\newblock {Appearance-and-Relation Networks for Video Classification}.
\newblock In {\em The IEEE Conference on Computer Vision and Pattern
  Recognition (CVPR)}, 2018.

\bibitem{want_CVPR2015_trajectory}
Limin Wang, Yu Qiao, and Xiaoou Tang.
\newblock {Action Recognition With Trajectory-Pooled Deep-Convolutional
  Descriptors}.
\newblock In {\em The IEEE Conference on Computer Vision and Pattern
  Recognition (CVPR)}, 2015.

\bibitem{wang_THUMOS2015_challenge}
Limin Wang, Zhe Wang, Yuanjun Xiong, and Yu Qiao.
\newblock {CUHK and SIAT Submission for THUMOS15 Action Recognition Challenge}.
\newblock {\em THUMOS’15 Action Recognition Challenge}, 2015.

\bibitem{wang2017untrimmednets}
Limin Wang, Yuanjun Xiong, Dahua Lin, and Luc Van~Gool.
\newblock Untrimmednets for weakly supervised action recognition and detection.
\newblock In {\em Proceedings of the IEEE conference on Computer Vision and
  Pattern Recognition}, pages 4325--4334, 2017.

\bibitem{wang_arxiv2015_good}
Limin Wang, Yuanjun Xiong, Zhe Wang, and Yu Qiao.
\newblock {Towards Good Practices for Very Deep Two-Stream ConvNets}.
\newblock {\em arXiv preprint arXiv:1507.02159}, 2015.

\bibitem{wang_ECCV2016_TSN}
Limin Wang, Yuanjun Xiong, Zhe Wang, Yu Qiao, Dahua Lin, Xiaoou Tang, and Luc
  Van~Gool.
\newblock {Temporal Segment Networks: Towards Good Practices for Deep Action
  Recognition}.
\newblock In {\em The European Conference on Computer Vision (ECCV)}, 2016.

\bibitem{wang_CVPR2018_nonlocal}
Xiaolong Wang, Ross Girshick, Abhinav Gupta, and Kaiming He.
\newblock {Non-Local Neural Networks}.
\newblock In {\em The IEEE Conference on Computer Vision and Pattern
  Recognition (CVPR)}, 2018.

\bibitem{wang_ECCV2018_videoGraph}
Xiaolong Wang and Abhinav Gupta.
\newblock {Videos as Space-Time Region Graphs}.
\newblock In {\em The European Conference on Computer Vision (ECCV)}, 2018.

\bibitem{CVPR2019_CycleTime}
Xiaolong Wang, Allan Jabri, and Alexei~A. Efros.
\newblock {Learning Correspondence from the Cycle-Consistency of Time}.
\newblock In {\em CVPR}, 2019.

\bibitem{wang2020attentionnas}
Xiaofang Wang, Xuehan Xiong, Maxim Neumann, AJ Piergiovanni, Michael~S. Ryoo,
  Anelia Angelova, Kris~M. Kitani, and Wei Hua.
\newblock Attentionnas: Spatiotemporal attention cell search for video
  classification, 2020.

\bibitem{wang2020symbiotic}
Xiaohan Wang, Linchao Zhu, Yu Wu, and Yi Yang.
\newblock Symbiotic attention for egocentric action recognition with
  object-centric alignment.
\newblock {\em IEEE Transactions on Pattern Analysis and Machine Intelligence},
  2020.

\bibitem{wang2018pulling}
Yang Wang and Minh Hoai.
\newblock Pulling actions out of context: Explicit separation for effective
  combination.
\newblock In {\em Proceedings of the IEEE Conference on Computer Vision and
  Pattern Recognition}, pages 7044--7053, 2018.

\bibitem{wang_CVPR2017_pyramid}
Yunbo Wang, Mingsheng Long, Jianmin Wang, and Philip~S. Yu.
\newblock {Spatiotemporal Pyramid Network for Video Action Recognition}.
\newblock In {\em The IEEE Conference on Computer Vision and Pattern
  Recognition (CVPR)}, 2017.

\bibitem{wei2019heuristic}
Zhipeng Wei, Jingjing Chen, Xingxing Wei, Linxi Jiang, Tat-Seng Chua, Fengfeng
  Zhou, and Yu-Gang Jiang.
\newblock Heuristic black-box adversarial attacks on video recognition models.
\newblock {\em arXiv preprint arXiv:1911.09449}, 2019.

\bibitem{weinzaepfel2019mimetics}
Philippe Weinzaepfel and Gr{\'e}gory Rogez.
\newblock Mimetics: Towards understanding human actions out of context.
\newblock {\em arXiv preprint arXiv:1912.07249}, 2019.

\bibitem{wu_CVPR2018_shift}
Bichen Wu, Alvin Wan, Xiangyu Yue, Peter Jin, Sicheng Zhao, Noah Golmant, Amir
  Gholaminejad, Joseph Gonzalez, and Kurt Keutzer.
\newblock {Shift: A Zero FLOP, Zero Parameter Alternative to Spatial
  Convolutions}.
\newblock In {\em The IEEE Conference on Computer Vision and Pattern
  Recognition (CVPR)}, 2018.

\bibitem{wu_CVPR2019_featureBank}
Chao-Yuan Wu, Christoph Feichtenhofer, Haoqi Fan, Kaiming He, Philipp
  Krahenbuhl, and Ross Girshick.
\newblock {Long-Term Feature Banks for Detailed Video Understanding}.
\newblock In {\em The IEEE Conference on Computer Vision and Pattern
  Recognition (CVPR)}, 2019.

\bibitem{wu_CVPR2020_multigrid}
Chao-Yuan Wu, Ross Girshick, Kaiming He, Christoph Feichtenhofer, and Philipp
  Krähenbühl.
\newblock {A Multigrid Method for Efficiently Training Video Models}.
\newblock In {\em The IEEE Conference on Computer Vision and Pattern
  Recognition (CVPR)}, 2020.

\bibitem{wu_CVPR2018_compressed}
Chao-Yuan Wu, Manzil Zaheer, Hexiang Hu, R. Manmatha, Alexander~J. Smola, and
  Philipp Krähenbühl.
\newblock {Compressed Video Action Recognition}.
\newblock In {\em The IEEE Conference on Computer Vision and Pattern
  Recognition (CVPR)}, 2018.

\bibitem{wu_CVPR2016_objectScene}
Zuxuan Wu, Yanwei Fu, Yu-Gang Jiang, and Leonid Sigal.
\newblock {Harnessing Object and Scene Semantics for Large-Scale Video
  Understanding}.
\newblock In {\em The IEEE Conference on Computer Vision and Pattern
  Recognition (CVPR)}, 2016.

\bibitem{wu_MM2016_multiStream}
Zuxuan Wu, Yu-Gang Jiang, Xi Wang, Hao Ye, and Xiangyang Xue.
\newblock {Multi-Stream Multi-Class Fusion of Deep Networks for Video
  Classification}.
\newblock In {\em The ACM Conference on Multimedia (MM)}, 2016.

\bibitem{wu_CVPR2019_adaframe}
Zuxuan Wu, Caiming Xiong, Chih-Yao Ma, Richard Socher, and Larry~S. Davis.
\newblock {AdaFrame: Adaptive Frame Selection for Fast Video Recognition}.
\newblock In {\em The IEEE Conference on Computer Vision and Pattern
  Recognition (CVPR)}, 2019.

\bibitem{wu2018unsupervised}
Zhirong Wu, Yuanjun Xiong, Stella~X Yu, and Dahua Lin.
\newblock Unsupervised feature learning via non-parametric instance
  discrimination.
\newblock In {\em Proceedings of the IEEE Conference on Computer Vision and
  Pattern Recognition}, pages 3733--3742, 2018.

\bibitem{wu_arxiv2016_survey}
Zuxuan Wu, Ting Yao, Yanwei Fu, and Yu-Gang Jiang.
\newblock {Deep Learning for Video Classification and Captioning}.
\newblock {\em arXiv preprint arXiv:1609.06782}, 2016.

\bibitem{xiao_arxiv2020_audioslowfast}
Fanyi Xiao, Yong~Jae Lee, Kristen Grauman, Jitendra Malik, and Christoph
  Feichtenhofer.
\newblock {Audiovisual SlowFast Networks for Video Recognition}.
\newblock {\em arXiv preprint arXiv:2001.08740}, 2020.

\bibitem{xie_CVPR2017_resnext}
Saining Xie, Ross Girshick, Piotr Dollar, Zhuowen Tu, and Kaiming He.
\newblock {Aggregated Residual Transformations for Deep Neural Networks}.
\newblock In {\em The IEEE Conference on Computer Vision and Pattern
  Recognition (CVPR)}, 2017.

\bibitem{xie_ECCV2018_S3D}
Saining Xie, Chen Sun, Jonathan Huang, Zhuowen Tu, and Kevin Murphy.
\newblock {Rethinking Spatiotemporal Feature Learning: Speed-Accuracy
  Trade-offs in Video Classification}.
\newblock In {\em The European Conference on Computer Vision (ECCV)}, 2018.

\bibitem{Xu19vcop}
Dejing Xu, Jun Xiao, Zhou Zhao, Jian Shao, Di Xie, and Yueting Zhuang.
\newblock Self-supervised spatiotemporal learning via video clip order
  prediction.
\newblock 2019.

\bibitem{xu_IVC2016_many2one}
Tiantian Xu, Fan Zhu, Edward~K. Wong, and Yi Fang.
\newblock {Dual Many-to-One-Encoder-based Transfer Learning for Cross-Dataset
  Human Action Recognition}.
\newblock {\em Image and Vision Computing}, 2016.

\bibitem{mtl_zsar_Xu_ECCV16}
Xun Xu, Timothy Hospedales, and Shaogang Gong.
\newblock {Multi-Task Zero-Shot Action Recognition with Prioritised Data
  Augmentation}.
\newblock In {\em ECCV}, 2016.

\bibitem{tran_zsar_Xu_IJCV17}
Xun Xu, Timothy Hospedales, and Shaogang Gong.
\newblock {Transductive Zero-Shot Action Recognition by Word-Vector Embedding}.
\newblock {\em IJCV}, 2017.

\bibitem{xu_CVPR2015_LCD}
Zhongwen Xu, Yi Yang, and Alexander~G. Hauptmann.
\newblock {A Discriminative CNN Video Representation for Event Detection}.
\newblock In {\em The IEEE Conference on Computer Vision and Pattern
  Recognition (CVPR)}, 2015.

\bibitem{yan2020sparse}
Huanqian Yan, Xingxing Wei, and Bo Li.
\newblock Sparse black-box video attack with reinforcement learning.
\newblock {\em arXiv preprint arXiv:2001.03754}, 2020.

\bibitem{stgcn2018aaai}
Sijie Yan, Yuanjun Xiong, and Dahua Lin.
\newblock Spatial temporal graph convolutional networks for skeleton-based
  action recognition.
\newblock In {\em AAAI}, 2018.

\bibitem{yang_arxiv2020_vthcl}
Ceyuan Yang, Yinghao Xu, Bo Dai, and Bolei Zhou.
\newblock {Video Representation Learning with Visual Tempo Consistency}.
\newblock In {\em arXiv preprint arXiv:2006.15489}, 2020.

\bibitem{yang_CVPR2020_TPN}
Ceyuan Yang, Yinghao Xu, Jianping Shi, Bo Dai, and Bolei Zhou.
\newblock {Temporal Pyramid Network for Action Recognition}.
\newblock In {\em The IEEE Conference on Computer Vision and Pattern
  Recognition (CVPR)}, 2020.

\bibitem{yang2020hierarchical}
Xitong Yang, Xiaodong Yang, Sifei Liu, Deqing Sun, Larry Davis, and Jan Kautz.
\newblock Hierarchical contrastive motion learning for video action
  recognition, 2020.

\bibitem{li_ICCV2015_tempStructure}
Li Yao, Atousa Torabi, Kyunghyun Cho, Nicolas Ballas, Christopher Pal, Hugo
  Larochelle, and Aaron Courville.
\newblock {Describing Videos by Exploiting Temporal Structure}.
\newblock In {\em The IEEE International Conference on Computer Vision (ICCV)},
  2015.

\bibitem{yeung_CVPR2016_glimpses}
Serena Yeung, Olga Russakovsky, Greg Mori, and Li Fei-Fei.
\newblock {End-to-end learning of action detection from frame glimpses in
  videos}.
\newblock In {\em The IEEE Conference on Computer Vision and Pattern
  Recognition (CVPR)}, 2016.

\bibitem{wang_BMVC2016_SRCNN}
Wang Yifan, Jie Song, Limin Wang, Luc Van~Gool, and Otmar Hilliges.
\newblock {Two-Stream SR-CNNs for Action Recognition in Videos}.
\newblock In {\em The British Machine Vision Conference (BMVC)}, 2016.

\bibitem{ng_CVPR2015_beyondLSTM}
Joe Yue-Hei~Ng, Matthew Hausknecht, Sudheendra Vijayanarasimhan, Oriol Vinyals,
  Rajat Monga, and George Toderici.
\newblock {Beyond Short Snippets: Deep Networks for Video Classification}.
\newblock In {\em The IEEE Conference on Computer Vision and Pattern
  Recognition (CVPR)}, 2015.

\bibitem{yun_ICCV2019_cutmix}
Sangdoo Yun, Dongyoon Han, Seong~Joon Oh, Sanghyuk Chun, Junsuk Choe, and
  Youngjoon Yoo.
\newblock {CutMix: Regularization Strategy to Train Strong Classifiers With
  Localizable Features}.
\newblock In {\em The IEEE International Conference on Computer Vision (ICCV)},
  2019.

\bibitem{zach_PRS2007_TVL1}
Christopher Zach, Thomas Pock, and Horst Bischof.
\newblock {A Duality based Approach for Realtime TV-L1 Optical Flow}.
\newblock {\em Joint Pattern Recognition Symposium}, 2007.

\bibitem{zhang_CVPR2016_motionVec}
Bowen Zhang, Limin Wang, Zhe Wang, Yu Qiao, and Hanli Wang.
\newblock {Real-time Action Recognition with Enhanced Motion Vector CNNs}.
\newblock In {\em The IEEE Conference on Computer Vision and Pattern
  Recognition (CVPR)}, 2016.

\bibitem{zhang2020pan}
Can Zhang, Yuexian Zou, Guang Chen, and Lei Gan.
\newblock Pan: Towards fast action recognition via learning persistence of
  appearance, 2020.

\bibitem{zhang_ICLR2018_mixup}
Hongyi Zhang, Moustapha Cisse, Yann~N. Dauphin, and David Lopez-Paz.
\newblock {Mixup: Beyond Empirical Risk Minimization}.
\newblock In {\em The International Conference on Learning Representations
  (ICLR)}, 2018.

\bibitem{zhang2020resnest}
Hang Zhang, Chongruo Wu, Zhongyue Zhang, Yi Zhu, Zhi Zhang, Haibin Lin, Yue
  Sun, Tong He, Jonas Muller, R. Manmatha, Mu Li, and Alexander Smola.
\newblock {ResNeSt: Split-Attention Networks}.
\newblock {\em arXiv preprint arXiv:2004.08955}, 2020.

\bibitem{zhang_ECCV2020_mesampler}
Hu Zhang, Linchao Zhu, Yi Zhu, and Yi Yang.
\newblock {Motion-Excited Sampler: Video Adversarial Attack with Sparked
  Prior}.
\newblock In {\em The European Conference on Computer Vision (ECCV)}, 2020.

\bibitem{zhang_Sensors2018_survey}
Hong-Bo Zhang, Yi-Xiang Zhang, Bineng Zhong, Qing Lei, Lijie Yang, Ji-Xiang Du,
  and Duan-Sheng Chen.
\newblock {A Comprehensive Survey of Vision-Based Human Action Recognition
  Methods}.
\newblock {\em Sensors}, 2018.

\bibitem{zhang2016colorful}
Richard Zhang, Phillip Isola, and Alexei~A Efros.
\newblock Colorful image colorization.
\newblock In {\em ECCV}, 2016.

\bibitem{Zhang_2017_CVPR}
Richard Zhang, Phillip Isola, and Alexei~A. Efros.
\newblock Split-brain autoencoders: Unsupervised learning by cross-channel
  prediction.
\newblock In {\em The IEEE Conference on Computer Vision and Pattern
  Recognition (CVPR)}, July 2017.

\bibitem{zhang_ICLR2020_V4D}
Shiwen Zhang, Sheng Guo, Weilin Huang, Matthew~R. Scott, and Limin Wang.
\newblock {V4D:4D Convolutional Neural Networks for Video-level Representation
  Learning}.
\newblock In {\em The International Conference on Learning Representations
  (ICLR)}, 2020.

\bibitem{zhang_CVPR2018_shufflenet}
Xiangyu Zhang, Xinyu Zhou, Mengxiao Lin, and Jian Sun.
\newblock {ShuffleNet: An Extremely Efficient Convolutional Neural Network for
  Mobile Devices}.
\newblock In {\em The IEEE Conference on Computer Vision and Pattern
  Recognition (CVPR)}, 2018.

\bibitem{zhang2020hierarchically}
Zehua Zhang and David Crandall.
\newblock Hierarchically decoupled spatial-temporal contrast for
  self-supervised video representation learning, 2020.

\bibitem{hacs}
Hang Zhao, Zhicheng Yan, Lorenzo Torresani, and Antonio Torralba.
\newblock {HACS: Human Action Clips and Segments Dataset for Recognition and
  Temporal Localization}.
\newblock {\em arXiv preprint arXiv:1712.09374}, 2019.

\bibitem{zhao_NIPS2018_trajconv}
Yue Zhao, Yuanjun Xiong, and Dahua Lin.
\newblock {Trajectory Convolution for Action Recognition}.
\newblock In {\em Advances in Neural Information Processing Systems (NeurIPS)},
  2018.

\bibitem{zhou_ECCV2018_TRN}
Bolei Zhou, Alex Andonian, Aude Oliva, and Antonio Torralba.
\newblock {Temporal Relational Reasoning in Videos}.
\newblock In {\em The European Conference on Computer Vision (ECCV)}, 2018.

\bibitem{zhou2017places}
Bolei Zhou, Agata Lapedriza, Aditya Khosla, Aude Oliva, and Antonio Torralba.
\newblock {Places: A 10 million Image Database for Scene Recognition}.
\newblock {\em IEEE Transactions on Pattern Analysis and Machine Intelligence
  (PAMI)}, 2017.

\bibitem{zhou_CVPR2018_MiCT}
Yizhou Zhou, Xiaoyan Sun, Zheng-Jun Zha, and Wenjun Zeng.
\newblock {MiCT: Mixed 3D/2D Convolutional Tube for Human Action Recognition}.
\newblock In {\em The IEEE Conference on Computer Vision and Pattern
  Recognition (CVPR)}, 2018.

\bibitem{zhu2020faster}
Linchao Zhu, Du Tran, Laura Sevilla-Lara, Yi Yang, Matt Feiszli, and Heng Wang.
\newblock Faster recurrent networks for efficient video classification.
\newblock In {\em AAAI}.

\bibitem{zhu_CVPR2017_multirate}
Linchao Zhu, Zhongwen Xu, and Yi Yang.
\newblock {Bidirectional Multirate Reconstruction for Temporal Modeling in
  Videos}.
\newblock In {\em The IEEE Conference on Computer Vision and Pattern
  Recognition (CVPR)}, 2017.

\bibitem{zhu_IJCV2017_vqa}
Linchao Zhu, Zhongwen Xu, Yi Yang, and Alex~G. Hauptmann.
\newblock {Uncovering Temporal Context for Video Question Answering}.
\newblock {\em International Journal of Computer Vision (IJCV)}, 2017.

\bibitem{zhu_CVPR2020_ActBERT}
Linchao Zhu and Yi Yang.
\newblock {ActBERT: Learning Global-Local Video-Text Representations}.
\newblock In {\em The IEEE Conference on Computer Vision and Pattern
  Recognition (CVPR)}, 2020.

\bibitem{zhu2020a3d}
Sijie Zhu, Taojiannan Yang, Matias Mendieta, and Chen Chen.
\newblock A3d: Adaptive 3d networks for video action recognition, 2020.

\bibitem{zhu_CVPR2016_keyVolume}
Wangjiang Zhu, Jie Hu, Gang Sun, Xudong Cao, and Yu Qiao.
\newblock {A Key Volume Mining Deep Framework for Action Recognition}.
\newblock In {\em The IEEE Conference on Computer Vision and Pattern
  Recognition (CVPR)}, 2016.

\bibitem{zhu_ACCV2018_hidden}
Yi Zhu, Zhenzhong Lan, Shawn Newsam, and Alexander~G. Hauptmann.
\newblock {Hidden Two-Stream Convolutional Networks for Action Recognition}.
\newblock In {\em The Asian Conference on Computer Vision (ACCV)}, 2018.

\bibitem{Zhu_CVPR2018_URL}
Yi Zhu, Yang Long, Yu Guan, Shawn Newsam, and Ling Shao.
\newblock {Towards Universal Representation for Unseen Action Recognition}.
\newblock In {\em The IEEE Conference on Computer Vision and Pattern
  Recognition (CVPR)}, 2018.

\bibitem{zhu_ECCVW2016_depth2action}
Yi Zhu and Shawn Newsam.
\newblock {Depth2Action: Exploring Embedded Depth for Large-Scale Action
  Recognition}.
\newblock In {\em The European Conference on Computer Vision (ECCV) Workshop},
  2016.

\bibitem{zhu_ACCV2018_RTS}
Yi Zhu and Shawn Newsam.
\newblock {Random Temporal Skipping for Multirate Video Analysis}.
\newblock In {\em The Asian Conference on Computer Vision (ACCV)}, 2018.

\bibitem{zolfaghari_ICCV2017_chained}
Mohammadreza Zolfaghari, Gabriel~L. Oliveira, Nima Sedaghat, and Thomas Brox.
\newblock {Chained Multi-Stream Networks Exploiting Pose, Motion, and
  Appearance for Action Classification and Detection}.
\newblock In {\em The IEEE International Conference on Computer Vision (ICCV)},
  2017.

\bibitem{zolfaghari_ECCV2018_ECO}
Mohammadreza Zolfaghari, Kamaljeet Singh, and Thomas Brox.
\newblock {ECO: Efficient Convolutional Network for Online Video
  Understanding}.
\newblock In {\em The European Conference on Computer Vision (ECCV)}, 2018.

\end{thebibliography}
}

\end{document}